\documentclass[sn-apa]{sn-jnl}%

\usepackage{multirow}%
\usepackage{mathrsfs}%
\usepackage[title]{appendix}%
\usepackage{xcolor}%
\usepackage{textcomp}%
\usepackage{manyfoot}%
\usepackage{booktabs}%
\usepackage{listings}%

\newif\ifshowcomments								
\showcommentstrue				%
\showcommentsfalse				%

\usepackage{booktabs}
\usepackage{wrapfig}

\usepackage[all]{nowidow} %
\usepackage{amsmath}
\usepackage{amsfonts}
\usepackage{amssymb}
\usepackage{graphicx}
\graphicspath{{graphics/}}
\usepackage{trfsigns}
\usepackage{caption} %
\usepackage{subcaption} %
\usepackage{enumitem}
\usepackage{color}
\usepackage{calc}
\usepackage{tabularx}
\usepackage{booktabs}
\usepackage{ragged2e}
\usepackage{hyperref}

\usepackage[acronym]{glossaries}
\usepackage{siunitx}

\usepackage{xstring}
\usepackage{catchfile}

\usepackage{adjustbox}

\usepackage{lscape}

\usepackage{gensymb}
\newcommand{\textcite}[1]{\citet{#1}}
\renewcommand{\cite}[1]{\citep{#1}}

\newcommand{\secsummary}{\paragraph*{Summary}}
\newcommand{\secoutlook}{\paragraph*{Outlook}}
\newcommand{\ownsec}[1]{\paragraph*{#1}}

\ifshowcomments
    \newcommand{\margtodo}
    {\marginpar{\textbf{\textcolor{blue}{ToDo}}}{}}
    \newcommand{\todo}[1]
    {{\textbf{\textcolor{blue}{(\margtodo{}#1)}}}{}}
    
    \newcommand{\margmajortodo}
    {\marginpar{\textbf{\textcolor{red}{ToDo}}}{}}
    \newcommand{\imp}[1]
    {{\textbf{\textcolor{red}{(\margmajortodo{}#1)}}}{}}
    
    \newcommand{\margidea}
    {\marginpar{\textbf{\textcolor{green}{Check}}}{}}
    \newcommand{\tocheck}[1]
    {{\textbf{\textcolor{green}{(\margidea{}#1)}}}{}}
    
    \newcommand{\margtocheck}
    {\marginpar{\textbf{\textcolor{cyan}{Idea}}}{}}
    \newcommand{\idea}[1]
    {{\textbf{\textcolor{cyan}{(\margtocheck{}#1)}}}{}}
    
    \newcommand{\margtochange}
    {\marginpar{\textbf{\textcolor{cyan}{Change}}}{}}
    \newcommand{\tochange}[1]
    {{\textbf{\textcolor{cyan}{(\margtochange{}#1)}}}{}}
\else
    \newcommand{\todo}[1]{}
    \newcommand{\imp}[1]{}
    \newcommand{\idea}[1]{}
    \newcommand{\tocheck}[1]{}
    \newcommand{\tochange}[1]{}
    
\fi
\definecolor{mygrey}{gray}{0.65}
\ifshowcomments  %
    
    \newcommand{\pcite}[2]{{\color{gray}(\cite{#1} #2)}}  %
    \newcommand{\margcomment}
    {\marginpar{\textbf{\textcolor{mygrey}{Com.}}}{}}
    \newcommand{\scomment}[1]
    {{\textit{\textcolor{mygrey}{(\margcomment{}#1)}}}{}}
\else  %
    \usepackage{comment}
    \newcommand{\pcite}[2]{}
    \newcommand{\scomment}[1]{}
    
\fi
\newcommand{\wrt}{w.r.t.\ } %

\newcommand{\idest}{i.e.\ } %
\newcommand{\eg}{e.g.\ } %
\newcommand{\cf}{cf.\ } %

\newcommand{\sota}{state-of-the-art}
\newcommand{\forklift}{forklift truck}
\newcommand{\rgbd}{RGB-D}
\newcommand{\kinect}{MS Kinect}

\newcommand{\maskrcnn}{Mask {R-CNN}}
\newcommand{\cubercnn}{Cube {R-CNN}}
\newcommand{\fasterrcnn}{Faster {R-CNN}}
\newcommand{\resnetfpn}{ResNet-50-FPN}
\newcommand{\cuberefine}{CubeRefine R-CNN}

\newcommand{\dsparcel}{Parcel3D}

\newcommand{\ap}[1]{$\text{AP}_{#1}$}
\newcommand{\boxap}[1]{$\text{Box AP}_{#1}$}
\newcommand{\maskap}[1]{$\text{Mask AP}_{#1}$}
\newcommand{\meshap}[1]{$\text{Mesh AP}_{#1}$}

\newcommand{\urlreview}{\href{https://a-nau.github.io/cv-in-logistics}{https://a-nau.github.io/cv-in-logistics}}

\newglossaryentry{i4.0}{name={Industry 4.0}, description={Industry 4.0}}
\newglossaryentry{forklift}{name={forklift truck}, description={}, plural={forklift trucks}}
\newacronym[longplural={Automated Guided Vehicles}]{agv}{AGV}{automated guided vehicle}
\newacronym{rfid}{RFID}{Radio-Frequency Identification}

\newacronym[longplural={time-of-flight cameras}]{tofc}{ToF camera}{time-of-flight camera}
\newacronym{pmd}{PMD}{Photonic Mixing Device}
\newacronym{roi}{ROI}{Region of Interest}
\newacronym{iou}{IoU}{Intersection over Union}
\newacronym{ar}{AR}{Augmented Reality}
\newacronym[longplural={Light Detection And Ranging}]{lidar}{LiDAR}{Light Detection And Ranging}
\newacronym[longplural={Frames per Second}]{fps}{FPS}{Frame per Second}
\newacronym{ocr}{OCR}{Optical Character Recognition}
\newacronym{gso}{GSO}{Google Scanned Objects}
\newacronym{lld}{LLD}{Large Logo Dataset}
\newacronym{rocauc}{ROC-AUC}{Area under the Curve of the Receiver Operator Characteristic}

\newacronym{AI}{AI}{Artificial Intelligence}
\newacronym{ML}{ML}{Machine Learning}
\newacronym{ransac}{RANSAC}{Random Sampling Consensus}
\newacronym[longplural={Artificial Neural Networks}]{nn}{ANN}{Artificial Neural Network}
\newacronym[longplural={Convolutional Neural Networks}]{cnn}{CNN}{Convolutional Neural Network}
\newacronym[longplural={Graph Convolutional Neural Networks}]{gcn}{GCN}{Graph Convolutional Neural Network}
\newacronym[longplural={Graph Neural Networks}]{gnn}{GNN}{Graph Neural Network}
\newacronym[longplural={Support Vector Machines}]{svm}{SVM}{Support Vector Machine}
\newacronym{dbscan}{DBSCAN}{Density-Based Spatial Clustering of Applications with Noise}
\newacronym{sgdm}{SGD+M}{Stochastic Gradient Descent with Momentum}
\newacronym{hmm}{HMM}{Hidden Markov Model}
\newacronym{hci}{HCI}{Human-Computer-Interaction}

\newacronym{eu}{EU}{European Union}

\newglossaryentry{poc}{name={proof of concept}, description={}}
\newglossaryentry{sota}{name={state-of-the-art}, description={}}

\makeglossaries

\usepackage[capitalize]{cleveref}
\crefname{section}{Section}{Sections}
\Crefname{section}{Section}{Sections}
\Crefname{table}{Table}{Tables}
\crefname{table}{Table}{Tables}
\usepackage{etoolbox}

\begin{document}

\title[Literature Review: Computer Vision Applications in Transportation Logistics and Warehousing]{Literature Review: Computer Vision Applications in Transportation Logistics and Warehousing}

\author[1,2]{\fnm{Alexander} \sur{Naumann}}\email{anaumann@fzi.de}

\author[1,2]{\fnm{Felix} \sur{Hertlein}}\email{hertlein@fzi.de}

\author[1,2]{\fnm{Laura} \sur{D\"orr}}\email{doerr@fzi.de}
\author[1,2]{\fnm{Steffen} \sur{Thoma}}\email{thoma@fzi.de}

\author[1,2]{\fnm{Kai} \sur{Furmans}}\email{kai.furmans@kit.edu}

\affil[1]{
    \orgname{FZI Research Center for Information Technology}%
}

\affil[2]{
    \orgname{Karlsruhe Institute of Technology (KIT)}%
}

\abstract{
    Computer vision applications in transportation logistics and warehousing have a huge potential for process automation.
    We present a structured literature review on research in the field to help leverage this potential.
    The literature is categorized \wrt the application, \idest the task it tackles and \wrt the computer vision techniques that are used.
    Regarding applications, we subdivide the literature in two areas: Monitoring, \idest observing and retrieving relevant information from the environment, and manipulation, where approaches are used to analyze and interact with the environment.
    Additionally, we point out directions for future research and link to recent developments in computer vision that are suitable for application in logistics.
    Finally, we present an overview of existing datasets and industrial solutions.
    The results of our analysis are also available online at \urlreview{}.
}

\maketitle

\section{Introduction}

\begin{figure}[t]
    \centering
    \includegraphics[width=0.6\textwidth]{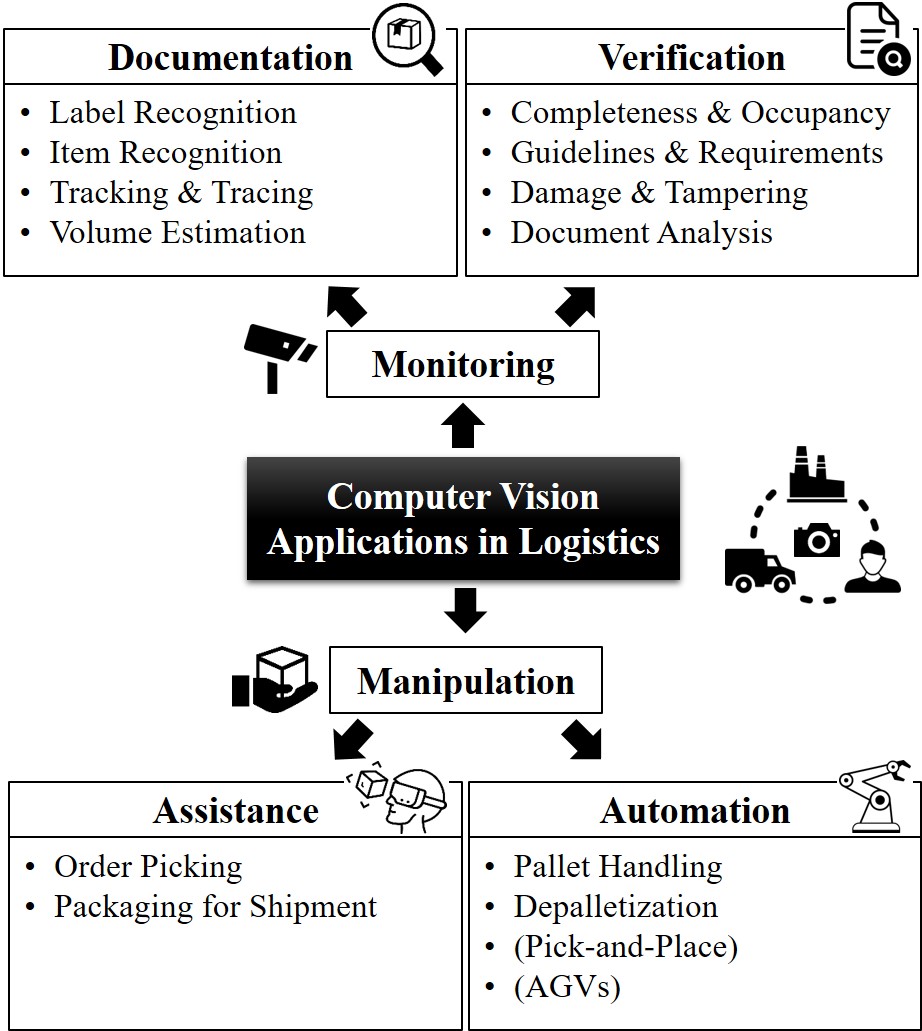}
    \vspace{0.3cm}
    \caption{
        Overview of the structure of this literature review.
        We provide tabular summaries for each area: Documentation (\cref{tab:overview:documentation}), Verification (\cref{tab:overview:verification}), Assistance (\cref{tab:overview:assistance_for_manipulation}) and Autonomous Manipulation (\cref{tab:overview:autonomous_manipulation}).
        Furthermore, we present a summary of datasets (\cref{sec:other:datasets}) and industry solutions (\cref{tab:overview:industry}).
    }
    \label{fig:overview}
\end{figure}

Companies frequently undervalue the strategic role of logistics \cite{tangStrategicRoleLogistics2019}, while at the same time, key trends in logistics are increasing cost pressure, process complexity and product individualization \cite{kerstenChancenDigitalenTransformation2017}.
To cope with these trends, technological advances are of major importance.
Especially transportation logistics and warehousing play a crucial role in every supply chain and have recently gained even higher importance due to the strongly growing e-commerce market.
Efficiently operating warehouses and transportation networks can reduce costs and thus, automating logistics processes is becoming increasingly important to handle the highly volatile shipping volumes \cite{azadehRobotizedAutomatedWarehouse2019}.
One promising path to increase efficiency is to utilize computer vision to automate time-consuming tasks.
We analyze the respective literature in this work and refer to \cref{fig:overview} for an overview of the topics covered.

There are literature reviews on artificial intelligence in smart logistics in general \cite{woschankReviewFurtherDirections2020}, and on the usage of machine learning for freight transportation and logistics applications \cite{tsolakiUtilizingMachineLearning2022a}.
Also, predictive maintenance of machines and the safety and security of operators in transportation networks are relevant research areas.
This work, however, focuses on freight handling using computer vision and does not review literature on downstream applications or machinery and operator considerations.
\textcite{borstellShortSurveyImage2018} presents a short survey of computer vision applications in logistics with a broader scope, however, no categorization from a computer vision perspective and no outlooks are presented.
\textcite{borstellBildbasierteZustandserfassungLogistik2021} extends this review and categorizes \wrt goods, operators, means of transport and space.
Unfortunately, only paper counts per category are given, without explicitly referring to the papers or presenting details.
We follow \textcite{borstellBildbasierteZustandserfassungLogistik2021} and adopt part of their categorization, such as distinguishing applications in monitoring and manipulation.
However, we are, to the best of the authors' knowledge, the first to present a detailed and structured review of the usage of computer vision in logistics contexts that covers applications, computer vision approaches and datasets as well as industrial solutions.
To summarize, the main contributions of our work are:

\begin{itemize}
    \item we present a detailed and structured review of the existing literature for computer vision in transportation logistics and warehousing with a categorization \wrt the application and the computer vision task at hand,
    \item we identify relevant areas in computer vision research which can foster further research on applications in transportation logistics and warehousing, and
    \item we present an overview of existing datasets and industry solutions.
\end{itemize}

~\\
In the following, we will first focus on the application-oriented view.
In \cref{sec:monitoring}, we review the literature on monitoring, in which the retrieval of environment information is considered. 
Next, in \cref{sec:manipulation}  we present applications where the interaction with the environment, \idest manipulation thereof, is a key factor.
After analyzing the literature from a practitioner's point of view, we categorize the literature also from a computer vision standpoint and present an overview of existing datasets and industrial solutions
in \cref{sec:other}.
The work concludes with \cref{sec:conclusion}.

\section{Monitoring}
\label{sec:monitoring}

Monitoring refers to observing processes in order to retrieve valuable information. 
The retrieval of information is based on visual perception. 
We distinguish two different cases: (1) Documentation, where we only want to retrieve information to document processes. 
Details are presented in \cref{sec:monitoring:documentation} and summarized in \cref{tab:overview:documentation}.
In this context, data is collected and stored, but not used as the basis for subsequent process automation.
(2) Verification, where the retrieved information is used to verify certain assumptions (\eg by comparing with a database of shipments). 
The relevant literature is discussed in \cref{sec:monitoring:verification} and summarized in \cref{tab:overview:verification}.
Note, that this application-oriented categorization causes large overlaps \wrt the computer vision approaches that are used and a separate categorization \wrt computer vision approaches will be presented in \cref{sec:other:cv}.

\subsection{Documentation}
\label{sec:monitoring:documentation}

The documentation of business processes is very important, especially in complex transportation networks, as frequently encountered in supply chains. 
To simplify the documentation of logistics processes, such as verifying incoming goods, several directions have been suggested in research. %
\begin{landscape}
    \begin{table}[t!]
    \centering\footnotesize
    \caption{Overview of the literature on Documentation (cf. \cref{sec:monitoring:documentation}).}
    \label{tab:overview:documentation}
    \begin{tabularx}{\linewidth}{>{\hsize=1.1\hsize\RaggedRight}X>{\hsize=1.8\hsize}X>{\hsize=1.2\hsize\RaggedRight}X>{\hsize=0.7\hsize\RaggedRight}X>{\hsize=0.3\hsize\RaggedRight}X>{\hsize=1.1\hsize\RaggedRight}X>{\hsize=.8\hsize\RaggedRight}X}
        \toprule
        Paper                                                    & Summary                                                                                             & Application                                           & Data Type                     & Public Dataset & Approach Type                        & Objects                                       \\
        \midrule
        \cite{brylkaAIbasedRecognitionDangerous2021}             & Detection of dangerous goods labels and volume estimation by pointcloud segmentation                & Depalletization, Label Recognition, Volume Estimation & RGB, \rgbd{}, Real, Synthetic &                & Deep Learning                        & Label, Pallet, Parcel                         \\
        \cite{mishraDevelopmentLowcostEmbedded2019}              & Low cost embedded vision system for the detection of 1D barcodes                                    & Label Recognition                                     & RGB                           &                & Classical Approach                   & Label                                         \\
        \cite{dorrLeanTrainingData2019}                          & Training data generation for shipping label recognition                                             & Label Recognition                                     & RGB, Synthetic                &                & Deep Learning                        & Label                                         \\
        \cite{suhRobustShippingLabel2019}                        & Label recognition by first detecting barcodes and then using this information for angle calibration & Label Recognition                                     & RGB                           &                & Classical Approach, Deep Learning    & Label                                         \\
        \cite{brylkaCameraBasedBarcode2020}                      & Detecting barcodes in images                                                                        & Label Recognition                                     & RGB, Synthetic                &                & Deep Learning                        & Label                                         \\
        \cite{wudhikarnDeepLearningBarcode2022}                  & Review on barcode detection                                                                         & Label Recognition                                     &                               &                &                                      & Label                                         \\
        \cite{kamnardsiri1DBarcodeDetection2022}                 & Comparison of CNNs for barcode detection                                                            & Label Recognition                                     & RGB, Real                     &                & Deep Learning                        & Label, Parcel                                 \\
        \cite{mayershoferFullySyntheticTrainingIndustrial2020}   & Synthetic dataset generation for applications in logistics                                          & Item Recognition                                      & RGB, Synthetic                &                & Deep Learning                        & Small Load Carrier                            \\
        \cite{mayershoferLOCOLogisticsObjects2020}               & Dataset for item recognition in logistics contexts                                                  & Item Recognition                                      & RGB, Real                     & x              & Deep Learning                        & Fork Lift, Pallet, Parcel, Small Load Carrier \\
        \cite{naumannRefinedPlaneSegmentation2020}               & Parcel side surface segmentation by exploiting plane detection                                      & Item Recognition                                      & RGB, Real                     & x              & Deep Learning                        & Parcel                                        \\
        \cite{naumannScrapeCutPasteLearn2022}                    & Automated instance segmentation dataset generation applied to parcel logistics                      & Item Recognition                                      & RGB, Synthetic                &                & Deep Learning                        & Arbitrary, Parcel                             \\
        \cite{shettyOpticalContainerCode2012}                    & Framework for OCR on containers                                                                     & Item Recognition                                      & RGB                           &                &                                      & Container/ Trailer                            \\
        \cite{dorrFullyAutomatedPackagingStructure2020}          & Localization of pallets and the analysis of their composition                                       & Item Recognition, Verify Completeness                 & RGB, Real                     &                & Deep Learning                        & Pallet, Small Load Carrier                    \\
        \cite{dorrTetraPackNetFourCornerBasedObject2021}         & Localization of pallets and the analysis of their composition                                       & Item Recognition, Verify Completeness                 & RGB, Real                     &                & Deep Learning                        & Pallet, Small Load Carrier                    \\
        \cite{weichertMarkerbasedTrackingSupport2010}            & Continuous detection, localization, and identification of parcels and bins in logistics processes   & Tracking and Tracing                                  & RGB                           &                & Classical Approach, Fiducial Markers & Container/ Trailer, Parcel                     \\
        \cite{borstellPalletMonitoringSystem2014}                & System for pallet monitoring                                                                        & Tracking and Tracing                                  & RGB, \rgbd{}, Real            &                & Classical Approach, Fiducial Markers & Pallet                                        \\
        \cite{clausenParcelTrackingDetection2019}                & Industry scale approach for tracking parcels in a logistics facility.                               & Tracking and Tracing                                  & RGB, Real                     &                & Deep Learning, Fiducial Markers      & Parcel                                        \\
        \cite{huCuboidDetectionTracking2021}                     & Tracking parcels inside a moving truck                                                              & Tracking and Tracing                                  & \rgbd{}, Real                 &                & Deep Learning                        & Parcel                                        \\
        \cite{rutinowskiReIdentificationWarehousingEntities2021} & Re-identification for chipwood pallet blocks of Euro pallets                                        & Tracking and Tracing                                  & RGB, Real                     & x              & Deep Learning                        & Pallet                                        \\
        \bottomrule
    \end{tabularx}
\end{table}

\begin{table}[t!]
    \centering\footnotesize
    \caption*{\textbf{\Cref{tab:overview:documentation}} (continuation): Overview of the literature in Documentation (cf. \cref{sec:monitoring:documentation}).}
    \label{tab:overview:documentation:b}
    \begin{tabularx}{\linewidth}{>{\hsize=1.1\hsize\RaggedRight}X>{\hsize=1.8\hsize}X>{\hsize=1.2\hsize\RaggedRight}X>{\hsize=0.7\hsize\RaggedRight}X>{\hsize=0.3\hsize\RaggedRight}X>{\hsize=1.1\hsize\RaggedRight}X>{\hsize=.8\hsize\RaggedRight}X}
        \toprule
        Paper                                                             & Summary                                                              & Application          & Data Type     & Public Dataset & Approach Type                        & Objects          \\
        \midrule
        \cite{kluttermannGraphRepresentationBased2022}                    & Re-identification for chipwood pallet blocks of Euro pallets         & Tracking and Tracing & RGB, Real     &                & Deep Learning                        & Pallet           \\
        \cite{rutinowskiDeepLearningBased2022}                            & Re-identification for chipwood pallet blocks of Euro pallets         & Tracking and Tracing & RGB, Real     & x              & Deep Learning                        & Pallet           \\
        \cite{borstellProzessintegrierteVolumenerfassungLogistischen2013} & Prototyped solutions for volume scanning using two MS Kinect cameras & Volume Estimation    & \rgbd{}, Real &                & Classical Approach                   & Pallet, Parcel   \\
        \cite{laotrakunchaiMeasurementSizeDistance2013}                   & Estimating the volume of a single parcel using a mobile device       & Volume Estimation    & RGB, Real     &                & Classical Approach, Pattern Matching & Parcel, Parcel   \\
        \cite{kucukDevelopmentDimensionsMeasurement2019}                  & Dimension estimation of static objects for logistic applications     & Volume Estimation    & \rgbd{}, Real &                & Classical Approach                   & Multiple, Parcel \\
        \bottomrule
    \end{tabularx}
\end{table}
    \begin{table}[t!]
    \centering\footnotesize
    \caption{Overview of the literature on Verification (cf. \cref{sec:monitoring:verification}).}
    \label{tab:overview:verification}
    \begin{tabularx}{\linewidth}{>{\hsize=1.1\hsize\RaggedRight}X>{\hsize=1.8\hsize}X>{\hsize=1.2\hsize\RaggedRight}X>{\hsize=0.7\hsize\RaggedRight}X>{\hsize=0.3\hsize\RaggedRight}X>{\hsize=1.1\hsize\RaggedRight}X>{\hsize=.8\hsize\RaggedRight}X}
        \toprule
        Paper                                            & Summary                                                                             & Application                           & Data Type            & Public Dataset & Approach Type                        & Objects                    \\
        \midrule
        \cite{nocetiMulticameraSystemDamage2018}         & Damage and tampering detection in a postal security framework from multiple cameras & Damage and Tampering Detection        & RGB, Real            &                & Classical Approach                   & Parcel                     \\
        \cite{malyshevArtificialNeuralNetwork2021}       & Concept for damaged parcel detection with CNNs                                      & Damage and Tampering Detection        &                      &                &                                      & Parcel                     \\
        \cite{naumannParcel3DShapeReconstruction2023}    & Damage assessment using single image 3D reconstruction                              & Damage and Tampering Detection        & RGB, Real, Synthetic &                & Deep Learning                        & Parcel                     \\
        \cite{dorrFullyAutomatedPackagingStructure2020}  & Localization of pallets and the analysis of their composition                       & Object Detection, Verify Completeness & RGB, Real            &                & Deep Learning                        & Pallet, Small Load Carrier \\
        \cite{dorrTetraPackNetFourCornerBasedObject2021} & Localization of pallets and the analysis of their composition                       & Object Detection, Verify Completeness & RGB, Real            &                & Deep Learning                        & Pallet, Small Load Carrier \\
        \cite{ozgurComparingSensorbasedCamerabased2016}  & Recognize the occupancy status of the load handling device of forklift trucks       & Verify Occupancy                      & RGB, Real            &                & Classical Approach, Fiducial Markers & Fork Lift                  \\
        \cite{liComputerVisionBased2021}                 & Recognize congestions on conveyor belts                                             & Verify Occupancy                      & RGB, Real            &                & Classical Approach                   & Conveyor Belt, Parcel      \\
        \bottomrule
    \end{tabularx}
\end{table}

\end{landscape}
The emphasis lies on automated vision-based information retrieval, and we present works on four different documentation tasks: label recognition (\cref{sec:monitoring:documentation:label_detection}), item recognition (\cref{sec:monitoring:documentation:object_detection}), tracking and tracing (\cref{sec:monitoring:documentation:tracking}), and volume estimation (\cref{sec:monitoring:documentation:volume_estimation}). 
For each section, we first describe the task and present an overview of the existing literature.
Subsequently, we summarize the findings, and briefly suggest further research directions.
Additionally, an overview of the literature reviewed in this section is presented in \cref{tab:overview:documentation}.

\subsubsection{Label Recognition}
\label{sec:monitoring:documentation:label_detection}

Transport labels uniquely identify shipments and thus, are fundamental for the organized management of goods.
Especially at every intersection point of the supply chain where goods are transferred, label detection is important to verify the completeness of the shipment.
Apart from transport labels, also other types of labels can contain relevant information that should be retrieved.
One such example are dangerous goods labels, which are crucial for the safety of operators and cargo.

\textcite{mishraDevelopmentLowcostEmbedded2019} present a low-cost embedded vision system for the detection of 1D barcodes.
They use traditional image processing to detect and rotate the barcode. 
They do not present results on the accuracy of the detections, however, focus on the execution time. %

\textcite{dorrLeanTrainingData2019} present different approaches to generate a targeted dataset for logistics transport label detection.
They take images of load carriers in realistic environments, where the load carriers have a colorful and easy-to-segment sheet of paper attached, where they would usually have a transport label.
This enables them to easily paste real transport labels onto these colorful dummy labels.
They investigate the trade-off between realism and randomness and find that accurate object detection models can also be trained on synthetic data only.
Contrary to human intuition, realism is not always advantageous, but using randomized backgrounds can yield good results.
The authors report \boxap{}s between \SI{0.65}{} and \SI{0.92}{} for different scenarios. 

\textcite{suhRobustShippingLabel2019} develop a label recognition pipeline that first detects barcodes on the shipping label, and then uses this information for angle calibration.
The angle calibration horizontally aligns the barcodes, and thus also the whole visible label. 
Afterwards, they go beyond label detection and use a second \gls*{cnn} to detect the bounding box around the address.
Finally, \gls*{ocr} is employed to extract the address as text.
The authors evaluate the barcode and address recognition accuracy and reach \SI{94.7}{\%} and \SI{93.62}{\%} respectively, while allowing random label rotation of up to \SI{20}{}$^{\circ}$. 

Focusing on maritime logistics, \textcite{shettyOpticalContainerCode2012} tackle Container \gls*{ocr}.
They focus on port logistics and present a framework composed of a container detection module, a decision engine, and a central risk management system.
The container detection module comprises an OCR module for container code retrieval which can be fused with \gls*{rfid} information of cranes and other equipment to increase robustness.
The paper suggests a framework and does not present empirical results or concrete implementations for the modules.

\textcite{brylkaCameraBasedBarcode2020} revise the problem of detecting barcodes in images.
In contrast to prior approaches, they tackle multiple real-world issues at the same time, such as poor illumination, noise, and motion blur.
Their approach consists of four consecutive steps: (1) localization, (2) segmentation and contour estimation, (3) orientation estimation and contour refinement, and (4) decoding with optional deblurring.
Modern \glspl*{cnn} are used, which is the reason for generating synthetic training datasets for each subtask.
They present two datasets for barcode localization, each comprising 25,000 images.
For segmentation, they generate 60,000 images, and finally, for deblurring a dataset of 300,000 images.
The evaluation is performed on a dataset of 400 real images, where each image contains several of 30 different barcodes.
The dataset is manually annotated and named AWB dataset.
They report a recall of \SI{0.446} and a precision of \SI{1.0} for the full approach.

\textcite{brylkaAIbasedRecognitionDangerous2021} also tackle the problem of label detection, however, since their focus is on dangerous good, we review their approach in \cref{sec:monitoring:verification:guidelines_requirements} on verifying guidelines and requirements.

\textcite{kamnardsiri1DBarcodeDetection2022} perform a case study by analyzing five different \gls*{nn} architectures for 1D barcode detection. 
They present two new datasets: InventBar and ParcelBar with 527 and 844 images, respectively.
Additionally, the evaluation considers several existing datasets.
Results show that YOLOv5 \cite{jocherUltralyticsYolov5V72022} performs best with an \ap{} of \SI{91.3}{} while YOLOx \cite{geYOLOXExceedingYOLO2021} is the fastest model considering the average runtime for all experiments.

\secsummary{}
Label recognition mostly focuses on barcodes \cite{mishraDevelopmentLowcostEmbedded2019,suhRobustShippingLabel2019,brylkaCameraBasedBarcode2020,kamnardsiri1DBarcodeDetection2022}, where also complex environments have been investigated.
For a more in-depth literature review of barcode detection, we refer to the survey of \textcite{wudhikarnDeepLearningBarcode2022}.
In addition, the detection of dangerous goods labels \cite{brylkaAIbasedRecognitionDangerous2021}, transport label detection \cite{dorrLeanTrainingData2019} and container \gls*{ocr} \cite{shettyOpticalContainerCode2012} have been tackled.

\secoutlook{}
Label detection requires object detection or semantic segmentation algorithms.
Both fields are very active areas of research \cite{zouObjectDetection202023,minaeeImageSegmentationUsing2022}.
Advances in these areas can be leveraged to improve accuracy, train with less data and increase robustness in difficult scenarios.
Especially barcode detection has been studied thoroughly and numerous datasets are publicly available. 
While \textcite{kamnardsiri1DBarcodeDetection2022} performed an analysis for a selection of algorithms, it would be interesting to analyze more diverse scenarios similar to \textcite{brylkaCameraBasedBarcode2020}.
Other fields, apart from barcode detection, lack the availability of diverse datasets and the effective use of synthetic data can be investigated.

\subsubsection{Item Recognition}
\label{sec:monitoring:documentation:object_detection}
We use item recognition in this context to refer to localizing and classifying relevant logistics objects or items in an image.
Computer vision taxonomy distinguishes object detection (\cf \textcite{zhaoObjectDetectionDeep2019} and \textcite{zouObjectDetection202023}) and semantic segmentation \cite{minaeeImageSegmentationUsing2022}, however, in our application-oriented context no such distinction is made.
Recognizing items is very helpful for documentation, \eg to identify or count incoming or outgoing goods.

\textcite{mayershoferLOCOLogisticsObjects2020} present the LOCO dataset which consists of 39,101 images in logistics environments, of which 5,593 images are annotated with bounding boxes.
Annotations are performed manually for five logistics-specific object categories: small load carrier, pallet, stillage, forklift and pallet truck.
The annotations are very unbalanced, since there are roughly 120,000 annotations for the class pallet, however less than 25,000 annotations for the second most frequent class small load carrier.
More specifically, the super-category load carrier, which includes pallet, small load carrier and stillage has 43 times more annotations than the super-category transportation vehicles, which includes pallet truck and forklift.
They report a \boxap{50} of $20.2$ using a \resnetfpn{} \cite{heDeepResidualLearning2016,linFeaturePyramidNetworks2017} when fine-tuning on LOCO.
Extensions to the dataset are planned, which include annotating more objects, incorporating 3D data and also providing segmentation masks.

Another logistics-specific dataset was presented by \textcite{mayershoferFullySyntheticTrainingIndustrial2020}.
The dataset contains synthetic training data and an industrial evaluation dataset that comprises \SI{1460}{} manually annotated images and focuses on five different types of small load carriers, as they are standardized by the VDA\footnote{Verband der Automobilindustrie e.V. (VDA), see \href{https://www.vda.de/en}{https://www.vda.de/en}.}.
The synthetic training data is created by choosing a random image as floor and dropping distractor objects, as well as the objects of interest onto this floor in a physics-based simulation.
Blender\footnote{See \href{https://www.blender.org}{https://www.blender.org}} is used, and since small load carriers are often stacked also the relation between them is modeled.
The synthetic dataset enables application for the real-world use-case of small load carrier detection, however, the detection quality is lower due to the domain gap.
The authors train Yolov3 \cite{redmonYOLOv3IncrementalImprovement2018} report a recall of $0.42$ and a precision of $0.4$ at an \gls{iou} of $50\%$.

\textcite{naumannRefinedPlaneSegmentation2020} tackle the problem of parcel segmentation and focus on the segmentation of all its side surfaces.
Additionally having plane-level segmentation information facilitates the comparison of parcel photos that were taken from different angles (\eg for tampering detection) since it allows normalizing each side surface view by applying a projective transformation.
In contrast to the other approaches, no task-specific dataset is needed.
Their method combines modern approaches for plane segmentation \cite{liuPlaneRCNN3DPlane2019} that were trained on indoor room data with approaches for contour detection \cite{cannyComputationalApproachEdge1986,soriaDenseExtremeInception2020}.
In this way, the approach is conditioned to focus on the geometry information and is less influenced by appearance changes, \eg through different parcel colors.
The authors report an average \gls{iou} over all segmentation masks of $85.2$.

While the previous works only used RGB images, \textcite{fontanaComparativeAssessmentParcel2021} present an approach for parcel detection based on \rgbd{} data.
They compare an approach that combines a \maskrcnn{} \cite{heMaskRCNN2017} with post-processing to a clustering approach on the depth data.
The clustering assumes prior knowledge of the box sizes, which limits its generalizability.
Both approaches show similar performance (errors of around \SI{5}{mm} and \SI{1}{\degree}), while the learned method is slightly better.

\textcite{naumannScrapeCutPasteLearn2022} present work on automated instance segmentation dataset generation.
They present a case study for parcels and automatically scrape relevant image data from the internet.
In the first step, only images with a homogenous background are kept and a class agnostics background removal technique is applied.
Afterwards, three different image selection processes are analyzed that are based on mask convexity, \gls*{cnn} inference, and manual selection.
Finally, the dataset is created by randomly pasting objects of interest together with distractor objects onto random backgrounds, similar to \textcite{dwibediCutPasteLearn2017}.
Results show that for the case study, manual selection of relevant parcel instances is not superior to simple pre-processing based on mask convexity (\maskap{} $81.5$ vs $86.2$).

\secsummary{}
Item recognition has been investigated for parcels \cite{naumannRefinedPlaneSegmentation2020,fontanaComparativeAssessmentParcel2021,naumannScrapeCutPasteLearn2022}, small load carriers \cite{mayershoferFullySyntheticTrainingIndustrial2020,dorrTetraPackNetFourCornerBasedObject2021} and general logistics objects such as pallets and forklifts \cite{mayershoferLOCOLogisticsObjects2020}.
Note, that other approaches also rely on object detection as part of their pipeline, however, this section focused on research where object detection is the main point of interest.

\secoutlook{}
Similar to \cref{sec:monitoring:documentation:label_detection} the computer vision tasks object detection and instance segmentation are relevant.
Both fields are very active areas of research \cite{zouObjectDetection202023,minaeeImageSegmentationUsing2022}, and advances can enable improvements in accuracy, training with less data and increasing robustness in difficult scenarios.
In general, sufficient and high-quality data frequently is a limiting factor, which leaves an opportunity for contributions.
To improve the transfer from synthetic training data to real-world applications, generative approaches such as Generative Adversarial Networks \cite{goodfellowGenerativeAdversarialNets2014} and diffusion models \cite{rombachHighResolutionImageSynthesis2022} can be used.

\subsubsection{Tracking and Tracing}
\label{sec:monitoring:documentation:tracking}

Logistics objects move along supply chains that usually comprise several parts.
Thus, in order to analyze the trajectory of specific items, it is necessary to be able to track them across these stages.
Tracking can refer to internal tracking (\eg within a facility) or global tracking (\eg across companies and facilities).
Especially for dangerous goods or in case an item gets lost, the information on the prior trajectory of an item is crucial to identify its whereabouts and thus, to guarantee operator safety and customer satisfaction.

\textcite{weichertMarkerbasedTrackingSupport2010} consider the continuous detection, localization, and identification of parcels and bins in logistics processes.
They consider roller and conveyor belt systems as typical representatives for automated transportation systems and suggest moving away from sensors such as light barriers and barcode readers and substituting them with low-cost cameras and \gls*{rfid} systems.
The combination of low-cost cameras and \gls*{rfid} systems allows one to identify an item either by detecting the marker with image processing or by reading out the data stored on the \gls*{rfid} tag.
The authors present case studies on the influence of the camera position, camera type, marker type and object distance on the marker detection accuracy.
They find a close camera position and high image resolution beneficial, while there is no clear winner for the marker type.

\textcite{borstellPalletMonitoringSystem2014} present a system for pallet monitoring.
They use a heterogeneous sensor setup and a system architecture with subsystems for pallet identification, pallet dimensioning, vehicle positioning and load change detection.

\textcite{clausenParcelTrackingDetection2019} present an industry-scale approach for tracking parcels on conveyor belts in a logistics facility.
They use a Siamese network \cite{bromleySignatureVerificationUsing1993} for parcel re-identification, to which they add a fully connected network with only one layer consisting of 1024 neurons.
A manually labeled dataset of 3,306 images from 37 different cameras, which contain a total of 14,248 parcels was created.
In addition to that, they present a calibration approach. %
Instead of manually calibrating the multi-camera framework, a single drive-by of a calibration parcel is enough for each conveyor belt.
They present extensive evaluations, also comparing to classical tracking approaches and show the superiority of their approach.
Currently, around 81\% of parcels are tracked correctly, while they name human interaction as a main cause of failure.

\textcite{nocetiMulticameraSystemDamage2018} also tackle the problem of tracking and tracing, however, since their focus is on damage and tampering detection, we review their approach in \cref{sec:monitoring:verification:damage_tampering}.

\textcite{huCuboidDetectionTracking2021} address the problem of tracking parcels inside a moving truck.
A multi-\rgbd{} camera setup is used to overcome the limited field of view and occlusions of a single camera setup.
They present a new calibration procedure using a sphere.
After calibration, their approach first creates a unified scene by merging the available \rgbd{} information.
On the resulting pointcloud, an approach for segmenting rectangular planes is applied.
These planes are used to find and evaluate box candidates.
To track a parcel, the position of its centroid, its size, and its rotation relative to the reference coordinate system is used.
They collect their own dataset with cuboids in different arrangements and report an average detection rate of \SI{0.976} while enabling real-time usage.

\textcite{rutinowskiReIdentificationWarehousingEntities2021} tackle the problem of re-identification for chipwood pallet blocks of Euro pallets.
For this purpose, they create a dataset consisting of 502 different pallet blocks and a total of 5,020 images.
The authors compare different architectures for the task and find a competitive algorithm for person re-identification \cite{sunPartModelsPerson2018} as the most suitable.
They report an \ap{} of \SI{98}{\%} and a matching accuracy of \SI{97}{\%}.

\textcite{kluttermannGraphRepresentationBased2022} use the dataset from \cite{rutinowskiReIdentificationWarehousingEntities2021} and present first results using anomaly-based re-identification of pallet blocks.
The authors identify anomalies by computing descriptive statistics of $16 \times 16$ image patches.
Detected anomalies are combined into a graph, thus, reducing the size of pallet block representation.
These graphs are processed by a Siamese \gls*{gnn} to map them into an embedding space.
Finally, in order to retrieve the matching pallet block for a given input, its nearest neighbor in the embedding space is used.
The accuracy of the approach is currently not competitive and is reported with \SI{27}{\%}.

\textcite{rutinowskiDeepLearningBased2022} also tackle the problem of re-identification for Euro pallets.
They present a new dataset consisting of 32,965 pallet blocks.
Of each pallet block four images are taken, resulting in a total of 131,860 images.
The dataset was generated automatically by monitoring conveyor belts in the warehouses of two German companies.
Similar to \cite{rutinowskiReIdentificationWarehousingEntities2021}, they apply a Part-based Convolutional Baseline (PCB) network \cite{sunPartModelsPerson2018} and report an accuracy of \SI{98}{\%}.

\secsummary{}
Tracking and tracing have been investigated for 
parcels \cite{weichertMarkerbasedTrackingSupport2010,nocetiMulticameraSystemDamage2018,clausenParcelTrackingDetection2019,huCuboidDetectionTracking2021},
and pallets \cite{borstellPalletMonitoringSystem2014,rutinowskiReIdentificationWarehousingEntities2021,rutinowskiDeepLearningBased2022, kluttermannGraphRepresentationBased2022}.
Approaches are mostly vision-based, partially rely on \rgbd{} imagery \cite{huCuboidDetectionTracking2021} and leverage literature from person re-identification \cite{rutinowskiDeepLearningBased2022}.
Moreover, whole systems for automating such processes \cite{borstellPalletMonitoringSystem2014, rutinowskijeromePotentialDeepLearning2022} and the monitoring of logistics road traffic \cite{benslimaneClassifyingLogisticVehicles2019} have been investigated.

\secoutlook{}
For the re-identification of parcels the approach of \textcite{ruiGeometryConstrainedCarRecognition2020} seems promising, as it is tailored towards cuboid-shaped objects.
Moreover, advances in feature matching \cite{sarlinSuperGlueLearningFeature2020} could be leveraged for re-identification.
For a review on re-identification, we refer to \textcite{khanSurveyAdvancesVisionbased2019} and \textcite{yeDeepLearningPerson2022}.

\subsubsection{Volume Estimation}
\label{sec:monitoring:documentation:volume_estimation}

Volume estimation refers to computing the volume of a single item or a set of items.
It is especially relevant to extract this information for optimizing downstream tasks, such as container loading.

DHL developed two prototyped solutions for volume scanning using two \kinect{} cameras in \citeyear{kuckelhausDHLLowcostSensor2013} \cite{kuckelhausDHLLowcostSensor2013,borstellProzessintegrierteVolumenerfassungLogistischen2013}.
The first solution is a gate approach while the second mounts cameras directly on a \forklift{} mast.
The volume is estimated by confining the overall pointcloud to a dedicated area and summing over a discretized height map.
Their system only needs \SI{250}{ms} for the capturing process to minimize the idle time during measurement.
Furthermore, they calibrate their system to achieve more accurate results.
They discretize the floor plane into tiles and estimate the height for each of them.
The final volume can then easily be computed by summing up over all tiles.
They analyze and compare different configurations that vary \wrt the relevant area and the camera setup.
The accuracy of their extent estimation ranges between \SI{10}{} and \SI{13}{mm} for the considered scenarios.

\textcite{laotrakunchaiMeasurementSizeDistance2013} present an approach for estimating the volume of a single parcel using a mobile device.
Their approach utilizes two different data modalities.
They use the cell phone acceleration sensor to measure the parcel extents by dragging the cell phone along its dimensions.
This information is complemented with two images (start and end of dragging) to enable measuring parcel extends from a distance.
For each image, the object region is selected manually and SURF \cite{baySpeededUpRobustFeatures2008} keypoints and descriptors are used for feature matching and subsequent disparity computation.
The final result is retrieved by applying a Gaussian weighted interpolation scheme.
Four different datasets are collected and the performance is analyzed for different object distances.
The average percentage error at distances of \SI{1}{m}, \SI{1.5}{m}, \SI{2}{m}, \SI{2.5}{m} and \SI{3}{m} is \SI{10.2}{\%}, while no clear trend on the dependence of the distance is present.

In addition to the detection of dangerous labels as will be presented in \cref{sec:monitoring:verification:guidelines_requirements}, \textcite{brylkaAIbasedRecognitionDangerous2021} also treat the problem of volume estimation.
They generate a dataset by using a setup with multiple depth cameras and fiducial markers.
This way, they automatically annotate 150 pallets with parcels that are always brown boxes and have 10 different dimensions.
They train BoNet \cite{yangLearningObjectBounding2019} and evaluate it on a separate validation dataset.
No details on the validation dataset or quantitative results are given.
The presented qualitative results look promising.

\textcite{kucukDevelopmentDimensionsMeasurement2019} develop a system for dimension estimation of static objects for logistic applications.
An \rgbd{} camera is employed to capture a pointcloud of the object.
Without going into details, the authors name spatial and temporal filters as post-processing of the pointcloud and do not report on the method used for finding the minimum bounding box for the object.
They perform evaluations on a range of objects, such as cylinders, tubes and cubes and report an error of less than \SI{0.5}{cm} in each dimension under good lightning conditions and a suitable exposure time.

Further specialized approaches include \cite{shvartsBulkMaterialVolume2014}, which heavily relies on manual input and \cite{sunObjectRecognitionVolume2020}, which only evaluates on four different parcels.

\secsummary{}
Volume estimation is mostly tackled using one \cite{kucukDevelopmentDimensionsMeasurement2019} or multiple \cite{kuckelhausDHLLowcostSensor2013,borstellProzessintegrierteVolumenerfassungLogistischen2013,brylkaAIbasedRecognitionDangerous2021} \rgbd{} cameras.
In addition to that, \textcite{laotrakunchaiMeasurementSizeDistance2013} investigated the usage of cell phones leveraging their acceleration sensor.

\secoutlook{}
Since \rgbd{} data is used frequently, studying the capabilities of pointcloud classification and pointcloud segmentation algorithms, as reviewed by \textcite{grilliReviewPointClouds2017} and \textcite{belloReviewDeepLearning2020}, seems promising.
New applications that, to the best of the authors' knowledge, have only been investigated commercially, can be considered.
Examples include measuring the load volume on a driving forklift.
For applications in scenarios with limited sensor availability, \eg during last-mile delivery, also approaches for single RGB shape reconstruction and volume estimation are interesting.
Related methodological literature is reviewed in \cite{khanThreeDimensionalReconstructionSingle2022}.
One very promising approach which has been used for shape reconstruction \cite{naumannParcel3DShapeReconstruction2023} is \cubercnn{} \cite{brazilOmni3DLargeBenchmark2023}.
By providing suitable data during training, it can also estimate scale and thus, volumes independent of otherwise necessary scale landmarks.

\subsection{Verification}
\label{sec:monitoring:verification}

Verification, in contrast to documentation, does not only encompass the mere retrieval of information but at the same time compares it to existing data.
In the following, we analyze approaches for checking completeness and occupancy (\cref{sec:monitoring:verification:completeness_occupancy}), checking guidelines and requirements (\cref{sec:monitoring:verification:guidelines_requirements}), detecting damage and tampering (\cref{sec:monitoring:verification:damage_tampering}) and finally, document analysis (\cref{sec:monitoring:verification:document}).
We present a full overview of the literature in \cref{tab:overview:verification}.

\subsubsection{Completeness and Occupancy}
\label{sec:monitoring:verification:completeness_occupancy}

Checking for completeness, \eg by counting the number of goods present, or retrieving the occupancy status of transportation containers and areas can play an important role to improve and speed up processes in logistics.

\textcite{liUsingKinectMonitoring2012} tackle the problem of monitoring warehouse order picking.
They utilize a \kinect{} to detect the picked items and check if any picking errors occur.
They restrict themselves to static box-shaped objects placed in a static basket and use 2D texture information as well as 3D geometric information to match recognized items to a database.
More precisely, they use SCARF \cite{thomasRealtimeRobustImage2011} descriptors and combine them with a volume estimation.
The evaluation of the approach yields a recognition accuracy close to \SI{100}{\%} for most of the eight objects that were tested in this limited scenario.

\textcite{ozgurComparingSensorbasedCamerabased2016} compare two approaches to recognize the occupancy status of the load handling device of forklift trucks.
One approach is sensor-based where an ultrasonic distance sensor is mounted onto the fork mast.
A pallet is recognized by monitoring the measured distance.
The second approach is camera-based, where the camera is mounted onto the ceiling to have a physically stable environment.
Fiducial markers are used to recognize the forklift and a color pattern is applied to the fork.
Finally, training data is gathered and a \gls*{svm} is trained.
The authors do not present quantitative results, however, mention that the sensor-based approach is superior since the configuration effort and the cost for installation and maintenance are lower while the accuracy is higher.

\textcite{dorrFullyAutomatedPackagingStructure2020} develop a system for automated packaging structure recognition, where the goal is the localization of uniformly packed pallets and the analysis of their composition.
They use a multi-step process: pallets are detected and for each pallet, the side faces are segmented.
The side face segmentation is rectified using a projective transformation.
On the rectified side face, a \gls*{cnn} is used to count the number of parcels and finally, the full packaging structure is determined.
The training and evaluation dataset contains a total of 1267 images.
The inter-unit segmentation, \idest the segmentation of the pallets is reported with \SI{0.9943}{} precision and a recall of \SI{1}{}.
The mean image error that takes all instances in the image into account is \SI{0.1564}{}.

In follow-up work, \textcite{dorrTetraPackNetFourCornerBasedObject2021} present a novel approach for the side face detection problem.
They extend CornerNet \cite{lawCornerNetDetectingObjects2018} to support the detection of arbitrary four-cornered polygons, instead of axis-aligned bounding boxes.
Their new model TetraPackNet shows significant improvements over a \maskrcnn{} on the dataset presented in \cite{dorrFullyAutomatedPackagingStructure2020}.
More precisely, the \maskap{} increases from \SI{58.7}{} to \SI{75.5}{}.

Finally, \textcite{liComputerVisionBased2021} present an approach to recognize congestions on conveyor belts. 
They use pre-processing steps in order to normalize the image of the observed area and subsequently employ edge detection techniques. 
They separate moving edges from static edges and use statistical information on static edges to make a prediction. 
The evaluation is performed on $160,000$ videos that were manually labeled, and they report \gls*{rocauc} of $0.9999$, which outperforms deep learning baselines they compare against.

\secsummary{}
Occupancy has been analyzed for forklift masts \cite{ozgurComparingSensorbasedCamerabased2016}  and conveyor belts \cite{liComputerVisionBased2021}.
Completeness checks have been investigated for pallets \cite{dorrFullyAutomatedPackagingStructure2020, dorrTetraPackNetFourCornerBasedObject2021} and order picking \cite{liUsingKinectMonitoring2012}.

\secoutlook{}
As the applicable computer vision techniques resemble those from \cref{sec:monitoring:documentation:object_detection}, we refer to the respective section.
Further applications include monitoring and verification of the complete packaging process of a pallet or truck.
By following the whole procedure, information is processed sequentially, which alleviates the common issue of occlusion. %
Also, counting the number of pallets within a truck or in a loading area is a relevant application.

\subsubsection{Guidelines and Requirements}
\label{sec:monitoring:verification:guidelines_requirements}

There are several guidelines and requirements when transporting dangerous goods (\cf \cite{unitednationsRecommendationsTransportDangerous2019}).
These guidelines help to protect the personnel handling the goods and the freight itself if they are recognized and followed carefully.

\textcite{brylkaAIbasedRecognitionDangerous2021} work on identifying dangerous goods by detecting the respective labels on parcels.
They generate a dataset of 1,000 manually labeled images and 50,000 synthetically generated images, which can be used for barcode and label detection.
The evaluation for barcodes is performed on a dataset, which has over 400 images with 840 barcode instances.
They improve upon \textcite{brylkaCameraBasedBarcode2020} by \SI{5}{\%} in recall.
For the detection of dangerous goods labels a new validation dataset is created by manually labeling 2,260 images with a total of 5,820 labels.
Since they consider passing under a camera arch, which yields a sequence of images, they report the detection rate per sequence.
The highest observed detection rate per sequence is \SI{96.2}{\%} at a recall of \SI{0.385}{} and a precision of \SI{0.976}{}.
Note, that they also tackle volume estimation.
The respective approach and results are reported in \cref{sec:monitoring:documentation:volume_estimation}.

\secsummary{}
Literature on verifying guidelines and requirements is very limited and focuses on classifying dangerous goods \cite{brylkaAIbasedRecognitionDangerous2021}.

\secoutlook{}
In addition to dangerous goods labels, also transportation requirements regarding orientation and maximum load can be investigated.
The former refers to checking whether a package can be transported upside down, while the latter refers to labels regarding sensible goods such as glass.
Furthermore, transportation units such as pallets might have special packaging requirements (\eg regarding the lid, foil or straps), that can be verified automatically.

\subsubsection{Damage and Tempering}
\label{sec:monitoring:verification:damage_tampering}

Logistics goods can be damaged or tampered with at any point in the supply chain and due to the steadily increasing presence of valuable goods such security considerations gain importance \cite{nocetiMulticameraSystemDamage2018}.
In order to pinpoint the time and place where such events occur, it is necessary to be able to recognize them automatically.
Damages can have several forms, such as water damage or deformation.
Also, tampering can be detectable in different ways: \eg new tape can be applied after opening a parcel or labels could be attached to or removed from a parcel.

\textcite{nocetiMulticameraSystemDamage2018} investigate damage and tampering detection in a postal security framework by extracting 3D shape and appearance information (\idest brightness patterns) from multiple cameras.
The authors present their detection method along with use-cases and a database storage of collected data for future reference, however, we will focus our attention on the vision-based detection approach.
Change detection \cite{staglianoOnlineSpaceVariantBackground2015} is used to fit parallelepipeds on binary masks of parcels.
Damage detection is then performed by comparing the obtained 3D shape with its expected shape, while tampering is based on the comparison of the parcel's side surfaces \cite{dalalHistogramsOrientedGradients2005}.
\textcite{nocetiMulticameraSystemDamage2018} report, that they reach an overall damage and tampering detection accuracy of over 90\%.

\textcite{malyshevArtificialNeuralNetwork2021} present a concept outlining the use of \glspl*{cnn} for damage detection.
They name deformation, rupture and moisture as possible categories of damage to a package.
Also examples of damages, which do not imply damage to the cargo are visualized exemplarily.

\textcite{naumannParcel3DShapeReconstruction2023} present a novel architecture for 3D shape reconstruction of cuboid-shaped and damaged parcels from single RGB images.
They combine estimating a 3D bounding box \cite{brazilOmni3DLargeBenchmark2023} with an iterative mesh refinement branch \cite{wangPixel2MeshGenerating3D2018}, to leverage the strong prior in form of the 3D bounding box while at the same time being able to adjust to damaged parcels.
Thus, their approach estimates the 3D mesh of the current, potentially deformed parcel shape, as well as its original pristine shape.
This enables not only damage classification but also damage quantification by directly comparing 3D meshes.
Training and evaluation are performed on \dsparcel{}, a novel synthetic dataset of intact and damaged parcels in diverse environments.
Their architecture \cuberefine{} performs best on intact parcels (\meshap{50} of $92.8$) and competitively for damaged ones (\meshap{50} of $70.7$).

\secsummary{}
Literature on damage and tampering detection is scarce and focuses on parcels \cite{nocetiMulticameraSystemDamage2018,naumannParcel3DShapeReconstruction2023}.
The most important requirement for such systems, according to the results of the questionnaire by \textcite{nocetiMulticameraSystemDamage2018}, is easy integration into existing processes.
More precisely, current processes should not be slowed down significantly and hardware installation should be easily possible within existing plants.
This, however, is challenging since the process and plant design can differ significantly within and across companies.

\secoutlook{}
Tampering detection approaches rely on generating viewpoint invariant parcel side surface representations, which can be computed from the parcel corner points.
Thus, incorporating recent advances in keypoint detection seems promising.
Moreover, prior knowledge of the cuboid shape can be leveraged by utilizing a vanishing point loss \cite{ruiGeometryConstrainedCarRecognition2020} or exploiting 2D/3D correspondences \cite{liRTM3DRealTimeMonocular2020}.
Damage pattern recognition, \idest identifying and clustering damages to recognize frequently occurring patterns is very interesting.
In addition to that, estimating the full 3D reconstruction of a parcel from \rgbd{} data has not been studied yet and would be very interesting to investigate for enabling detailed damage quantification.
Finally, there is no dataset for analyzing different cases of damages, such as water damage or ruptures in the packaging, and damages for other objects such as pallets have not been investigated yet.

\subsubsection{Document Analysis}
\label{sec:monitoring:verification:document}

Shipments are always accompanied by documents that provide additional insights such as product types, product quantities, and product prices.
Thus, to obtain detailed information about the shipment, it is also necessary to automatically process documents.
The first step towards this goal is to unwarp and rectify documents that might have been crumbled or warped during the transportation process.
This task is called document rectification and it has gained a lot of attention recently \cite{maDocUNetDocumentImage2018,xieDewarpingDocumentImage2020,markovitzCanYouRead2020,fengDocTrDocumentImage2021,xueFourierDocumentRestoration2022,dasLearningIsometricSurface2022,jiangRevisitingDocumentImage2022,wangUDocGANUnpairedDocument2022}.
Once the document is rectified, contents can be analyzed using \gls*{ocr}.
We refer to reviews for handwritten \cite{memonHandwrittenOpticalCharacter2020} and typed \cite{subramaniSurveyDeepLearning2021}  \gls*{ocr}.
Moreover, document structure recognition plays an important role 
\cite{pintoTableExtractionUsing2003,riadClassificationInformationExtraction2017,subramaniSurveyDeepLearning2021,liDiTSelfsupervisedPretraining2022}.

\secsummary{}
Approaches in the area of document analysis tackle the general problem and we are not aware of literature focusing on logistics use-cases.
Document rectification \cite{maDocUNetDocumentImage2018,xieDewarpingDocumentImage2020,markovitzCanYouRead2020,fengDocTrDocumentImage2021,xueFourierDocumentRestoration2022,dasLearningIsometricSurface2022,jiangRevisitingDocumentImage2022,wangUDocGANUnpairedDocument2022} has gained a lot of attention recently.
Moreover, \gls*{ocr} \cite{memonHandwrittenOpticalCharacter2020,subramaniSurveyDeepLearning2021} and document structure recognition \cite{pintoTableExtractionUsing2003,riadClassificationInformationExtraction2017,subramaniSurveyDeepLearning2021, liDiTSelfsupervisedPretraining2022} are important problems that are investigated in the literature.

\secoutlook{}
The above-mentioned techniques can be applied to documents that are relevant to logistics processes, such as delivery notes.
Especially detecting and interpreting human annotations on such documents seems very promising.

\section{Manipulation}
\label{sec:manipulation}

While for monitoring the focus was on information retrieval, we now focus on the interaction with the environment. 
We follow \textcite{borstellBildbasierteZustandserfassungLogistik2021} and divide this section according to the degree of automation into two parts: machine-supported tasks and fully autonomous manipulation.
The former, also called assistance, refers to providing helpful additional information to human operators and will be treated in \cref{sec:manipulation:assistance} and is summarized in \cref{tab:overview:assistance_for_manipulation}. 
The final objective is not to fully automate a process, but instead to ease the workload for operators.
Literature on fully autonomous manipulation will be presented in \cref{sec:manipulation:autonomous} and is summarized in \cref{tab:overview:autonomous_manipulation}. 
In contrast to the literature on assistance, the focus here is not on providing additional information but on helping to solve the task at hand autonomously.
Note, that literature is categorized \wrt the goal that is pursued and not the achieved degree of automation.

\subsection{Assistance for Manual Manipulation}
\label{sec:manipulation:assistance}

Nowadays, most logistics processes are still handled manually.
To ease the workload for human operators, assistance systems can be developed.
Note, that the research presented here specifically strives towards assisting a human operator, as opposed to aiming at fully automating a process.
We identified literature focusing on order picking, which is discussed in \cref{sec:manipulation:assistance:order_picking} and on the packaging process, which is presented in \cref{sec:manipulation:assistance:packaging}. 
For a general overview of all literature, we refer to \cref{tab:overview:assistance_for_manipulation}.

\subsubsection{Order Picking}
\label{sec:manipulation:assistance:order_picking}

Orders commonly comprise several items, which need to be collected for shipping.
The process of assembling an order is also referred to as order picking, and it is essential for the efficiency of warehouses.

\textcite{reifEntwicklungUndEvaluierung2009} investigates the suitability of \gls*{ar} for order picking from an operator's perspective.
During order picking, static text with product information as well as dynamic 3D information, \eg, about the source or target position, can be displayed.
The authors perform a study that shows a steep learning curve and an improvement in order picking time and error rate in comparison to order picking with a paper-based list.
Further, the subjective workload is not higher when using \gls*{ar}, while the mental workload is lower due to the provided helpful information.
Nonetheless, they recommend further longtime studies to analyze the effects and note that the learning curve depends on the individuals.

\begin{landscape}
    \begin{table}[t!]
    \centering\footnotesize
    \caption{Overview of the literature on Assistance for Manipulation (cf. \cref{sec:manipulation:assistance})}
    \label{tab:overview:assistance_for_manipulation}
    \begin{tabularx}{\linewidth}{>{\hsize=1.1\hsize\RaggedRight}X>{\hsize=1.8\hsize}X>{\hsize=1.2\hsize\RaggedRight}X>{\hsize=0.7\hsize\RaggedRight}X>{\hsize=0.3\hsize\RaggedRight}X>{\hsize=1.1\hsize\RaggedRight}X>{\hsize=.8\hsize\RaggedRight}X}
        \toprule
        Paper                                                & Summary                                                                                     & Application            & Data Type          & Public Dataset & Approach Type      & Objects         \\
        \midrule
        \cite{reifEntwicklungUndEvaluierung2009}             & Analysis of the suitability of augmented reality for order picking                          & Order Picking          &                    &                &                    &                 \\
        \cite{liUsingKinectMonitoring2012}                   & Monitoring warehouse order picking                                                          & Order Picking          & \rgbd{}, Real      &                & Classical Approach & Other Packaging \\
        \cite{grzeszickCameraassistedPickbyfeel2016}         & Assist the picking process with wearables                                                   & Order Picking          & Real               &                & Deep Learning      & Label           \\
        \cite{hochsteinPackassistentAssistenzsystemFuer2016} & Assistance system based on augmented reality for quality control during the packing process & Packaging for Shipment & RGB, \rgbd{}, Real &                &                    & Other, Parcel   \\
        \cite{maettigUntersuchungEinsatzesAugmented2016}     & Study to analyze how the packaging process can be improved by using augmented reality       & Packaging for Shipment & Real               &                &                    &                 \\
        \bottomrule
    \end{tabularx}
\end{table}
    \begin{table}[t!]
    \centering\footnotesize
    \caption{Overview of the literature on Autonomous Manipulation (cf. \cref{sec:manipulation:autonomous})}
    \label{tab:overview:autonomous_manipulation}
    \begin{tabularx}{\linewidth}{>{\hsize=1.1\hsize\RaggedRight}X>{\hsize=1.8\hsize}X>{\hsize=1.2\hsize\RaggedRight}X>{\hsize=0.7\hsize\RaggedRight}X>{\hsize=0.3\hsize\RaggedRight}X>{\hsize=1.1\hsize\RaggedRight}X>{\hsize=.8\hsize\RaggedRight}X}
        \toprule
        Paper                                              & Summary                                                                                      & Application                                         & Data Type                     & Public Dataset & Approach Type                                       & Objects               \\
        \midrule
        \cite{thamer3DrobotVisionSystem2013}               & Pointcloud segmentation for container unloading                                              & Depalletization                                     & Pointcloud, Real, Synthetic   &                & Classical Approach, Pattern Matching                & Container/ Trailer    \\
        \cite{thamer3DComputerVisionAutomation2014}        & Pointcloud segmentation for container unloading                                              & Depalletization                                     & Pointcloud, Real, Synthetic   &                & Classical Approach, Deep Learning, Pattern Matching & Container/ Trailer    \\
        \cite{prasseNewApproachesSingularization2015}      & Depalletization using a robot arm and low-cost 3D sensors                                    & Depalletization                                     & \rgbd{}, Real, Synthetic      &                & Classical Approach, Pattern Matching                & Pallet, Parcel        \\
        \cite{arpentiRGBDRecognitionLocalization2020}      & Optimized depalletization using a single \rgbd{} camera                                      & Depalletization                                     & \rgbd{}, Real                 &                & Classical Approach                                  & Pallet, Parcel        \\
        \cite{chiaravalliIntegrationMultiCameraVision2020} & Depalletizing using a robot with a fixed time-of-flight camera and an eye-in-hand RGB camera & Depalletization                                     & RGB, \rgbd{}, Real            &                & Classical Approach, Pattern Matching                & Parcel                \\
        \cite{brylkaAIbasedRecognitionDangerous2021}       & Detection of dangerous goods labels and volume estimation by pointcloud segmentation         & Depalletization, Label Detection, Volume Estimation & RGB, \rgbd{}, Real, Synthetic &                & Deep Learning                                       & Label, Pallet, Parcel \\
        \cite{vargaVisionbasedAutonomousLoad2014}          & Pallet detection and localization for an autonomous forklift using a stereo camera           & Pallet Handling                                     & RGB, Real                     &                & Classical Approach, Pattern Matching                & Pallet                \\
        \cite{vargaImprovedAutonomousLoad2015}             & Pallet detection and localization for an autonomous forklift using a stereo camera           & Pallet Handling                                     & RGB, Real                     &                & Classical Approach, Pattern Matching                & Pallet                \\
        \cite{haanpaaMachineVisionAlgorithms2016}          & Pallet engagement in the context of military logistics                                       & Pallet Handling                                     & RGB, \rgbd{}, Real            &                & Classical Approach, Fiducial Markers                & Pallet                \\
        \cite{vargaRobustPalletDetection2016}              & Pallet detection and localization for an autonomous forklift using a stereo camera           & Pallet Handling                                     & RGB, Real                     &                & Classical Approach, Pattern Matching                & Pallet                \\
        \cite{xiaoPalletRecognitionLocalization2017}       & Pallet recognition and localization by using an \rgbd{} camera                               & Pallet Handling                                     & \rgbd{}, Real                 &                & Classical Approach                                  & Pallet                \\
        \cite{molterRealtimePalletLocalization2018}        & Pallet detection and localization using a time-of-flight camera                              & Pallet Handling                                     & \rgbd{}, Real                 &                & Classical Approach, Pattern Matching                & Pallet                \\
        \cite{molterSemiAutomaticPalletPickup2019}         & Driver assistance system for a forklift truck                                                & Pallet Handling                                     & \rgbd{}, Real                 &                & Classical Approach, Pattern Matching                & Pallet                \\
        \cite{mohamedDetectionLocalisationTracking2020}    & Pallet recognition and tracking using only an onboard laser rangefinder                      & Pallet Handling                                     & Pointcloud, Real              &                & Deep Learning                                       & Pallet                \\
        \bottomrule
    \end{tabularx}
\end{table}
\end{landscape}

\textcite{grzeszickCameraassistedPickbyfeel2016} present an approach to assist the picking process with wearables.
The picker is equipped with a smartwatch and a low-cost camera.
The camera is used for activity recognition, which triggers further processing if determined that the current activity is picking an object. 
When the activity recognition recognizes a picking process an image is taken and analyzed with barcode detection and a \gls*{cnn} to check whether the correct item was picked.
The smartwatch is used to display the information relevant to the next pick and gives tactile feedback regarding the success of the pick.
The authors report $80.1\%$ accuracy for action recognition and $89\%$ for the recognition of clearly visible barcodes.

\secsummary{}
Assistance solutions for order picking focus on using \gls*{ar} \cite{reifEntwicklungUndEvaluierung2009} and wearables \cite{grzeszickCameraassistedPickbyfeel2016}.
In both cases, results suggest a steep learning curve and a positive impact on the operator's performance.

\secoutlook{}
In general, \gls*{ar} has a high potential for assistance during manual manipulation processes.
Its usefulness for the area of logistics has been analyzed by \textcite{stoltzAugmentedRealityWarehouse2017} and an industry perspective is provided by DHL \cite{glocknerAugmentedRealityLogistics2020}.
Recent plane detection algorithms \cite{liuPlaneRCNN3DPlane2019} can help to reduce sensor requirements to enable a broader acceptance of \gls*{ar} techniques.
Furthermore, it would be interesting to investigate the suitability of \gls*{ar} interfaces for different applications, since for example glasses or a projector might not be feasible for some use-cases.
Finally, research on the utility of wearables in different logistics scenarios is an interesting topic for research.

\subsubsection{Packaging for Shipment}
\label{sec:manipulation:assistance:packaging}

Once all items of an order have been assembled, they need to be stored safely and efficiently inside a transport unit, such as a parcel.
This process is crucial for customer satisfaction since damaged items are a frequent cause for complaints.

\textcite{hochsteinPackassistentAssistenzsystemFuer2016} develop an assistance system based on \gls*{ar} for quality control during the packing process.
They use one RGB camera and one \gls*{tofc}.
Commercial software for object detection is used for monitoring the packing process.
\textcite{hochsteinPackassistentAssistenzsystemFuer2016} do not recognize all articles that are placed into a parcel but focus on the localization within the box.
As assistance, the packing list and other relevant information are projected onto the workplace.
The authors constructed a working prototype focusing on ergonomics and privacy, however, no user study was conducted.

\textcite{maettigUntersuchungEinsatzesAugmented2016} perform a study to analyze how the packaging process can be improved by using \gls*{ar}.
They name pressure on time, quality, and costs as potential areas of improvement enabled by the usage of \gls*{ar} for packaging.
They perform a study with two groups of 10 people and the results confirm the hypothesis that \gls*{ar} helps to improve time efficiency when packing order suggestions are displayed.
In addition, it helps optimizing costs, \idest identifying the best parcel size.

\secsummary{}
Literature on assistance during the packaging process focuses on the usage of \gls*{ar} techniques \cite{hochsteinPackassistentAssistenzsystemFuer2016,maettigUntersuchungEinsatzesAugmented2016}.
Overlaying digital information with the visual perception of the world can help to improve efficiency and reduce errors during the packaging process.

\secoutlook{}
Since the same techniques as mentioned in \cref{sec:manipulation:assistance:order_picking} are relevant, we refer to the respective outlook.
In addition to that, further in-depth studies of the performance and acceptance of such techniques are promising research directions.

\subsection{Autonomous Manipulation}
\label{sec:manipulation:autonomous}

While most tasks are still handed by humans or by \gls*{hci}, research is striving for fully autonomous solutions. 
In this section, we present research that works towards this goal.
Since we prioritize the long-term objective of automation, works are presented independent of the current level of automation that they achieve. 
First, literature on pallet handling is reviewed in \cref{sec:manipulation:autonomous:pallet}, and subsequently literature on depalletization in \cref{sec:manipulation:autonomous:depalletization}.
Afterwards, we present insights into logistics-related approaches for pick-and-place in \cref{sec:manipulation:autonomous:pick} and \glspl*{agv} in \cref{sec:manipulation:autonomous:agv} briefly.
Moreover, we provide a general overview of the literature in \cref{tab:overview:autonomous_manipulation}.

\subsubsection{Pallet Handling}
\label{sec:manipulation:autonomous:pallet}
Pallet handling refers to the task of automatically recognizing, localizing and interacting with pallets.
Due to the ubiquity of pallets in the context of material handling, this is a crucial task that is frequently needed in logistics processes.

\textcite{vargaVisionbasedAutonomousLoad2014} present an approach for pallet detection and localization for \glspl*{agv} using a stereo camera.
During the considered loading and unloading operations, the \gls*{agv} is assumed to stop at a distance of approximately \SI{2.5}{m} from the pallet.
To engage with the pallet, the 3D position must be provided accurately.
Accuracy is defined as a deviation of at most \SI{5}{cm} and \SI{1}{\degree}.
The presented pipeline consists of first performing stereo image rectification and stereo matching.
Subsequently, pallet detection is applied to the left image.
This result is then employed for exterior reconstruction and plane fitting.
They define a model for a pallet, which consists of three legs that are separated by empty pockets.
The relative proportions of each area are assumed to be known and constant.
They use a sliding window approach and design features that represent the assumed pallet model.
These features are used to train an AdaBoost \cite{freundDesiciontheoreticGeneralizationOnline1995} classifier.
The datasets for training and testing were gathered manually for this work.
The detection rate for weak positives is reported as \SI{94}{\%}, the one for strong positives as \SI{84}{\%}, while having a false positive rate of \SI{1.5}{\%}.

\textcite{vargaImprovedAutonomousLoad2015} follow up on their work \cite{vargaVisionbasedAutonomousLoad2014} and again use a stereo camera system to detect pallets for an autonomous forklift.
Their approach is based on a fixed-size sliding window, where multiple scales of the image are used as input and a fixed aspect ratio of the pallet is assumed.
They compute eight image channels based on the camera input: a grayscale image, the gradient magnitude and the oriented gradient magnitude at six different orientations.
A random forest (Adaboost \cite{freundDesiciontheoreticGeneralizationOnline1995}) is used to train a classifier.
The training dataset Viano2 contains 7,124 images with 9,047 pallets and the test dataset Viano3-5 contains 467 images with 891 pallets.
Their best model achieves \SI{78.1}{\%} precise matches on the test set.
The authors also perform a test, during which all operations were successful after fine-tuning for the new scenario.
They name illumination as an open challenge, which is tackled in their subsequent work \cite{vargaRobustPalletDetection2016}.

\textcite{haanpaaMachineVisionAlgorithms2016} focus on the problem of pallet engagement in the context of military logistics.
A multi-sensor setup is used that comprises two \glspl*{tofc} and an RGB camera.
They consider a pipeline consisting of different steps and vary the type of input information.
In addition to that, they resort to fiduciary markers for certain tasks.
The authors do not provide a quantitative analysis of their results.

\textcite{xiaoPalletRecognitionLocalization2017} present an approach for pallet recognition and localization by using an \rgbd{} camera.
A region-growing algorithm \cite{xiaoThreedimensionalPointCloud2013} is used to segment the depth image into planar patches.
A heuristic for pre-processing is employed, and afterwards, pattern matching on the remaining segments is applied.
The pattern matching focuses on the pallet only and a set of five different pallet base models is used.
The pattern matching does not rely on any color information but instead works on a binary image that belongs to a planar patch.
The authors present qualitative data to exemplary show how well the approach works, however, no quantitative results are presented.
As failure cases, they mention that the pallet could be too dark or that items obscuring the pallet can change the observed pallet pattern.

\textcite{molterRealtimePalletLocalization2018} present an approach for pallet detection and localization using a \gls*{tofc}. 
The camera is mounted on top of the back side of the fork of a forklift. 
They remove the ground plane within the pointcloud using a RANSAC \cite{fischlerRandomSampleConsensus1987} approach. 
Subsequently, a region-growing algorithm is used to determine surface clusters. 
For each cluster, the centroid and normal vector are calculated, and filtering for vertical planes is applied. 
Afterwards, centroid pairs and triple pairs are computed and finally, a check using the geometrical information on the pallet is performed to obtain a candidate. 
The authors report, that in the static scenario, almost all pallets were detected correctly, while in the dynamic one only \SI{55}{\%} were localized correctly.
In a follow-up work, \textcite{molterSemiAutomaticPalletPickup2019} present an advanced driver assistance system for forklifts operated by humans. 
The driver assistance system utilizes the pallet detection approach \cite{molterRealtimePalletLocalization2018} and combines it with trajectory planning and control for the forklift, as well as a user interface inside the forklift.

\textcite{mohamedDetectionLocalisationTracking2020} present an approach for pallet recognition and tracking using only the onboard laser rangefinder of a forklift.
The data from the laser scanner is used to create top views of the occupancy of the floor.
These top-view images are then processed by a \fasterrcnn{} \cite{renFasterRCNNRealTime2017}.
To enable robust localization and tracking over time a Kalman filter \cite{kalmanNewApproachLinear1960} is used.
The authors evaluate on a real-world dataset containing 340 labeled top-view images, which is augmented by rotation and displacement to 1020 images.
\SI{714}{} samples are used for training and \SI{306}{} are used for testing.
An average accuracy of \SI{99.58}{\%} is reported.

\secsummary{}
The problem of pallet localization has been tackled frequently in literature.
Most approaches employ classical computer vision techniques and frequently pattern matching is part of the pipeline \cite{vargaVisionbasedAutonomousLoad2014,vargaImprovedAutonomousLoad2015,vargaRobustPalletDetection2016,haanpaaMachineVisionAlgorithms2016,xiaoPalletRecognitionLocalization2017, molterRealtimePalletLocalization2018}.
Recently, deep learning has been used to identify pallets from laser rangefinder data \cite{mohamedDetectionLocalisationTracking2020}.
For details on earlier approaches, we refer to the literature review on the pallet loading problem by \textcite{vargas-osorioLiteratureReviewPallet2016} and to \textcite{mohamedDetectionLocalisationTracking2020}.

\secoutlook{}
For pallet engagement, the usage of recent pointcloud classification and segmentation algorithms (cf \cite{grilliReviewPointClouds2017,belloReviewDeepLearning2020}) has not yet been investigated.
Due to the high accuracy of \sota{} 2D object detection algorithms \cite{xieSegFormerSimpleEfficient2021} they are very well-suited for pallet detection, provided a sufficiently large dataset is available.

\subsubsection{Depalletization}
\label{sec:manipulation:autonomous:depalletization}

In warehouses, hand-held scanners are still very common to verify that a package is on the correct pallet and to do inventory.
Also the subsequent step of depalletization involves a lot of human labor and can still be slow and error-prone.

\textcite{thamer3DrobotVisionSystem2013} develop a segmentation technique for pointclouds in logistics applications.
More precisely, they present a system for the automatic unloading of containers.
With the knowledge of the spatial relationship between the sensor and the container, they filter out the background in a first step.
The pointcloud now only contains objects of interest and needs to be segmented.
They pursue a two-step approach inspired by \cite{schnabelShapeRecognition3D2007,scholz-reiterApproach3DObject2011} and start with applying the graph-based segmentation technique by \textcite{felzenszwalbEfficientGraphBasedImage2004}.
To improve on these results, they additionally implement an iterative process for region-growing.
Subsequently, \textcite{thamer3DrobotVisionSystem2013} fit surface patches onto the segments of the pointcloud \cite{moreLevenbergMarquardtAlgorithmImplementation1978}.
In order to find objects within the surface patches, they define models for boxes, cylinders and sacks which they try to fit.
The fitting process consists of a directed graph, where the surface patches are the vertices and the edges save information on possible object matches.
By employing distance metrics, they find possible candidates for objects and check them against the models for their predefined shapes.
The evaluation of the approach is done with 54 different real-world packaging scenarios and further 51 synthetically generated ones.
They report better performance on the synthetic data, recognizing \SI{83}{\%} of the labeled graspable goods within the scene.
The authors argue, that the approach has high potential, however, was not capable to run in real-time in \citeyear{thamer3DrobotVisionSystem2013}.

\textcite{thamer3DComputerVisionAutomation2014} is a subsequent work, which also tackles the problem of detecting differently shaped logistics goods to enable automated processing by robotic systems.
The classes box, barrel and sack are considered and artificial training data is generated for each class.
Their approach does not operate on the points directly, but instead uses Viewpoint Feature Histograms (VFH) \cite{rusuFast3DRecognition2010}.
As pre-processing steps, points belonging to the background are removed and denoising techniques are applied.
In addition, object candidates are separated by applying clustering on their Euclidean distance.
\glspl*{svm} and \glspl*{nn} are used for the final classification and trained on 500 simulated training examples per shape class.
The authors report a classification accuracy above \SI{90}{\%} on synthetic data, while the performance on real data is significantly lower in most cases.
Moreover, \glspl*{svm} seem to perform better on synthetic data, while the \gls*{nn} seems more robust and handles real data better.

In \citeyear{prasseNewApproachesSingularization2015}, \textcite{prasseNewApproachesSingularization2015} used a robot arm and low-cost 3D sensors to perform the task of depalletization.
They present two approaches:
The first uses a \gls*{pmd} sensor and a pre-determined model of loading situations, while the second employs a 3D scanning approach by dynamically positioning a structured light sensor with the robot arm.
The loading units are assumed to have a cuboid shape and are determined by employing a \gls*{ransac} \cite{fischlerRandomSampleConsensus1987} algorithm.
\textcite{prasseNewApproachesSingularization2015} evaluate the \gls*{pmd} approach on real data and report the deviation of package dimensions for 9 parcels.
Since the deviation in height is less than their threshold of \SI{10}{mm}, they argue that the approach is suitable for application in logistics.
The second approach is evaluated on synthetic data only and the influence of the following factors on accuracy and runtime are investigated: pointcloud size, number of iterations, parcel dimensions, randomly transformed pointclouds, simulated Gaussian sensor noise, and number of detected parcel faces.

\textcite{arpentiRGBDRecognitionLocalization2020} present an approach for depalletization using a single \rgbd{} camera.
The information from the \rgbd{} camera is supplemented by a case database, which contains the product barcode, the number of boxes on a pallet, their dimensions and one image of each textured face of the box.
Since they have an existing database, image segmentation is based on matching the SIFT \cite{loweObjectRecognitionLocal1999} features of the textures.
If sufficient matches are found, the homography between the taken image and the one from the database is computed.
Because this only works for textured faces, the watershed transform \cite{meyerColorImageSegmentation1992} is used for the other cases.
The segmentation is then used, in order to determine the 3D plane of the face from the depth information and all points from the segmentation are associated with it.
The minimum area enclosing rectangle around this pointcloud is then used as input for the geometrical module.
The purpose of the geometrical module is to find matching faces for the candidate faces in the database.
This is done by matching the visible dimensions to the database information and the pose is estimated.
For the evaluation, a database of nine cases that are organized in ten different settings is considered.
The authors report an accuracy of \SI{98}{\%}.
The authors also perform a case study on the whole depalletizing process.
Here, a black tendon is placed behind the pallet to reduce reflections for the \rgbd{} camera.
One configuration was tested and all parcels from this configuration were correctly depalletized.

\textcite{chiaravalliIntegrationMultiCameraVision2020} present an approach towards depalletizing using a robot with a fixed \gls*{tofc} and an eye-in-hand RGB camera.
The sizes of the boxes as well as the plane through the top of the highest parcel are assumed to be known.
Canny edge detection \cite{cannyComputationalApproachEdge1986} and the Hough transform \cite{houghMethodMeansRecognizing1962} are used to determine the edges of the boxes.
A connectivity graph is then used to generate box hypotheses and an optimization problem is solved using a genetic algorithm to identify the relevant boxes.
Afterwards, their pipeline comprises the following steps: gap localization, gap alignment, insertion test, insertion complete, and collection.
Note, that the goal is not to grasp the parcel, but instead to pull it to a desired position.
The box detection and pose estimation is analyzed on 125 depth images.
The authors report an average standard deviation of \SI{3.05}{mm} for the box position and a position error of \SI{3.60}{mm}.
For evaluation, six boxes of equal size are positioned in three rows of two parcels each, with no initial gap.

\secsummary{}
All approaches use depth information for depalletization and rely on base patterns.
Most approaches operate on pointclouds \cite{thamer3DrobotVisionSystem2013,prasseNewApproachesSingularization2015, chiaravalliIntegrationMultiCameraVision2020}, while \cite{thamer3DComputerVisionAutomation2014} rely on Viewpoint Feature Histograms (VFH) \cite{rusuFast3DRecognition2010}.
Currently, approaches do not use \glspl*{gnn} for this task and oftentimes additional prior information is assumed.

\secoutlook{}
Similar to the case of pallet handling, the usage of pointcloud classification and segmentation algorithms (\cf \cite{grilliReviewPointClouds2017,belloReviewDeepLearning2020}) has not yet been investigated and suggests interesting future applications.
Furthermore, the approaches work towards automated pallet handling, however, to enable reliable real-world deployment higher robustness and end-to-end integration are necessary.

\subsubsection{Pick-and-Place}
\label{sec:manipulation:autonomous:pick}
The importance of pick-and-place for logistics use-cases is manifested for example in the Amazon picking challenge \cite{correllAnalysisObservationsFirst2018}.
Since the applications of this task reach far beyond logistics, we refer to existing literature reviews \cite{bicchiRoboticGraspingContact2000,baiObjectDetectionRecognition2020,kleebergerSurveyLearningBasedRobotic2020,duVisionbasedRoboticGrasping2021}.
Works with a focus on logistics include \cite{rennieDatasetImprovedRGBDBased2016, zeng2016multi,boniniFastDeployableAutonomous2016,schwarzNimbRoPickingVersatile2017,wahrmannAutonomousFlexibleRobotic2019, pavlichenkoKittingBotMobileManipulation2019}.

\subsubsection{\glspl*{agv}}
\label{sec:manipulation:autonomous:agv}
\glspl*{agv} are important in the logistics domain and beyond \cite{croninStateoftheArtReviewAutonomous2019, fottnerAutonomousSystemsIntralogistics2021}.
Key components of systems include visual SLAM \cite{miethSurveyPotentialsIndoor2019} and planning \& control \cite{fragapanePlanningControlAutonomous2021}.
Research with a particular focus on logistics includes \cite{himstedtOnlineSemanticMapping2017,kumbharAutomatedGuidedVehicles2018, sabattiniPANRobotsProjectAdvanced2018,yangCubeSLAMMonocular3D2019,zhouObjectDetectionMapping2021}.
Note, that also autonomous driving has a huge potential for logistics \cite{mullerAutomatedTrucksRoad2019}, however, is not considered as part of this review.

\section{Computer Vision Perspective}
\label{sec:other}

The goal of this section is to review the presented literature from a computer vision perspective.
We first present a categorization of the literature \wrt the computer vision tasks they tackle in \cref{sec:other:cv}. 
Subsequently, we present a brief overview of existing, publicly available datasets in \cref{sec:other:datasets} and of industrial solutions in \cref{sec:other:industry}.

\subsection{Methodological Categorization}
\label{sec:other:cv}

We briefly categorize the reviewed literature according to the computer vision task they solve. %
Approaches for 2D and 3D data are presented, and we refer to \textcite{riestockSurveyDepthCameras2019} for an overview of sensors commonly used in logistics.

\ownsec{Marker-based Detection}
Fiducial markers are commonly used to facilitate visual identification.
Those markers include barcodes \cite{mishraDevelopmentLowcostEmbedded2019}, ArUcos \cite{haanpaaMachineVisionAlgorithms2016} and other markers \cite{garibottoIndustrialExploitationComputer1998,reifEntwicklungUndEvaluierung2009,weichertMarkerbasedTrackingSupport2010}.
Using markers is still a valid and robust approach, where the overhead of marking all necessary goods is feasible.

\ownsec{Edge Detection}
Edge detection has been applied for interacting with pallets \cite{molterRealtimePalletLocalization2018} and parcels \cite{naumannRefinedPlaneSegmentation2020}.

\ownsec{Object Detection}
Several approaches \cite{vargaVisionbasedAutonomousLoad2014,vargaRobustPalletDetection2016,shettyOpticalContainerCode2012,mayershoferLOCOLogisticsObjects2020,malyshevArtificialNeuralNetwork2021} tackle the problem of object detection, \idest localizing and classifying an object in an image.
Apart from RGB images, object detection has been applied to top-view occupancy images of 2D laser range finders \cite{mohamedDetectionLocalisationTracking2020}.

\ownsec{Instance Segmentation}
Instance segmentation tackles the problem of identifying each instance of a class separately by assigning pixel-wise correspondences.
Research in the area of logistics includes \cite{shvartsBulkMaterialVolume2014,suhRobustShippingLabel2019,naumannRefinedPlaneSegmentation2020,mayershoferFullySyntheticTrainingIndustrial2020,brylkaAIbasedRecognitionDangerous2021,liComputerVisionBased2021}, where packing structure recognition \cite{dorrFullyAutomatedPackagingStructure2020,dorrTetraPackNetFourCornerBasedObject2021} exploits this information to determine the pallet composition.
Furthermore, \cite{himstedtOnlineSemanticMapping2017,sunObjectRecognitionVolume2020} use instance segmentation together with depth information.

\ownsec{Object Re-Identification and Tracking}
The same object might occur multiple times in images, \eg at a different location for a different point in time.
The goal of object re-identification is to be able to find all images in which a target object is visible.
A very popular area of application for this task is person re-identification.
The existing literature considers boxes \cite{liUsingKinectMonitoring2012}, parcels \cite{nocetiMulticameraSystemDamage2018}, and pallets \cite{rutinowskiReIdentificationWarehousingEntities2021,rutinowskiDeepLearningBased2022,kluttermannGraphRepresentationBased2022}.
Moreover, work on tracking parcels \cite{clausenParcelTrackingDetection2019} has been proposed.

\ownsec{Action Recognition}
Identifying the activity currently pursued by a person can be a very helpful task.
For example, research has been done on determining if a worker is currently picking an object, to use this information for order picking \cite{grzeszickCameraassistedPickbyfeel2016}.

\ownsec{3D Object Detection}
Plane segmentation is used for detecting parcels \cite{huCuboidDetectionTracking2021} and pallets \cite{xiaoPalletRecognitionLocalization2017, molterRealtimePalletLocalization2018}.
Moreover, clustering approaches \cite{fontanaComparativeAssessmentParcel2021} and deep learning-based pointcloud classification \cite{brylkaAIbasedRecognitionDangerous2021} have been used.

\ownsec{3D Shape Reconstruction}
There are approaches for dataset generation that heavily rely on fiducial markers \cite{mihalyiRobust3DObject2015}.
Furthermore, digital measurement with \rgbd{} cameras has been investigated in static scenes with traditional methods \cite{kuckelhausDHLLowcostSensor2013,sonMethodConstructAutomatic2017,kucukDevelopmentDimensionsMeasurement2019}.
Finally, \textcite{naumannParcel3DShapeReconstruction2023} tackle single image 3D reconstruction for assessing parcel damages.

\subsection{Datasets}
\label{sec:other:datasets}

The availability of freely available datasets is limited, as we can infer from Tables \ref{tab:overview:documentation}, \ref{tab:overview:verification}, \ref{tab:overview:assistance_for_manipulation} and \ref{tab:overview:autonomous_manipulation}.
Datasets exist for instance segmentation of logistics objects \cite{mayershoferLOCOLogisticsObjects2020} and parcels \cite{naumannRefinedPlaneSegmentation2020,naumannParcel3DShapeReconstruction2023}.
Furthermore, chipwood re-identification \cite{rutinowskiReIdentificationWarehousingEntities2021,rutinowskiDeepLearningBased2022} datasets have been presented.
Most of the presented works, however, utilize datasets that are not available to the public and are mostly described only briefly.

\subsection{Industry Services}
\label{sec:other:industry}

Due to the huge potential of computer vision applications in logistics and the market potential of such solutions, numerous companies offer commercial products and services in this area.
We present an extract of relevant companies and selected products in \cref{tab:overview:industry}.
This overview shows that companies are actively working on solutions for all the areas mentioned in \cref{sec:monitoring} and \ref{sec:manipulation}.

The overview presented here is not complete.
We provide further details online on our project website and invite the audience to use it and contribute.

\begin{landscape}
    \begin{table}[t!]
    \centering\footnotesize
    \caption{Overview of Industry Solutions (cf. \cref{sec:other:industry}).}
    \label{tab:overview:industry}
    \begin{tabularx}{\linewidth}{>{\hsize=0.5\hsize\RaggedRight}X>{\hsize=0.7\hsize\RaggedRight}X>{\hsize=1.8\hsize\RaggedRight}X>{\hsize=1\hsize\RaggedRight}X}
        \toprule
        Company              & Product                                               & Description                                                                                                                           & Application                                                                          \\
        \midrule
        CARGOMETER GmbH      & CARGOMETER                                            & On the fly measurement and weighing for a moving forklift. Hardware is installed at loading gates, where every pallet has to pass by. & Object Detection, Pallet Handling, Tracking and Tracing, Volume Estimation           \\
        Cognex               & 3D-A1000 dimensioning system                          & Capture moving objects in 2D and 3D for identification, volume estimation and damage detection.                                       & Damage and Tampering Detection, Label Detection, Object Detection, Volume Estimation \\
        FIZYR                & Parcel handling                                       & Flexible AI-powered software solution, e.g. for parcel pick-and-place. Detects deformation, damages, wrinkles and labels.             & Label Detection, Object Detection, Pick-and-Place                                    \\
        HIKROBOT             & Single Piece Separation System                        & Real-time positioning of packages to ensure that packages pass through a gate individually and at a pre-defined rate.                 & Sorting                                                                              \\
        KEYENCE              & AGV safety precautions                                & Laser scanner for AGVs to ensure safety                                                                                               & AGVs, Object Detection                                                               \\
        LMI                  & 3D Scanning and Inspection for the Packaging Industry & Package filling inspection by scanning and 3D measuring to ensure the correct fill-level                                              & Object Detection, Quality Control                                                    \\
        Loadscan             & Load Volume Scanner system                            & Measuring the volume of the material loaded in a truck or trailer bin. Based on laser scanner and proprietary software.               & Volume Estimation                                                                    \\
        Locus Robotics       & Directed Picking                                      & Support during picking process by actively directing workers to their next picking location.                                          & Order Picking                                                                        \\
        Omron                & Vision and Inspection                                 & Label inspection for ISO and GS1 standards                                                                                            & Label Detection                                                                      \\
        River Systems        & Mobile Sort                                           & Picking and sorting solution to enable batch picking with subsequent sorting into distinct orders.                                    & Order Picking, Sorting                                                               \\
        TeamViewer           & xPick                                                 & Augmented reality solution to support order picking, sorting, inventory control and other processes in logistics and warehousing      & Augmented Reality Assistance, Order Picking                                          \\
        VC Vision Components & Advanced Driver-Assistance System                     & Advanced driver-assistance systems for forklifts                                                                                      & AGVs                                                                                 \\
        VITRONIC             & Air Freight Measurement                               & Air Freight Measurement system                                                                                                        & Volume Estimation                                                                    \\
        Valerann             & Monitor                                               & Monitoring of road traffic                                                                                                            & Vehicle Traffic                                                                      \\
        Zebra                & Packing and Staging                                   & Order confirmation and shipping label printing for hands-free packing and staging in warehouses and distribution centers.             & Other Assistance                                                                     \\
        \bottomrule
    \end{tabularx}
\end{table}
\end{landscape}

\section{Conclusion}
\label{sec:conclusion}

We present a detailed overview of the literature on computer vision applications in transportation logistics and warehousing.
First, we consider monitoring applications, where the goal is the retrieval of relevant information from visual input.
Afterwards, applications were considered where not only observing an environment is important, but also the interaction therewith.
In addition to the summarization and categorization according to the area of application, we provide an overview that is based on the computer vision task at hand.
Finally, we provide an overview of existing datasets and industry solutions.
We conclude that computer vision applications in transportation logistics and warehousing are crucial to increase efficiency in this highly competitive domain, which explains the active research in academia and industry.
All data is available in a structured format on \urlreview{} and we invite the interested readers to contribute.

\bibliography{literature/literature.bib}%

\begin{thebibliography}{}
\renewcommand{\doi}[1]{\url{https://doi.org/#1}}
\bibcommenthead

\bibitem [\protect \citeauthoryear {%
Arpenti%
\ \protect \BOthers {.}}{%
Arpenti%
\ \protect \BOthers {.}}{%
{\protect \APACyear {2020}}%
}]{%
arpentiRGBDRecognitionLocalization2020}
\APACinsertmetastar {%
arpentiRGBDRecognitionLocalization2020}%
\begin{APACrefauthors}%
Arpenti, P.%
, Caccavale, R.%
, Paduano, G.%
, Andrea~Fontanelli, G.%
, Lippiello, V.%
, Villani, L.%
\BCBL {} Siciliano, B.%
\end{APACrefauthors}%
\unskip\
\newblock
\APACrefYearMonthDay{2020}{{\APACmonth{10}}}{}.
\newblock
{\BBOQ}\APACrefatitle {{{RGB-D Recognition}} and {{Localization}} of {{Cases}} for {{Robotic Depalletizing}} in {{Supermarkets}}} {{{RGB-D Recognition}} and {{Localization}} of {{Cases}} for {{Robotic Depalletizing}} in {{Supermarkets}}}.{\BBCQ}
\newblock
\APACjournalVolNumPages{IEEE Robotics and Automation Letters}{5}{4}{6233--6238,}
\newblock
\begin{APACrefDOI} \doi{10/gh547b} \end{APACrefDOI}
\newblock

\newblock

\PrintBackRefs{\CurrentBib}

\bibitem [\protect \citeauthoryear {%
Azadeh%
, De~Koster%
\BCBL {}\ \BBA {} Roy%
}{%
Azadeh%
\ \protect \BOthers {.}}{%
{\protect \APACyear {2019}}%
}]{%
azadehRobotizedAutomatedWarehouse2019}
\APACinsertmetastar {%
azadehRobotizedAutomatedWarehouse2019}%
\begin{APACrefauthors}%
Azadeh, K.%
, De~Koster, R.%
\BCBL {} Roy, D.%
\end{APACrefauthors}%
\unskip\
\newblock
\APACrefYearMonthDay{2019}{{\APACmonth{07}}}{}.
\newblock
{\BBOQ}\APACrefatitle {Robotized and {{Automated Warehouse Systems}}: {{Review}} and {{Recent Developments}}} {Robotized and {{Automated Warehouse Systems}}: {{Review}} and {{Recent Developments}}}.{\BBCQ}
\newblock
\APACjournalVolNumPages{Transportation Science}{53}{4}{917--945,}
\newblock
\begin{APACrefDOI} \doi{10/ggdgbq} \end{APACrefDOI}
\newblock

\newblock

\PrintBackRefs{\CurrentBib}

\bibitem [\protect \citeauthoryear {%
Bai%
\ \protect \BOthers {.}}{%
Bai%
\ \protect \BOthers {.}}{%
{\protect \APACyear {2020}}%
}]{%
baiObjectDetectionRecognition2020}
\APACinsertmetastar {%
baiObjectDetectionRecognition2020}%
\begin{APACrefauthors}%
Bai, Q.%
, Li, S.%
, Yang, J.%
, Song, Q.%
, Li, Z.%
\BCBL {} Zhang, X.%
\end{APACrefauthors}%
\unskip\
\newblock
\APACrefYearMonthDay{2020}{}{}.
\newblock
{\BBOQ}\APACrefatitle {Object {{Detection Recognition}} and {{Robot Grasping Based}} on {{Machine Learning}}: {{A Survey}}} {Object {{Detection Recognition}} and {{Robot Grasping Based}} on {{Machine Learning}}: {{A Survey}}}.{\BBCQ}
\newblock
\APACjournalVolNumPages{IEEE Access}{8}{}{181855--181879,}
\newblock
\begin{APACrefDOI} \doi{10.1109/ACCESS.2020.3028740} \end{APACrefDOI}
\newblock

\newblock

\PrintBackRefs{\CurrentBib}

\bibitem [\protect \citeauthoryear {%
Bay%
, Ess%
, Tuytelaars%
\BCBL {}\ \BBA {} Van~Gool%
}{%
Bay%
\ \protect \BOthers {.}}{%
{\protect \APACyear {2008}}%
}]{%
baySpeededUpRobustFeatures2008}
\APACinsertmetastar {%
baySpeededUpRobustFeatures2008}%
\begin{APACrefauthors}%
Bay, H.%
, Ess, A.%
, Tuytelaars, T.%
\BCBL {} Van~Gool, L.%
\end{APACrefauthors}%
\unskip\
\newblock
\APACrefYearMonthDay{2008}{{\APACmonth{06}}}{}.
\newblock
{\BBOQ}\APACrefatitle {Speeded-{{Up Robust Features}} ({{SURF}})} {Speeded-{{Up Robust Features}} ({{SURF}})}.{\BBCQ}
\newblock
\APACjournalVolNumPages{Computer Vision and Image Understanding}{110}{3}{346--359,}
\newblock
\begin{APACrefDOI} \doi{10/ffsc9r} \end{APACrefDOI}
\newblock

\newblock

\PrintBackRefs{\CurrentBib}

\bibitem [\protect \citeauthoryear {%
Bello%
, Yu%
, Wang%
, Adam%
\BCBL {}\ \BBA {} Li%
}{%
Bello%
\ \protect \BOthers {.}}{%
{\protect \APACyear {2020}}%
}]{%
belloReviewDeepLearning2020}
\APACinsertmetastar {%
belloReviewDeepLearning2020}%
\begin{APACrefauthors}%
Bello, S.A.%
, Yu, S.%
, Wang, C.%
, Adam, J.M.%
\BCBL {} Li, J.%
\end{APACrefauthors}%
\unskip\
\newblock
\APACrefYearMonthDay{2020}{{\APACmonth{01}}}{}.
\newblock
{\BBOQ}\APACrefatitle {Review: {{Deep Learning}} on {{3D Point Clouds}}} {Review: {{Deep Learning}} on {{3D Point Clouds}}}.{\BBCQ}
\newblock
\APACjournalVolNumPages{Remote Sensing}{12}{11}{1729,}
\newblock
\begin{APACrefDOI} \doi{10.3390/rs12111729} \end{APACrefDOI}
\newblock

\newblock

\PrintBackRefs{\CurrentBib}

\bibitem [\protect \citeauthoryear {%
Benslimane%
, Tamayo%
\BCBL {}\ \BBA {} {de La Fortelle}%
}{%
Benslimane%
\ \protect \BOthers {.}}{%
{\protect \APACyear {2019}}%
}]{%
benslimaneClassifyingLogisticVehicles2019}
\APACinsertmetastar {%
benslimaneClassifyingLogisticVehicles2019}%
\begin{APACrefauthors}%
Benslimane, S.%
, Tamayo, S.%
\BCBL {} {de La Fortelle}, A.%
\end{APACrefauthors}%
\unskip\
\newblock
\APACrefYearMonthDay{2019}{{\APACmonth{06}}}{}.
\newblock
{\BBOQ}\APACrefatitle {Classifying Logistic Vehicles in Cities Using {{Deep}} Learning} {Classifying logistic vehicles in cities using {{Deep}} learning}.{\BBCQ}
\newblock
\APACjournalVolNumPages{arXiv:1906.11895 [cs, stat]}{}{}{,}
\newblock
{\href{https://arxiv.org/abs/1906.11895}{{arxiv:1906.11895}}}
\newblock
 {[cs, stat]}
\PrintBackRefs{\CurrentBib}

\bibitem [\protect \citeauthoryear {%
Bicchi%
\ \BBA {} Kumar%
}{%
Bicchi%
\ \BBA {} Kumar%
}{%
{\protect \APACyear {2000}}%
}]{%
bicchiRoboticGraspingContact2000}
\APACinsertmetastar {%
bicchiRoboticGraspingContact2000}%
\begin{APACrefauthors}%
Bicchi, A.%
\BCBT {}\ \BBA {} Kumar, V.%
\end{APACrefauthors}%
\unskip\
\newblock
\APACrefYearMonthDay{2000}{{\APACmonth{04}}}{}.
\newblock
{\BBOQ}\APACrefatitle {Robotic Grasping and Contact: A Review} {Robotic grasping and contact: A review}.{\BBCQ}
\newblock
 \APACrefbtitle {Proceedings 2000 {{ICRA}}. {{Millennium Conference}}. {{IEEE International Conference}} on {{Robotics}} and {{Automation}}. {{Symposia Proceedings}} ({{Cat}}. {{No}}.{{00CH37065}})} {Proceedings 2000 {{ICRA}}. {{Millennium Conference}}. {{IEEE International Conference}} on {{Robotics}} and {{Automation}}. {{Symposia Proceedings}} ({{Cat}}. {{No}}.{{00CH37065}})}\ (\BVOL~1, \BPG~348-353 vol.1).
\PrintBackRefs{\CurrentBib}

\bibitem [\protect \citeauthoryear {%
Bonini%
, Urru%
\BCBL {}\ \BBA {} Echelmeyer%
}{%
Bonini%
\ \protect \BOthers {.}}{%
{\protect \APACyear {2016}}%
}]{%
boniniFastDeployableAutonomous2016}
\APACinsertmetastar {%
boniniFastDeployableAutonomous2016}%
\begin{APACrefauthors}%
Bonini, M.%
, Urru, A.%
\BCBL {} Echelmeyer, W.%
\end{APACrefauthors}%
\unskip\
\newblock
\APACrefYearMonthDay{2016}{}{}.
\newblock
{\BBOQ}\APACrefatitle {Fast {{Deployable Autonomous Systems}} for {{Order Picking}} - {{How Small}} and {{Medium Size Enterprises Can Benefit}} from the {{Automation}} of the {{Picking Process}}:} {Fast {{Deployable Autonomous Systems}} for {{Order Picking}} - {{How Small}} and {{Medium Size Enterprises Can Benefit}} from the {{Automation}} of the {{Picking Process}}:}.{\BBCQ}
\newblock
 \APACrefbtitle {Proceedings of the 13th {{International Conference}} on {{Informatics}} in {{Control}}, {{Automation}} and {{Robotics}}} {Proceedings of the 13th {{International Conference}} on {{Informatics}} in {{Control}}, {{Automation}} and {{Robotics}}}\ (\BPGS\ 479--484).
\newblock
\APACaddressPublisher{{Lisbon, Portugal}}{{SCITEPRESS - Science and and Technology Publications}}.
\PrintBackRefs{\CurrentBib}

\bibitem [\protect \citeauthoryear {%
Borstell%
}{%
Borstell%
}{%
{\protect \APACyear {2018}}%
}]{%
borstellShortSurveyImage2018}
\APACinsertmetastar {%
borstellShortSurveyImage2018}%
\begin{APACrefauthors}%
Borstell, H.%
\end{APACrefauthors}%
\unskip\
\newblock
\APACrefYearMonthDay{2018}{}{}.
\newblock
{\BBOQ}\APACrefatitle {A Short Survey on Image Processing in Logistics} {A short survey on image processing in logistics}.{\BBCQ}
\newblock
 \APACrefbtitle {11th {{International Doctoral Student Workshop}} on {{Logistics}}.} {11th {{International Doctoral Student Workshop}} on {{Logistics}}.}
\newblock
\APACaddressPublisher{{Magdeburg}}{}.
\PrintBackRefs{\CurrentBib}

\bibitem [\protect \citeauthoryear {%
Borstell%
}{%
Borstell%
}{%
{\protect \APACyear {2021}}%
}]{%
borstellBildbasierteZustandserfassungLogistik2021}
\APACinsertmetastar {%
borstellBildbasierteZustandserfassungLogistik2021}%
\begin{APACrefauthors}%
Borstell, H.%
\end{APACrefauthors}%
\unskip\
\newblock
\APACrefYear{2021}.
\unskip\
\newblock
\APACrefbtitle {{Bildbasierte Zustandserfassung in der Logistik}} {{Bildbasierte Zustandserfassung in der Logistik}}\ \APACtypeAddressSchool {\BUPhD}{}{}.
\unskip\
\newblock
\APACaddressSchool {{Magdeburg}}{Otto-von-Guericke-Universit\"at Magdeburg}.
\PrintBackRefs{\CurrentBib}

\bibitem [\protect \citeauthoryear {%
Borstell%
, Cao%
, Kluth%
\BCBL {}\ \BBA {} Richter%
}{%
Borstell%
\ \protect \BOthers {.}}{%
{\protect \APACyear {2013}}%
}]{%
borstellProzessintegrierteVolumenerfassungLogistischen2013}
\APACinsertmetastar {%
borstellProzessintegrierteVolumenerfassungLogistischen2013}%
\begin{APACrefauthors}%
Borstell, H.%
, Cao, L.%
, Kluth, J.%
\BCBL {} Richter, K.%
\end{APACrefauthors}%
\unskip\
\newblock
\APACrefYearMonthDay{2013}{}{}.
\newblock
{\BBOQ}\APACrefatitle {{Prozessintegrierte Volumenerfassung von logistischen Palettenstrukturen auf Basis von Low-Cost- Tiefenbildsensoren}} {{Prozessintegrierte Volumenerfassung von logistischen Palettenstrukturen auf Basis von Low-Cost- Tiefenbildsensoren}}.{\BBCQ}
\newblock
 \APACrefbtitle {{Tagungsband / 3D-NordOst 2013, 16. Anwendungsbezogener Workshop zur Erfassung, Modellierung, Verarbeitung und Auswertung von 3D-Daten, im Rahmen der GFaI-Workshop-Familie NordOst, Berlin, 12./13. Dezember 2013}} {{Tagungsband / 3D-NordOst 2013, 16. Anwendungsbezogener Workshop zur Erfassung, Modellierung, Verarbeitung und Auswertung von 3D-Daten, im Rahmen der GFaI-Workshop-Familie NordOst, Berlin, 12./13. Dezember 2013}}\ (\BPG~11).
\PrintBackRefs{\CurrentBib}

\bibitem [\protect \citeauthoryear {%
Borstell%
\ \protect \BOthers {.}}{%
Borstell%
\ \protect \BOthers {.}}{%
{\protect \APACyear {2014}}%
}]{%
borstellPalletMonitoringSystem2014}
\APACinsertmetastar {%
borstellPalletMonitoringSystem2014}%
\begin{APACrefauthors}%
Borstell, H.%
, Kluth, J.%
, Jaeschke, M.%
, Plate, C.%
, Gebert, B.%
\BCBL {} Richter, K.%
\end{APACrefauthors}%
\unskip\
\newblock
\APACrefYearMonthDay{2014}{{\APACmonth{10}}}{}.
\newblock
{\BBOQ}\APACrefatitle {Pallet Monitoring System Based on a Heterogeneous Sensor Network for Transparent Warehouse Processes} {Pallet monitoring system based on a heterogeneous sensor network for transparent warehouse processes}.{\BBCQ}
\newblock
 \APACrefbtitle {2014 {{Sensor Data Fusion}}: {{Trends}}, {{Solutions}}, {{Applications}} ({{SDF}})} {2014 {{Sensor Data Fusion}}: {{Trends}}, {{Solutions}}, {{Applications}} ({{SDF}})}\ (\BPGS\ 1--6).
\PrintBackRefs{\CurrentBib}

\bibitem [\protect \citeauthoryear {%
Brazil%
\ \protect \BOthers {.}}{%
Brazil%
\ \protect \BOthers {.}}{%
{\protect \APACyear {2023}}%
}]{%
brazilOmni3DLargeBenchmark2023}
\APACinsertmetastar {%
brazilOmni3DLargeBenchmark2023}%
\begin{APACrefauthors}%
Brazil, G.%
, Kumar, A.%
, Straub, J.%
, Ravi, N.%
, Johnson, J.%
\BCBL {} Gkioxari, G.%
\end{APACrefauthors}%
\unskip\
\newblock
\APACrefYearMonthDay{2023}{{\APACmonth{06}}}{}.
\newblock
{\BBOQ}\APACrefatitle {{{Omni3D}}: {{A Large Benchmark}} and {{Model}} for {{3D Object Detection}} in the {{Wild}}} {{{Omni3D}}: {{A Large Benchmark}} and {{Model}} for {{3D Object Detection}} in the {{Wild}}}.{\BBCQ}
\newblock
 \APACrefbtitle {2023 {{IEEE}}/{{CVF Conference}} on {{Computer Vision}} and {{Pattern Recognition}} ({{CVPR}}).} {2023 {{IEEE}}/{{CVF Conference}} on {{Computer Vision}} and {{Pattern Recognition}} ({{CVPR}}).}
\newblock
\APACaddressPublisher{{Vancouver, Canada}}{}.
\PrintBackRefs{\CurrentBib}

\bibitem [\protect \citeauthoryear {%
Bromley%
, Guyon%
, LeCun%
, S{\"a}ckinger%
\BCBL {}\ \BBA {} Shah%
}{%
Bromley%
\ \protect \BOthers {.}}{%
{\protect \APACyear {1993}}%
}]{%
bromleySignatureVerificationUsing1993}
\APACinsertmetastar {%
bromleySignatureVerificationUsing1993}%
\begin{APACrefauthors}%
Bromley, J.%
, Guyon, I.%
, LeCun, Y.%
, S{\"a}ckinger, E.%
\BCBL {} Shah, R.%
\end{APACrefauthors}%
\unskip\
\newblock
\APACrefYearMonthDay{1993}{}{}.
\newblock
{\BBOQ}\APACrefatitle {Signature Verification Using a "Siamese" Time Delay Neural Network} {Signature verification using a "siamese" time delay neural network}.{\BBCQ}
\newblock
\APACjournalVolNumPages{Advances in neural information processing systems}{6}{}{,}
\newblock

\newblock

\PrintBackRefs{\CurrentBib}

\bibitem [\protect \citeauthoryear {%
Brylka%
, Bierwirth%
\BCBL {}\ \BBA {} Schwanecke%
}{%
Brylka%
\ \protect \BOthers {.}}{%
{\protect \APACyear {2021}}%
}]{%
brylkaAIbasedRecognitionDangerous2021}
\APACinsertmetastar {%
brylkaAIbasedRecognitionDangerous2021}%
\begin{APACrefauthors}%
Brylka, R.%
, Bierwirth, B.%
\BCBL {} Schwanecke, U.%
\end{APACrefauthors}%
\unskip\
\newblock
\APACrefYearMonthDay{2021}{{\APACmonth{12}}}{}.
\newblock
{\BBOQ}\APACrefatitle {{{AI-based}} Recognition of Dangerous Goods Labels and Metric Package Features} {{{AI-based}} recognition of dangerous goods labels and metric package features}.{\BBCQ}
\newblock
 \APACrefbtitle {Proceedings of the {{Hamburg International Conference}} of {{Logistics}} ({{HICL}})} {Proceedings of the {{Hamburg International Conference}} of {{Logistics}} ({{HICL}})}\ (\BPGS\ 245--272).
\newblock
\APACaddressPublisher{}{{epubli}}.
\PrintBackRefs{\CurrentBib}

\bibitem [\protect \citeauthoryear {%
Brylka%
, Schwanecke%
\BCBL {}\ \BBA {} Bierwirth%
}{%
Brylka%
\ \protect \BOthers {.}}{%
{\protect \APACyear {2020}}%
}]{%
brylkaCameraBasedBarcode2020}
\APACinsertmetastar {%
brylkaCameraBasedBarcode2020}%
\begin{APACrefauthors}%
Brylka, R.%
, Schwanecke, U.%
\BCBL {} Bierwirth, B.%
\end{APACrefauthors}%
\unskip\
\newblock
\APACrefYearMonthDay{2020}{{\APACmonth{08}}}{}.
\newblock
{\BBOQ}\APACrefatitle {Camera {{Based Barcode Localization}} and {{Decoding}} in {{Real-World Applications}}} {Camera {{Based Barcode Localization}} and {{Decoding}} in {{Real-World Applications}}}.{\BBCQ}
\newblock
 \APACrefbtitle {2020 {{International Conference}} on {{Omni-layer Intelligent Systems}} ({{COINS}})} {2020 {{International Conference}} on {{Omni-layer Intelligent Systems}} ({{COINS}})}\ (\BPGS\ 1--8).
\PrintBackRefs{\CurrentBib}

\bibitem [\protect \citeauthoryear {%
Canny%
}{%
Canny%
}{%
{\protect \APACyear {1986}}%
}]{%
cannyComputationalApproachEdge1986}
\APACinsertmetastar {%
cannyComputationalApproachEdge1986}%
\begin{APACrefauthors}%
Canny, J.%
\end{APACrefauthors}%
\unskip\
\newblock
\APACrefYearMonthDay{1986}{{\APACmonth{11}}}{}.
\newblock
{\BBOQ}\APACrefatitle {A {{Computational Approach}} to {{Edge Detection}}} {A {{Computational Approach}} to {{Edge Detection}}}.{\BBCQ}
\newblock
\APACjournalVolNumPages{IEEE Transactions on Pattern Analysis and Machine Intelligence}{PAMI-8}{6}{679--698,}
\newblock
\begin{APACrefDOI} \doi{10/fn3fdk} \end{APACrefDOI}
\newblock

\newblock

\PrintBackRefs{\CurrentBib}

\bibitem [\protect \citeauthoryear {%
Chiaravalli%
, Palli%
, Monica%
, Aleotti%
\BCBL {}\ \BBA {} Rizzini%
}{%
Chiaravalli%
\ \protect \BOthers {.}}{%
{\protect \APACyear {2020}}%
}]{%
chiaravalliIntegrationMultiCameraVision2020}
\APACinsertmetastar {%
chiaravalliIntegrationMultiCameraVision2020}%
\begin{APACrefauthors}%
Chiaravalli, D.%
, Palli, G.%
, Monica, R.%
, Aleotti, J.%
\BCBL {} Rizzini, D.L.%
\end{APACrefauthors}%
\unskip\
\newblock
\APACrefYearMonthDay{2020}{{\APACmonth{09}}}{}.
\newblock
{\BBOQ}\APACrefatitle {Integration of a {{Multi-Camera Vision System}} and {{Admittance Control}} for {{Robotic Industrial Depalletizing}}} {Integration of a {{Multi-Camera Vision System}} and {{Admittance Control}} for {{Robotic Industrial Depalletizing}}}.{\BBCQ}
\newblock
 \APACrefbtitle {2020 25th {{IEEE International Conference}} on {{Emerging Technologies}} and {{Factory Automation}} ({{ETFA}})} {2020 25th {{IEEE International Conference}} on {{Emerging Technologies}} and {{Factory Automation}} ({{ETFA}})}\ (\BVOL~1, \BPGS\ 667--674).
\PrintBackRefs{\CurrentBib}

\bibitem [\protect \citeauthoryear {%
Clausen%
, Zelenka%
, Schwede%
\BCBL {}\ \BBA {} Koch%
}{%
Clausen%
\ \protect \BOthers {.}}{%
{\protect \APACyear {2019}}%
}]{%
clausenParcelTrackingDetection2019}
\APACinsertmetastar {%
clausenParcelTrackingDetection2019}%
\begin{APACrefauthors}%
Clausen, S.%
, Zelenka, C.%
, Schwede, T.%
\BCBL {} Koch, R.%
\end{APACrefauthors}%
\unskip\
\newblock
\APACrefYearMonthDay{2019}{}{}.
\newblock
{\BBOQ}\APACrefatitle {Parcel {{Tracking}} by {{Detection}} in {{Large Camera Networks}}} {Parcel {{Tracking}} by {{Detection}} in {{Large Camera Networks}}}.{\BBCQ}
\newblock
 T.~Brox, A.~Bruhn\BCBL {}\ \BBA {} M.~Fritz\ (\BEDS), \APACrefbtitle {Pattern {{Recognition}}} {Pattern {{Recognition}}}\ (\BPGS\ 89--104).
\newblock
\APACaddressPublisher{{Cham}}{{Springer International Publishing}}.
\PrintBackRefs{\CurrentBib}

\bibitem [\protect \citeauthoryear {%
Correll%
\ \protect \BOthers {.}}{%
Correll%
\ \protect \BOthers {.}}{%
{\protect \APACyear {2018}}%
}]{%
correllAnalysisObservationsFirst2018}
\APACinsertmetastar {%
correllAnalysisObservationsFirst2018}%
\begin{APACrefauthors}%
Correll, N.%
, Bekris, K.E.%
, Berenson, D.%
, Brock, O.%
, Causo, A.%
, Hauser, K.%
\BDBL {}Wurman, P.R.%
\end{APACrefauthors}%
\unskip\
\newblock
\APACrefYearMonthDay{2018}{{\APACmonth{01}}}{}.
\newblock
{\BBOQ}\APACrefatitle {Analysis and {{Observations From}} the {{First Amazon Picking Challenge}}} {Analysis and {{Observations From}} the {{First Amazon Picking Challenge}}}.{\BBCQ}
\newblock
\APACjournalVolNumPages{IEEE Transactions on Automation Science and Engineering}{15}{1}{172--188,}
\newblock
\begin{APACrefDOI} \doi{10.1109/TASE.2016.2600527} \end{APACrefDOI}
\newblock

\newblock

\PrintBackRefs{\CurrentBib}

\bibitem [\protect \citeauthoryear {%
Cronin%
, Conway%
\BCBL {}\ \BBA {} Walsh%
}{%
Cronin%
\ \protect \BOthers {.}}{%
{\protect \APACyear {2019}}%
}]{%
croninStateoftheArtReviewAutonomous2019}
\APACinsertmetastar {%
croninStateoftheArtReviewAutonomous2019}%
\begin{APACrefauthors}%
Cronin, C.%
, Conway, A.%
\BCBL {} Walsh, J.%
\end{APACrefauthors}%
\unskip\
\newblock
\APACrefYearMonthDay{2019}{{\APACmonth{06}}}{}.
\newblock
{\BBOQ}\APACrefatitle {State-of-the-{{Art Review}} of {{Autonomous Intelligent Vehicles}} ({{AIV}}) {{Technologies}} for the {{Automotive}} and {{Manufacturing Industry}}} {State-of-the-{{Art Review}} of {{Autonomous Intelligent Vehicles}} ({{AIV}}) {{Technologies}} for the {{Automotive}} and {{Manufacturing Industry}}}.{\BBCQ}
\newblock
 \APACrefbtitle {2019 30th {{Irish Signals}} and {{Systems Conference}} ({{ISSC}})} {2019 30th {{Irish Signals}} and {{Systems Conference}} ({{ISSC}})}\ (\BPGS\ 1--6).
\PrintBackRefs{\CurrentBib}

\bibitem [\protect \citeauthoryear {%
Dalal%
\ \BBA {} Triggs%
}{%
Dalal%
\ \BBA {} Triggs%
}{%
{\protect \APACyear {2005}}%
}]{%
dalalHistogramsOrientedGradients2005}
\APACinsertmetastar {%
dalalHistogramsOrientedGradients2005}%
\begin{APACrefauthors}%
Dalal, N.%
\BCBT {}\ \BBA {} Triggs, B.%
\end{APACrefauthors}%
\unskip\
\newblock
\APACrefYearMonthDay{2005}{{\APACmonth{06}}}{}.
\newblock
{\BBOQ}\APACrefatitle {Histograms of Oriented Gradients for Human Detection} {Histograms of oriented gradients for human detection}.{\BBCQ}
\newblock
 \APACrefbtitle {2005 {{IEEE Computer Society Conference}} on {{Computer Vision}} and {{Pattern Recognition}} ({{CVPR}}'05)} {2005 {{IEEE Computer Society Conference}} on {{Computer Vision}} and {{Pattern Recognition}} ({{CVPR}}'05)}\ (\BVOL~1, \BPG~886-893 vol. 1).
\PrintBackRefs{\CurrentBib}

\bibitem [\protect \citeauthoryear {%
Das%
, Ma%
, Shu%
\BCBL {}\ \BBA {} Samaras%
}{%
Das%
\ \protect \BOthers {.}}{%
{\protect \APACyear {2022}}%
}]{%
dasLearningIsometricSurface2022}
\APACinsertmetastar {%
dasLearningIsometricSurface2022}%
\begin{APACrefauthors}%
Das, S.%
, Ma, K.%
, Shu, Z.%
\BCBL {} Samaras, D.%
\end{APACrefauthors}%
\unskip\
\newblock
\APACrefYearMonthDay{2022}{}{}.
\newblock
{\BBOQ}\APACrefatitle {Learning an {{Isometric Surface Parameterization}} for {{Texture Unwrapping}}} {Learning an {{Isometric Surface Parameterization}} for {{Texture Unwrapping}}}.{\BBCQ}
\newblock
 S.~Avidan, G.~Brostow, M.~Ciss{\'e}, G.M.~Farinella\BCBL {}\ \BBA {} T.~Hassner\ (\BEDS), \APACrefbtitle {Computer {{Vision}} \textendash{} {{ECCV}} 2022} {Computer {{Vision}} \textendash{} {{ECCV}} 2022}\ (\BVOL\ 13697, \BPGS\ 580--597).
\newblock
\APACaddressPublisher{{Cham}}{{Springer Nature Switzerland}}.
\PrintBackRefs{\CurrentBib}

\bibitem [\protect \citeauthoryear {%
D{\"o}rr%
, Brandt%
, Meyer%
\BCBL {}\ \BBA {} Pouls%
}{%
D{\"o}rr%
\ \protect \BOthers {.}}{%
{\protect \APACyear {2019}}%
}]{%
dorrLeanTrainingData2019}
\APACinsertmetastar {%
dorrLeanTrainingData2019}%
\begin{APACrefauthors}%
D{\"o}rr, L.%
, Brandt, F.%
, Meyer, A.%
\BCBL {} Pouls, M.%
\end{APACrefauthors}%
\unskip\
\newblock
\APACrefYearMonthDay{2019}{{\APACmonth{12}}}{}.
\newblock
{\BBOQ}\APACrefatitle {Lean {{Training Data Generation}} for {{Planar Object Detection Models}} in {{Unsteady Logistics Contexts}}} {Lean {{Training Data Generation}} for {{Planar Object Detection Models}} in {{Unsteady Logistics Contexts}}}.{\BBCQ}
\newblock
 \APACrefbtitle {{{IEEE International Conference On Machine Learning And Applications}} ({{ICMLA}})} {{{IEEE International Conference On Machine Learning And Applications}} ({{ICMLA}})}\ (\BPGS\ 329--334).
\newblock
\APACaddressPublisher{{Boca Raton, FL, USA}}{{IEEE}}.
\PrintBackRefs{\CurrentBib}

\bibitem [\protect \citeauthoryear {%
D{\"o}rr%
, Brandt%
, Naumann%
\BCBL {}\ \BBA {} Pouls%
}{%
D{\"o}rr%
\ \protect \BOthers {.}}{%
{\protect \APACyear {2021}}%
}]{%
dorrTetraPackNetFourCornerBasedObject2021}
\APACinsertmetastar {%
dorrTetraPackNetFourCornerBasedObject2021}%
\begin{APACrefauthors}%
D{\"o}rr, L.%
, Brandt, F.%
, Naumann, A.%
\BCBL {} Pouls, M.%
\end{APACrefauthors}%
\unskip\
\newblock
\APACrefYearMonthDay{2021}{}{}.
\newblock
{\BBOQ}\APACrefatitle {{{TetraPackNet}}: {{Four-Corner-Based Object Detection}} in {{Logistics Use-Cases}}} {{{TetraPackNet}}: {{Four-Corner-Based Object Detection}} in {{Logistics Use-Cases}}}.{\BBCQ}
\newblock
 \APACrefbtitle {{{DAGM German Conference}} on {{Pattern Recognition}}.} {{{DAGM German Conference}} on {{Pattern Recognition}}.}
\PrintBackRefs{\CurrentBib}

\bibitem [\protect \citeauthoryear {%
D{\"o}rr%
, Brandt%
, Pouls%
\BCBL {}\ \BBA {} Naumann%
}{%
D{\"o}rr%
\ \protect \BOthers {.}}{%
{\protect \APACyear {2020}}%
}]{%
dorrFullyAutomatedPackagingStructure2020}
\APACinsertmetastar {%
dorrFullyAutomatedPackagingStructure2020}%
\begin{APACrefauthors}%
D{\"o}rr, L.%
, Brandt, F.%
, Pouls, M.%
\BCBL {} Naumann, A.%
\end{APACrefauthors}%
\unskip\
\newblock
\APACrefYearMonthDay{2020}{{\APACmonth{08}}}{}.
\newblock
{\BBOQ}\APACrefatitle {Fully-{{Automated Packaging Structure Recognition}} in {{Logistics Environments}}} {Fully-{{Automated Packaging Structure Recognition}} in {{Logistics Environments}}}.{\BBCQ}
\newblock
 \APACrefbtitle {International {{Conference}} on {{Emerging Technologies}} and {{Factory Automation}}.} {International {{Conference}} on {{Emerging Technologies}} and {{Factory Automation}}.}
\PrintBackRefs{\CurrentBib}

\bibitem [\protect \citeauthoryear {%
Du%
, Wang%
, Lian%
\BCBL {}\ \BBA {} Zhao%
}{%
Du%
\ \protect \BOthers {.}}{%
{\protect \APACyear {2021}}%
}]{%
duVisionbasedRoboticGrasping2021}
\APACinsertmetastar {%
duVisionbasedRoboticGrasping2021}%
\begin{APACrefauthors}%
Du, G.%
, Wang, K.%
, Lian, S.%
\BCBL {} Zhao, K.%
\end{APACrefauthors}%
\unskip\
\newblock
\APACrefYearMonthDay{2021}{{\APACmonth{03}}}{}.
\newblock
{\BBOQ}\APACrefatitle {Vision-Based Robotic Grasping from Object Localization, Object Pose Estimation to Grasp Estimation for Parallel Grippers: A Review} {Vision-based robotic grasping from object localization, object pose estimation to grasp estimation for parallel grippers: A review}.{\BBCQ}
\newblock
\APACjournalVolNumPages{Artificial Intelligence Review}{54}{3}{1677--1734,}
\newblock
\begin{APACrefDOI} \doi{10.1007/s10462-020-09888-5} \end{APACrefDOI}
\newblock

\newblock

\PrintBackRefs{\CurrentBib}

\bibitem [\protect \citeauthoryear {%
Dwibedi%
, Misra%
\BCBL {}\ \BBA {} Hebert%
}{%
Dwibedi%
\ \protect \BOthers {.}}{%
{\protect \APACyear {2017}}%
}]{%
dwibediCutPasteLearn2017}
\APACinsertmetastar {%
dwibediCutPasteLearn2017}%
\begin{APACrefauthors}%
Dwibedi, D.%
, Misra, I.%
\BCBL {} Hebert, M.%
\end{APACrefauthors}%
\unskip\
\newblock
\APACrefYearMonthDay{2017}{{\APACmonth{10}}}{}.
\newblock
{\BBOQ}\APACrefatitle {Cut, {{Paste}} and {{Learn}}: {{Surprisingly Easy Synthesis}} for {{Instance Detection}}} {Cut, {{Paste}} and {{Learn}}: {{Surprisingly Easy Synthesis}} for {{Instance Detection}}}.{\BBCQ}
\newblock
 \APACrefbtitle {2017 {{IEEE International Conference}} on {{Computer Vision}} ({{ICCV}})} {2017 {{IEEE International Conference}} on {{Computer Vision}} ({{ICCV}})}\ (\BPGS\ 1310--1319).
\newblock
\APACaddressPublisher{{Venice}}{{IEEE}}.
\PrintBackRefs{\CurrentBib}

\bibitem [\protect \citeauthoryear {%
Felzenszwalb%
\ \BBA {} Huttenlocher%
}{%
Felzenszwalb%
\ \BBA {} Huttenlocher%
}{%
{\protect \APACyear {2004}}%
}]{%
felzenszwalbEfficientGraphBasedImage2004}
\APACinsertmetastar {%
felzenszwalbEfficientGraphBasedImage2004}%
\begin{APACrefauthors}%
Felzenszwalb, P.F.%
\BCBT {}\ \BBA {} Huttenlocher, D.P.%
\end{APACrefauthors}%
\unskip\
\newblock
\APACrefYearMonthDay{2004}{{\APACmonth{09}}}{}.
\newblock
{\BBOQ}\APACrefatitle {Efficient {{Graph-Based Image Segmentation}}} {Efficient {{Graph-Based Image Segmentation}}}.{\BBCQ}
\newblock
\APACjournalVolNumPages{International Journal of Computer Vision}{59}{2}{167--181,}
\newblock
\begin{APACrefDOI} \doi{10/fdmw8q} \end{APACrefDOI}
\newblock

\newblock

\PrintBackRefs{\CurrentBib}

\bibitem [\protect \citeauthoryear {%
Feng%
, Wang%
, Zhou%
, Deng%
\BCBL {}\ \BBA {} Li%
}{%
Feng%
\ \protect \BOthers {.}}{%
{\protect \APACyear {2021}}%
}]{%
fengDocTrDocumentImage2021}
\APACinsertmetastar {%
fengDocTrDocumentImage2021}%
\begin{APACrefauthors}%
Feng, H.%
, Wang, Y.%
, Zhou, W.%
, Deng, J.%
\BCBL {} Li, H.%
\end{APACrefauthors}%
\unskip\
\newblock
\APACrefYearMonthDay{2021}{{\APACmonth{10}}}{}.
\newblock
{\BBOQ}\APACrefatitle {{{DocTr}}: {{Document Image Transformer}} for {{Geometric Unwarping}} and {{Illumination Correction}}} {{{DocTr}}: {{Document Image Transformer}} for {{Geometric Unwarping}} and {{Illumination Correction}}}.{\BBCQ}
\newblock
 \APACrefbtitle {Proceedings of the 29th {{ACM International Conference}} on {{Multimedia}}} {Proceedings of the 29th {{ACM International Conference}} on {{Multimedia}}}\ (\BPGS\ 273--281).
\newblock
\APACaddressPublisher{{New York, NY, USA}}{{Association for Computing Machinery}}.
\PrintBackRefs{\CurrentBib}

\bibitem [\protect \citeauthoryear {%
Fischler%
\ \BBA {} Bolles%
}{%
Fischler%
\ \BBA {} Bolles%
}{%
{\protect \APACyear {1987}}%
}]{%
fischlerRandomSampleConsensus1987}
\APACinsertmetastar {%
fischlerRandomSampleConsensus1987}%
\begin{APACrefauthors}%
Fischler, M.A.%
\BCBT {}\ \BBA {} Bolles, R.C.%
\end{APACrefauthors}%
\unskip\
\newblock
\APACrefYearMonthDay{1987}{}{}.
\newblock
{\BBOQ}\APACrefatitle {Random {{Sample Consensus}}: {{A Paradigm}} for {{Model Fitting}} with {{Applications}} to {{Image Analysis}} and {{Automated Cartography}}} {Random {{Sample Consensus}}: {{A Paradigm}} for {{Model Fitting}} with {{Applications}} to {{Image Analysis}} and {{Automated Cartography}}}.{\BBCQ}
\newblock
 \APACrefbtitle {Readings in {{Computer Vision}}} {Readings in {{Computer Vision}}}\ (\BPGS\ 726--740).
\newblock
\APACaddressPublisher{}{{Elsevier}}.
\PrintBackRefs{\CurrentBib}

\bibitem [\protect \citeauthoryear {%
Fontana%
, Zarotti%
\BCBL {}\ \BBA {} Lodi~Rizzini%
}{%
Fontana%
\ \protect \BOthers {.}}{%
{\protect \APACyear {2021}}%
}]{%
fontanaComparativeAssessmentParcel2021}
\APACinsertmetastar {%
fontanaComparativeAssessmentParcel2021}%
\begin{APACrefauthors}%
Fontana, E.%
, Zarotti, W.%
\BCBL {} Lodi~Rizzini, D.%
\end{APACrefauthors}%
\unskip\
\newblock
\APACrefYearMonthDay{2021}{{\APACmonth{08}}}{}.
\newblock
{\BBOQ}\APACrefatitle {A {{Comparative Assessment}} of {{Parcel Box Detection Algorithms}} for {{Industrial Applications}}} {A {{Comparative Assessment}} of {{Parcel Box Detection Algorithms}} for {{Industrial Applications}}}.{\BBCQ}
\newblock
 \APACrefbtitle {2021 {{European Conference}} on {{Mobile Robots}} ({{ECMR}})} {2021 {{European Conference}} on {{Mobile Robots}} ({{ECMR}})}\ (\BPGS\ 1--6).
\PrintBackRefs{\CurrentBib}

\bibitem [\protect \citeauthoryear {%
Fottner%
\ \protect \BOthers {.}}{%
Fottner%
\ \protect \BOthers {.}}{%
{\protect \APACyear {2021}}%
}]{%
fottnerAutonomousSystemsIntralogistics2021}
\APACinsertmetastar {%
fottnerAutonomousSystemsIntralogistics2021}%
\begin{APACrefauthors}%
Fottner, J.%
, Clauer, D.%
, Hormes, F.%
, Freitag, M.%
, Beinke, T.%
, Overmeyer, L.%
\BDBL {}Thomas, F.%
\end{APACrefauthors}%
\unskip\
\newblock
\APACrefYear{2021}.
\newblock
\APACrefbtitle {Autonomous {{Systems}} in {{Intralogistics}} \textendash{} {{State}} of the {{Art}} and {{Future Research Challenges}}} {Autonomous {{Systems}} in {{Intralogistics}} \textendash{} {{State}} of the {{Art}} and {{Future Research Challenges}}}\ (\PrintOrdinal{Second}\ \BEd).
\newblock
\APACaddressPublisher{{DE}}{{Bundesvereinigung Logistik (BVL) e.V.}}
\PrintBackRefs{\CurrentBib}

\bibitem [\protect \citeauthoryear {%
Fragapane%
, {de Koster}%
, Sgarbossa%
\BCBL {}\ \BBA {} Strandhagen%
}{%
Fragapane%
\ \protect \BOthers {.}}{%
{\protect \APACyear {2021}}%
}]{%
fragapanePlanningControlAutonomous2021}
\APACinsertmetastar {%
fragapanePlanningControlAutonomous2021}%
\begin{APACrefauthors}%
Fragapane, G.%
, {de Koster}, R.%
, Sgarbossa, F.%
\BCBL {} Strandhagen, J.O.%
\end{APACrefauthors}%
\unskip\
\newblock
\APACrefYearMonthDay{2021}{{\APACmonth{10}}}{}.
\newblock
{\BBOQ}\APACrefatitle {Planning and Control of Autonomous Mobile Robots for Intralogistics: {{Literature}} Review and Research Agenda} {Planning and control of autonomous mobile robots for intralogistics: {{Literature}} review and research agenda}.{\BBCQ}
\newblock
\APACjournalVolNumPages{European Journal of Operational Research}{294}{2}{405--426,}
\newblock
\begin{APACrefDOI} \doi{10.1016/j.ejor.2021.01.019} \end{APACrefDOI}
\newblock

\newblock

\PrintBackRefs{\CurrentBib}

\bibitem [\protect \citeauthoryear {%
Freund%
\ \BBA {} Schapire%
}{%
Freund%
\ \BBA {} Schapire%
}{%
{\protect \APACyear {1995}}%
}]{%
freundDesiciontheoreticGeneralizationOnline1995}
\APACinsertmetastar {%
freundDesiciontheoreticGeneralizationOnline1995}%
\begin{APACrefauthors}%
Freund, Y.%
\BCBT {}\ \BBA {} Schapire, R.E.%
\end{APACrefauthors}%
\unskip\
\newblock
\APACrefYearMonthDay{1995}{}{}.
\newblock
{\BBOQ}\APACrefatitle {A Desicion-Theoretic Generalization of on-Line Learning and an Application to Boosting} {A desicion-theoretic generalization of on-line learning and an application to boosting}.{\BBCQ}
\newblock
 P.~Vit{\'a}nyi\ (\BED), \APACrefbtitle {Computational {{Learning Theory}}} {Computational {{Learning Theory}}}\ (\BPGS\ 23--37).
\newblock
\APACaddressPublisher{{Berlin, Heidelberg}}{{Springer}}.
\PrintBackRefs{\CurrentBib}

\bibitem [\protect \citeauthoryear {%
Garibotto%
\ \protect \BOthers {.}}{%
Garibotto%
\ \protect \BOthers {.}}{%
{\protect \APACyear {1998}}%
}]{%
garibottoIndustrialExploitationComputer1998}
\APACinsertmetastar {%
garibottoIndustrialExploitationComputer1998}%
\begin{APACrefauthors}%
Garibotto, G.%
, Masciangelo, S.%
, Bassino, P.%
, Coelho, C.%
, Pavan, A.%
\BCBL {} Marson, M.%
\end{APACrefauthors}%
\unskip\
\newblock
\APACrefYearMonthDay{1998}{}{}.
\newblock
{\BBOQ}\APACrefatitle {Industrial Exploitation of Computer Vision in Logistic Automation: Autonomous Control of an Intelligent Forklift Truck} {Industrial exploitation of computer vision in logistic automation: Autonomous control of an intelligent forklift truck}.{\BBCQ}
\newblock
 \APACrefbtitle {Proceedings. 1998 {{IEEE International Conference}} on {{Robotics}} and {{Automation}} ({{Cat}}. {{No}}.{{98CH36146}})} {Proceedings. 1998 {{IEEE International Conference}} on {{Robotics}} and {{Automation}} ({{Cat}}. {{No}}.{{98CH36146}})}\ (\BVOL~2, \BPGS\ 1459--1464).
\newblock
\APACaddressPublisher{{Leuven, Belgium}}{{IEEE}}.
\PrintBackRefs{\CurrentBib}

\bibitem [\protect \citeauthoryear {%
Ge%
, Liu%
, Wang%
, Li%
\BCBL {}\ \BBA {} Sun%
}{%
Ge%
\ \protect \BOthers {.}}{%
{\protect \APACyear {2021}}%
}]{%
geYOLOXExceedingYOLO2021}
\APACinsertmetastar {%
geYOLOXExceedingYOLO2021}%
\begin{APACrefauthors}%
Ge, Z.%
, Liu, S.%
, Wang, F.%
, Li, Z.%
\BCBL {} Sun, J.%
\end{APACrefauthors}%
\unskip\
\newblock
\APACrefYearMonthDay{2021}{{\APACmonth{08}}}{}.
\newblock
\APACrefbtitle {{{YOLOX}}: {{Exceeding YOLO Series}} in 2021} {{{YOLOX}}: {{Exceeding YOLO Series}} in 2021}\ (\BNUM\ arXiv:2107.08430).
\newblock
\APACaddressPublisher{}{{arXiv}}.
\PrintBackRefs{\CurrentBib}

\bibitem [\protect \citeauthoryear {%
Glockner%
, Jannek%
, Mahn%
\BCBL {}\ \BBA {} Theis%
}{%
Glockner%
\ \protect \BOthers {.}}{%
{\protect \APACyear {2020}}%
}]{%
glocknerAugmentedRealityLogistics2020}
\APACinsertmetastar {%
glocknerAugmentedRealityLogistics2020}%
\begin{APACrefauthors}%
Glockner, H.%
, Jannek, K.%
, Mahn, J.%
\BCBL {} Theis, B.%
\end{APACrefauthors}%
\unskip\
\newblock
\APACrefYearMonthDay{2020}{}{}.
\newblock
\APACrefbtitle {Augmented {{Reality}} in {{Logistics}}: {{Changing}} the {{Way}} We See {{Logistics}} - {{A DHL Perspective}}} {Augmented {{Reality}} in {{Logistics}}: {{Changing}} the {{Way}} we see {{Logistics}} - {{A DHL Perspective}}}\ \APACbVolEdTR{}{\BTR{}}.
\PrintBackRefs{\CurrentBib}

\bibitem [\protect \citeauthoryear {%
Goodfellow%
\ \protect \BOthers {.}}{%
Goodfellow%
\ \protect \BOthers {.}}{%
{\protect \APACyear {2014}}%
}]{%
goodfellowGenerativeAdversarialNets2014}
\APACinsertmetastar {%
goodfellowGenerativeAdversarialNets2014}%
\begin{APACrefauthors}%
Goodfellow, I.%
, {Pouget-Abadie}, J.%
, Mirza, M.%
, Xu, B.%
, {Warde-Farley}, D.%
, Ozair, S.%
\BDBL {}Bengio, Y.%
\end{APACrefauthors}%
\unskip\
\newblock
\APACrefYearMonthDay{2014}{}{}.
\newblock
{\BBOQ}\APACrefatitle {Generative Adversarial Nets} {Generative adversarial nets}.{\BBCQ}
\newblock
 Z.~Ghahramani, M.~Welling, C.~Cortes, N.~Lawrence\BCBL {}\ \BBA {} K.~Weinberger\ (\BEDS), \APACrefbtitle {Advances in Neural Information Processing Systems} {Advances in neural information processing systems}\ (\BVOL~27).
\newblock
\APACaddressPublisher{}{{Curran Associates, Inc.}}
\PrintBackRefs{\CurrentBib}

\bibitem [\protect \citeauthoryear {%
Grilli%
, Menna%
\BCBL {}\ \BBA {} Remondino%
}{%
Grilli%
\ \protect \BOthers {.}}{%
{\protect \APACyear {2017}}%
}]{%
grilliReviewPointClouds2017}
\APACinsertmetastar {%
grilliReviewPointClouds2017}%
\begin{APACrefauthors}%
Grilli, E.%
, Menna, F.%
\BCBL {} Remondino, F.%
\end{APACrefauthors}%
\unskip\
\newblock
\APACrefYearMonthDay{2017}{{\APACmonth{02}}}{}.
\newblock
{\BBOQ}\APACrefatitle {A Review of Point Clouds Segmentation and Classification Algorithms} {A review of point clouds segmentation and classification algorithms}.{\BBCQ}
\newblock
\APACjournalVolNumPages{ISPRS - International Archives of the Photogrammetry, Remote Sensing and Spatial Information Sciences}{XLII-2/W3}{}{339--344,}
\newblock
\begin{APACrefDOI} \doi{10.5194/isprs-archives-XLII-2-W3-339-2017} \end{APACrefDOI}
\newblock

\newblock

\PrintBackRefs{\CurrentBib}

\bibitem [\protect \citeauthoryear {%
Grzeszick%
, Feldhorst%
, Mosblech%
, Fink%
\BCBL {}\ \BBA {} Ten~Hompel%
}{%
Grzeszick%
\ \protect \BOthers {.}}{%
{\protect \APACyear {2016}}%
}]{%
grzeszickCameraassistedPickbyfeel2016}
\APACinsertmetastar {%
grzeszickCameraassistedPickbyfeel2016}%
\begin{APACrefauthors}%
Grzeszick, R.%
, Feldhorst, S.%
, Mosblech, C.%
, Fink, G.A.%
\BCBL {} Ten~Hompel, M.%
\end{APACrefauthors}%
\unskip\
\newblock
\APACrefYearMonthDay{2016}{}{}.
\newblock
{\BBOQ}\APACrefatitle {Camera-Assisted {{Pick-by-feel}}} {Camera-assisted {{Pick-by-feel}}}.{\BBCQ}
\newblock
\APACjournalVolNumPages{Logistics Journal : Proceedings}{2016}{}{,}
\newblock
\begin{APACrefDOI} \doi{10.2195/lj_proc_grzeszick_en_201610_01} \end{APACrefDOI}
\newblock

\newblock

\PrintBackRefs{\CurrentBib}

\bibitem [\protect \citeauthoryear {%
Haanpaa%
, Beach%
\BCBL {}\ \BBA {} Cohen%
}{%
Haanpaa%
\ \protect \BOthers {.}}{%
{\protect \APACyear {2016}}%
}]{%
haanpaaMachineVisionAlgorithms2016}
\APACinsertmetastar {%
haanpaaMachineVisionAlgorithms2016}%
\begin{APACrefauthors}%
Haanpaa, D.%
, Beach, G.%
\BCBL {} Cohen, C.J.%
\end{APACrefauthors}%
\unskip\
\newblock
\APACrefYearMonthDay{2016}{{\APACmonth{10}}}{}.
\newblock
{\BBOQ}\APACrefatitle {Machine Vision Algorithms for Robust Pallet Engagement and Stacking} {Machine vision algorithms for robust pallet engagement and stacking}.{\BBCQ}
\newblock
 \APACrefbtitle {2016 {{IEEE Applied Imagery Pattern Recognition Workshop}} ({{AIPR}})} {2016 {{IEEE Applied Imagery Pattern Recognition Workshop}} ({{AIPR}})}\ (\BPGS\ 1--8).
\newblock
\APACaddressPublisher{{Washington, DC, USA}}{{IEEE}}.
\PrintBackRefs{\CurrentBib}

\bibitem [\protect \citeauthoryear {%
He%
, Gkioxari%
, Dollar%
\BCBL {}\ \BBA {} Girshick%
}{%
He%
\ \protect \BOthers {.}}{%
{\protect \APACyear {2017}}%
}]{%
heMaskRCNN2017}
\APACinsertmetastar {%
heMaskRCNN2017}%
\begin{APACrefauthors}%
He, K.%
, Gkioxari, G.%
, Dollar, P.%
\BCBL {} Girshick, R.%
\end{APACrefauthors}%
\unskip\
\newblock
\APACrefYearMonthDay{2017}{{\APACmonth{10}}}{}.
\newblock
{\BBOQ}\APACrefatitle {Mask {{R-CNN}}} {Mask {{R-CNN}}}.{\BBCQ}
\newblock
 \APACrefbtitle {{{IEEE International Conference}} on {{Computer Vision}} ({{ICCV}})} {{{IEEE International Conference}} on {{Computer Vision}} ({{ICCV}})}\ (\BPGS\ 2980--2988).
\newblock
\APACaddressPublisher{{Venice}}{{IEEE}}.
\PrintBackRefs{\CurrentBib}

\bibitem [\protect \citeauthoryear {%
He%
, Zhang%
, Ren%
\BCBL {}\ \BBA {} Sun%
}{%
He%
\ \protect \BOthers {.}}{%
{\protect \APACyear {2016}}%
}]{%
heDeepResidualLearning2016}
\APACinsertmetastar {%
heDeepResidualLearning2016}%
\begin{APACrefauthors}%
He, K.%
, Zhang, X.%
, Ren, S.%
\BCBL {} Sun, J.%
\end{APACrefauthors}%
\unskip\
\newblock
\APACrefYearMonthDay{2016}{{\APACmonth{06}}}{}.
\newblock
{\BBOQ}\APACrefatitle {Deep {{Residual Learning}} for {{Image Recognition}}} {Deep {{Residual Learning}} for {{Image Recognition}}}.{\BBCQ}
\newblock
 \APACrefbtitle {{{IEEE Conference}} on {{Computer Vision}} and {{Pattern Recognition}} ({{CVPR}})} {{{IEEE Conference}} on {{Computer Vision}} and {{Pattern Recognition}} ({{CVPR}})}\ (\BPGS\ 770--778).
\newblock
\APACaddressPublisher{{Las Vegas, NV, USA}}{{IEEE}}.
\PrintBackRefs{\CurrentBib}

\bibitem [\protect \citeauthoryear {%
Himstedt%
\ \BBA {} Maehle%
}{%
Himstedt%
\ \BBA {} Maehle%
}{%
{\protect \APACyear {2017}}%
}]{%
himstedtOnlineSemanticMapping2017}
\APACinsertmetastar {%
himstedtOnlineSemanticMapping2017}%
\begin{APACrefauthors}%
Himstedt, M.%
\BCBT {}\ \BBA {} Maehle, E.%
\end{APACrefauthors}%
\unskip\
\newblock
\APACrefYearMonthDay{2017}{{\APACmonth{07}}}{}.
\newblock
{\BBOQ}\APACrefatitle {Online Semantic Mapping of Logistic Environments Using {{RGB-D}} Cameras:} {Online semantic mapping of logistic environments using {{RGB-D}} cameras:}.{\BBCQ}
\newblock
\APACjournalVolNumPages{International Journal of Advanced Robotic Systems}{}{}{,}
\newblock
\begin{APACrefDOI} \doi{10/gcd6jq} \end{APACrefDOI}
\newblock

\newblock

\PrintBackRefs{\CurrentBib}

\bibitem [\protect \citeauthoryear {%
Hochstein%
, Gl{\"o}ckle%
, Meyer%
\BCBL {}\ \BBA {} Furmans%
}{%
Hochstein%
\ \protect \BOthers {.}}{%
{\protect \APACyear {2016}}%
}]{%
hochsteinPackassistentAssistenzsystemFuer2016}
\APACinsertmetastar {%
hochsteinPackassistentAssistenzsystemFuer2016}%
\begin{APACrefauthors}%
Hochstein, M.%
, Gl{\"o}ckle, J.%
, Meyer, T.%
\BCBL {} Furmans, K.%
\end{APACrefauthors}%
\unskip\
\newblock
\APACrefYearMonthDay{2016}{}{}.
\newblock
{\BBOQ}\APACrefatitle {{Packassistent \textendash{} Assistenzsystem f\"ur die Qualit\"atskontrolle w\"ahrend des Packprozesses}} {{Packassistent \textendash{} Assistenzsystem f\"ur die Qualit\"atskontrolle w\"ahrend des Packprozesses}}.{\BBCQ}
\newblock
\APACjournalVolNumPages{Logistics Journal : Proceedings}{2016}{}{,}
\newblock
\begin{APACrefDOI} \doi{10.2195/lj_proc_hochstein_de_201610_01} \end{APACrefDOI}
\newblock

\newblock

\PrintBackRefs{\CurrentBib}

\bibitem [\protect \citeauthoryear {%
Hough%
}{%
Hough%
}{%
{\protect \APACyear {1962}}%
}]{%
houghMethodMeansRecognizing1962}
\APACinsertmetastar {%
houghMethodMeansRecognizing1962}%
\begin{APACrefauthors}%
Hough, P.V.C.%
\end{APACrefauthors}%
\unskip\
\newblock
\APACrefYearMonthDay{1962}{{\APACmonth{12}}}{}.
\newblock
\APACrefbtitle {Method and Means for Recognizing Complex Patterns} {Method and means for recognizing complex patterns}\ (\BNUM\ US3069654A).
\PrintBackRefs{\CurrentBib}

\bibitem [\protect \citeauthoryear {%
Hu%
, Immel%
, Janosovits%
, Lauer%
\BCBL {}\ \BBA {} Stiller%
}{%
Hu%
\ \protect \BOthers {.}}{%
{\protect \APACyear {2021}}%
}]{%
huCuboidDetectionTracking2021}
\APACinsertmetastar {%
huCuboidDetectionTracking2021}%
\begin{APACrefauthors}%
Hu, H.%
, Immel, F.%
, Janosovits, J.%
, Lauer, M.%
\BCBL {} Stiller, C.%
\end{APACrefauthors}%
\unskip\
\newblock
\APACrefYearMonthDay{2021}{{\APACmonth{08}}}{}.
\newblock
{\BBOQ}\APACrefatitle {A {{Cuboid Detection}} and {{Tracking System}} Using {{A Multi RGBD Camera Setup}} for {{Intelligent Manipulation}} and {{Logistics}}} {A {{Cuboid Detection}} and {{Tracking System}} using {{A Multi RGBD Camera Setup}} for {{Intelligent Manipulation}} and {{Logistics}}}.{\BBCQ}
\newblock
 \APACrefbtitle {2021 {{IEEE}} 17th {{International Conference}} on {{Automation Science}} and {{Engineering}} ({{CASE}})} {2021 {{IEEE}} 17th {{International Conference}} on {{Automation Science}} and {{Engineering}} ({{CASE}})}\ (\BPGS\ 1097--1103).
\PrintBackRefs{\CurrentBib}

\bibitem [\protect \citeauthoryear {%
Jiang%
\ \protect \BOthers {.}}{%
Jiang%
\ \protect \BOthers {.}}{%
{\protect \APACyear {2022}}%
}]{%
jiangRevisitingDocumentImage2022}
\APACinsertmetastar {%
jiangRevisitingDocumentImage2022}%
\begin{APACrefauthors}%
Jiang, X.%
, Long, R.%
, Xue, N.%
, Yang, Z.%
, Yao, C.%
\BCBL {} Xia, G\BHBI S.%
\end{APACrefauthors}%
\unskip\
\newblock
\APACrefYearMonthDay{2022}{{\APACmonth{06}}}{}.
\newblock
{\BBOQ}\APACrefatitle {Revisiting {{Document Image Dewarping}} by {{Grid Regularization}}} {Revisiting {{Document Image Dewarping}} by {{Grid Regularization}}}.{\BBCQ}
\newblock
 \APACrefbtitle {2022 {{IEEE}}/{{CVF Conference}} on {{Computer Vision}} and {{Pattern Recognition}} ({{CVPR}})} {2022 {{IEEE}}/{{CVF Conference}} on {{Computer Vision}} and {{Pattern Recognition}} ({{CVPR}})}\ (\BPGS\ 4533--4542).
\newblock
\APACaddressPublisher{{New Orleans, LA, USA}}{{IEEE}}.
\PrintBackRefs{\CurrentBib}

\bibitem [\protect \citeauthoryear {%
Jocher%
\ \protect \BOthers {.}}{%
Jocher%
\ \protect \BOthers {.}}{%
{\protect \APACyear {2022}}%
}]{%
jocherUltralyticsYolov5V72022}
\APACinsertmetastar {%
jocherUltralyticsYolov5V72022}%
\begin{APACrefauthors}%
Jocher, G.%
, Chaurasia, A.%
, Stoken, A.%
, Borovec, J.%
, NanoCode012%
, Kwon, Y.%
\BDBL {}Jain, M.%
\end{APACrefauthors}%
\unskip\
\newblock
\APACrefYearMonthDay{2022}{{\APACmonth{11}}}{}.
\newblock
\APACrefbtitle {Ultralytics/Yolov5: V7.0 - {{YOLOv5 SOTA Realtime Instance Segmentation}}.} {Ultralytics/yolov5: V7.0 - {{YOLOv5 SOTA Realtime Instance Segmentation}}.}
\newblock
\APAChowpublished {Zenodo}.
\PrintBackRefs{\CurrentBib}

\bibitem [\protect \citeauthoryear {%
Kalman%
}{%
Kalman%
}{%
{\protect \APACyear {1960}}%
}]{%
kalmanNewApproachLinear1960}
\APACinsertmetastar {%
kalmanNewApproachLinear1960}%
\begin{APACrefauthors}%
Kalman, R.E.%
\end{APACrefauthors}%
\unskip\
\newblock
\APACrefYearMonthDay{1960}{}{}.
\newblock
{\BBOQ}\APACrefatitle {A New Approach to Linear Filtering and Prediction Problems} {A new approach to linear filtering and prediction problems}.{\BBCQ}
\newblock
\APACjournalVolNumPages{Transactions of the ASME\textendash Journal of Basic Engineering}{82}{Series D}{35--45,}
\newblock

\newblock

\PrintBackRefs{\CurrentBib}

\bibitem [\protect \citeauthoryear {%
Kamnardsiri%
, Charoenkwan%
, Malang%
\BCBL {}\ \BBA {} Wudhikarn%
}{%
Kamnardsiri%
\ \protect \BOthers {.}}{%
{\protect \APACyear {2022}}%
}]{%
kamnardsiri1DBarcodeDetection2022}
\APACinsertmetastar {%
kamnardsiri1DBarcodeDetection2022}%
\begin{APACrefauthors}%
Kamnardsiri, T.%
, Charoenkwan, P.%
, Malang, C.%
\BCBL {} Wudhikarn, R.%
\end{APACrefauthors}%
\unskip\
\newblock
\APACrefYearMonthDay{2022}{{\APACmonth{11}}}{}.
\newblock
{\BBOQ}\APACrefatitle {{{1D Barcode Detection}}: {{Novel Benchmark Datasets}} and {{Comprehensive Comparison}} of {{Deep Convolutional Neural Network Approaches}}} {{{1D Barcode Detection}}: {{Novel Benchmark Datasets}} and {{Comprehensive Comparison}} of {{Deep Convolutional Neural Network Approaches}}}.{\BBCQ}
\newblock
\APACjournalVolNumPages{Sensors}{22}{22}{8788,}
\newblock
\begin{APACrefDOI} \doi{10.3390/s22228788} \end{APACrefDOI}
\newblock

\newblock

\PrintBackRefs{\CurrentBib}

\bibitem [\protect \citeauthoryear {%
Kersten%
, Seiter%
, von See%
, Hackius%
\BCBL {}\ \BBA {} Maurer%
}{%
Kersten%
\ \protect \BOthers {.}}{%
{\protect \APACyear {2017}}%
}]{%
kerstenChancenDigitalenTransformation2017}
\APACinsertmetastar {%
kerstenChancenDigitalenTransformation2017}%
\begin{APACrefauthors}%
Kersten, W.%
, Seiter, M.%
, von See, B.%
, Hackius, N.%
\BCBL {} Maurer, T.%
\end{APACrefauthors}%
\unskip\
\newblock
\APACrefYear{2017}.
\newblock
\APACrefbtitle {{Chancen der digitalen Transformation: Trends und Strategien in Logistik und Supply Chain Management}} {{Chancen der digitalen Transformation: Trends und Strategien in Logistik und Supply Chain Management}}.
\newblock
\APACaddressPublisher{{Hamburg}}{{DVV Media Group GmbH}}.
\PrintBackRefs{\CurrentBib}

\bibitem [\protect \citeauthoryear {%
M.S.U.~Khan%
, Pagani%
, Liwicki%
, Stricker%
\BCBL {}\ \BBA {} Afzal%
}{%
M.S.U.~Khan%
\ \protect \BOthers {.}}{%
{\protect \APACyear {2022}}%
}]{%
khanThreeDimensionalReconstructionSingle2022}
\APACinsertmetastar {%
khanThreeDimensionalReconstructionSingle2022}%
\begin{APACrefauthors}%
Khan, M.S.U.%
, Pagani, A.%
, Liwicki, M.%
, Stricker, D.%
\BCBL {} Afzal, M.Z.%
\end{APACrefauthors}%
\unskip\
\newblock
\APACrefYearMonthDay{2022}{{\APACmonth{09}}}{}.
\newblock
{\BBOQ}\APACrefatitle {Three-{{Dimensional Reconstruction}} from a {{Single RGB Image Using Deep Learning}}: {{A Review}}} {Three-{{Dimensional Reconstruction}} from a {{Single RGB Image Using Deep Learning}}: {{A Review}}}.{\BBCQ}
\newblock
\APACjournalVolNumPages{Journal of Imaging}{8}{9}{225,}
\newblock
\begin{APACrefDOI} \doi{10.3390/jimaging8090225} \end{APACrefDOI}
\newblock

\newblock

\PrintBackRefs{\CurrentBib}

\bibitem [\protect \citeauthoryear {%
S.D.~Khan%
\ \BBA {} Ullah%
}{%
S.D.~Khan%
\ \BBA {} Ullah%
}{%
{\protect \APACyear {2019}}%
}]{%
khanSurveyAdvancesVisionbased2019}
\APACinsertmetastar {%
khanSurveyAdvancesVisionbased2019}%
\begin{APACrefauthors}%
Khan, S.D.%
\BCBT {}\ \BBA {} Ullah, H.%
\end{APACrefauthors}%
\unskip\
\newblock
\APACrefYearMonthDay{2019}{{\APACmonth{05}}}{}.
\newblock
{\BBOQ}\APACrefatitle {A Survey of Advances in Vision-Based Vehicle Re-Identification} {A survey of advances in vision-based vehicle re-identification}.{\BBCQ}
\newblock
\APACjournalVolNumPages{Computer Vision and Image Understanding}{182}{}{50--63,}
\newblock
\begin{APACrefDOI} \doi{10.1016/j.cviu.2019.03.001} \end{APACrefDOI}
\newblock

\newblock

\PrintBackRefs{\CurrentBib}

\bibitem [\protect \citeauthoryear {%
Kleeberger%
, Bormann%
, Kraus%
\BCBL {}\ \BBA {} Huber%
}{%
Kleeberger%
\ \protect \BOthers {.}}{%
{\protect \APACyear {2020}}%
}]{%
kleebergerSurveyLearningBasedRobotic2020}
\APACinsertmetastar {%
kleebergerSurveyLearningBasedRobotic2020}%
\begin{APACrefauthors}%
Kleeberger, K.%
, Bormann, R.%
, Kraus, W.%
\BCBL {} Huber, M.F.%
\end{APACrefauthors}%
\unskip\
\newblock
\APACrefYearMonthDay{2020}{{\APACmonth{12}}}{}.
\newblock
{\BBOQ}\APACrefatitle {A {{Survey}} on {{Learning-Based Robotic Grasping}}} {A {{Survey}} on {{Learning-Based Robotic Grasping}}}.{\BBCQ}
\newblock
\APACjournalVolNumPages{Current Robotics Reports}{1}{4}{239--249,}
\newblock
\begin{APACrefDOI} \doi{10.1007/s43154-020-00021-6} \end{APACrefDOI}
\newblock

\newblock

\PrintBackRefs{\CurrentBib}

\bibitem [\protect \citeauthoryear {%
Kl{\"u}ttermann%
, Rutinowski%
, Reining%
, Roidl%
\BCBL {}\ \BBA {} M{\"u}ller%
}{%
Kl{\"u}ttermann%
\ \protect \BOthers {.}}{%
{\protect \APACyear {2022}}%
}]{%
kluttermannGraphRepresentationBased2022}
\APACinsertmetastar {%
kluttermannGraphRepresentationBased2022}%
\begin{APACrefauthors}%
Kl{\"u}ttermann, S.%
, Rutinowski, J.%
, Reining, C.%
, Roidl, M.%
\BCBL {} M{\"u}ller, E.%
\end{APACrefauthors}%
\unskip\
\newblock
\APACrefYearMonthDay{2022}{{\APACmonth{12}}}{}.
\newblock
{\BBOQ}\APACrefatitle {Towards {{Graph Representation}} Based {{Re-Identification}} of {{Chipwood Pallet Blocks}}} {Towards {{Graph Representation}} based {{Re-Identification}} of {{Chipwood Pallet Blocks}}}.{\BBCQ}
\newblock
 \APACrefbtitle {2022 21st {{IEEE International Conference}} on {{Machine Learning}} and {{Applications}} ({{ICMLA}})} {2022 21st {{IEEE International Conference}} on {{Machine Learning}} and {{Applications}} ({{ICMLA}})}\ (\BPGS\ 1543--1550).
\PrintBackRefs{\CurrentBib}

\bibitem [\protect \citeauthoryear {%
K{\"u}ckelhaus%
}{%
K{\"u}ckelhaus%
}{%
{\protect \APACyear {2013}}%
}]{%
kuckelhausDHLLowcostSensor2013}
\APACinsertmetastar {%
kuckelhausDHLLowcostSensor2013}%
\begin{APACrefauthors}%
K{\"u}ckelhaus, M.%
\end{APACrefauthors}%
\unskip\
\newblock
\APACrefYearMonthDay{2013}{}{}.
\newblock
\APACrefbtitle {{{DHL}} - {{Low-cost}} Sensor Technology} {{{DHL}} - {{Low-cost}} sensor technology}\ [{{DHL Customer Solutions}} \& {{Innovation}}].
\newblock
{Troisdorf, Germany}.
\PrintBackRefs{\CurrentBib}

\bibitem [\protect \citeauthoryear {%
Kucuk%
, Al~Muallim%
, Y{\i}lmaz%
\BCBL {}\ \BBA {} Kahraman%
}{%
Kucuk%
\ \protect \BOthers {.}}{%
{\protect \APACyear {2019}}%
}]{%
kucukDevelopmentDimensionsMeasurement2019}
\APACinsertmetastar {%
kucukDevelopmentDimensionsMeasurement2019}%
\begin{APACrefauthors}%
Kucuk, H.%
, Al~Muallim, M.T.%
, Y{\i}lmaz, F.%
\BCBL {} Kahraman, M.%
\end{APACrefauthors}%
\unskip\
\newblock
\APACrefYearMonthDay{2019}{{\APACmonth{03}}}{}.
\newblock
{\BBOQ}\APACrefatitle {Development of a Dimensions Measurement System Based on Depth Camera for Logistic Applications} {Development of a dimensions measurement system based on depth camera for logistic applications}.{\BBCQ}
\newblock
 D.P.~Nikolaev, P.~Radeva, A.~Verikas\BCBL {}\ \BBA {} J.~Zhou\ (\BEDS), \APACrefbtitle {Eleventh {{International Conference}} on {{Machine Vision}} ({{ICMV}} 2018)} {Eleventh {{International Conference}} on {{Machine Vision}} ({{ICMV}} 2018)}\ (\BPG~93).
\newblock
\APACaddressPublisher{{Munich, Germany}}{{SPIE}}.
\PrintBackRefs{\CurrentBib}

\bibitem [\protect \citeauthoryear {%
Kumbhar%
, Thombare%
\BCBL {}\ \BBA {} Salunkhe%
}{%
Kumbhar%
\ \protect \BOthers {.}}{%
{\protect \APACyear {2018}}%
}]{%
kumbharAutomatedGuidedVehicles2018}
\APACinsertmetastar {%
kumbharAutomatedGuidedVehicles2018}%
\begin{APACrefauthors}%
Kumbhar, S.G.%
, Thombare, R.B.%
\BCBL {} Salunkhe, A.B.%
\end{APACrefauthors}%
\unskip\
\newblock
\APACrefYearMonthDay{2018}{{\APACmonth{09}}}{}.
\newblock
{\BBOQ}\APACrefatitle {Automated {{Guided Vehicles}} for {{Small Manufacturing Enterprises}}: {{A Review}}} {Automated {{Guided Vehicles}} for {{Small Manufacturing Enterprises}}: {{A Review}}}.{\BBCQ}
\newblock
\APACjournalVolNumPages{SAE International Journal of Materials and Manufacturing}{11}{3}{253--258,}
\newblock
\begin{APACrefDOI} \doi{10/ggfhm9} \end{APACrefDOI}
\newblock

\newblock

\PrintBackRefs{\CurrentBib}

\bibitem [\protect \citeauthoryear {%
Laotrakunchai%
, Wongkaew%
\BCBL {}\ \BBA {} Patanukhom%
}{%
Laotrakunchai%
\ \protect \BOthers {.}}{%
{\protect \APACyear {2013}}%
}]{%
laotrakunchaiMeasurementSizeDistance2013}
\APACinsertmetastar {%
laotrakunchaiMeasurementSizeDistance2013}%
\begin{APACrefauthors}%
Laotrakunchai, S.%
, Wongkaew, A.%
\BCBL {} Patanukhom, K.%
\end{APACrefauthors}%
\unskip\
\newblock
\APACrefYearMonthDay{2013}{{\APACmonth{12}}}{}.
\newblock
{\BBOQ}\APACrefatitle {Measurement of {{Size}} and {{Distance}} of {{Objects Using Mobile Devices}}} {Measurement of {{Size}} and {{Distance}} of {{Objects Using Mobile Devices}}}.{\BBCQ}
\newblock
 \APACrefbtitle {2013 {{International Conference}} on {{Signal-Image Technology Internet-Based Systems}}} {2013 {{International Conference}} on {{Signal-Image Technology Internet-Based Systems}}}\ (\BPGS\ 156--161).
\PrintBackRefs{\CurrentBib}

\bibitem [\protect \citeauthoryear {%
Law%
\ \BBA {} Deng%
}{%
Law%
\ \BBA {} Deng%
}{%
{\protect \APACyear {2018}}%
}]{%
lawCornerNetDetectingObjects2018}
\APACinsertmetastar {%
lawCornerNetDetectingObjects2018}%
\begin{APACrefauthors}%
Law, H.%
\BCBT {}\ \BBA {} Deng, J.%
\end{APACrefauthors}%
\unskip\
\newblock
\APACrefYearMonthDay{2018}{}{}.
\newblock
{\BBOQ}\APACrefatitle {{{CornerNet}}: {{Detecting Objects}} as {{Paired Keypoints}}} {{{CornerNet}}: {{Detecting Objects}} as {{Paired Keypoints}}}.{\BBCQ}
\newblock
 \APACrefbtitle {Proceedings of the {{European Conference}} on {{Computer Vision}} ({{ECCV}})} {Proceedings of the {{European Conference}} on {{Computer Vision}} ({{ECCV}})}\ (\BPGS\ 734--750).
\PrintBackRefs{\CurrentBib}

\bibitem [\protect \citeauthoryear {%
J.~Li%
\ \protect \BOthers {.}}{%
J.~Li%
\ \protect \BOthers {.}}{%
{\protect \APACyear {2022}}%
}]{%
liDiTSelfsupervisedPretraining2022}
\APACinsertmetastar {%
liDiTSelfsupervisedPretraining2022}%
\begin{APACrefauthors}%
Li, J.%
, Xu, Y.%
, Lv, T.%
, Cui, L.%
, Zhang, C.%
\BCBL {} Wei, F.%
\end{APACrefauthors}%
\unskip\
\newblock
\APACrefYearMonthDay{2022}{{\APACmonth{03}}}{}.
\newblock
{\BBOQ}\APACrefatitle {{{DiT}}: {{Self-supervised Pre-training}} for {{Document Image Transformer}}} {{{DiT}}: {{Self-supervised Pre-training}} for {{Document Image Transformer}}}.{\BBCQ}
\newblock
\APACjournalVolNumPages{arXiv:2203.02378 [cs]}{}{}{,}
\newblock
{\href{https://arxiv.org/abs/2203.02378}{{arxiv:2203.02378}}}
\newblock
 {[cs]}
\PrintBackRefs{\CurrentBib}

\bibitem [\protect \citeauthoryear {%
P.~Li%
, Zhao%
, Liu%
\BCBL {}\ \BBA {} Cao%
}{%
P.~Li%
\ \protect \BOthers {.}}{%
{\protect \APACyear {2020}}%
}]{%
liRTM3DRealTimeMonocular2020}
\APACinsertmetastar {%
liRTM3DRealTimeMonocular2020}%
\begin{APACrefauthors}%
Li, P.%
, Zhao, H.%
, Liu, P.%
\BCBL {} Cao, F.%
\end{APACrefauthors}%
\unskip\
\newblock
\APACrefYearMonthDay{2020}{}{}.
\newblock
{\BBOQ}\APACrefatitle {{{RTM3D}}: {{Real-Time Monocular 3D Detection}} from {{Object Keypoints}} for {{Autonomous Driving}}} {{{RTM3D}}: {{Real-Time Monocular 3D Detection}} from {{Object Keypoints}} for {{Autonomous Driving}}}.{\BBCQ}
\newblock
 A.~Vedaldi, H.~Bischof, T.~Brox\BCBL {}\ \BBA {} J\BHBI M.~Frahm\ (\BEDS), \APACrefbtitle {Computer {{Vision}} \textendash{} {{ECCV}} 2020} {Computer {{Vision}} \textendash{} {{ECCV}} 2020}\ (\BVOL\ 12348, \BPGS\ 644--660).
\newblock
\APACaddressPublisher{{Cham}}{{Springer International Publishing}}.
\PrintBackRefs{\CurrentBib}

\bibitem [\protect \citeauthoryear {%
X.~Li%
, Chen%
, Thomas%
\BCBL {}\ \BBA {} MacDonald%
}{%
X.~Li%
\ \protect \BOthers {.}}{%
{\protect \APACyear {2012}}%
}]{%
liUsingKinectMonitoring2012}
\APACinsertmetastar {%
liUsingKinectMonitoring2012}%
\begin{APACrefauthors}%
Li, X.%
, Chen, I.Y\BHBI H.%
, Thomas, S.%
\BCBL {} MacDonald, B.A.%
\end{APACrefauthors}%
\unskip\
\newblock
\APACrefYearMonthDay{2012}{}{}.
\newblock
{\BBOQ}\APACrefatitle {Using {{Kinect}} for Monitoring Warehouse Order Picking Operations} {Using {{Kinect}} for monitoring warehouse order picking operations}.{\BBCQ}
\newblock
\APACjournalVolNumPages{Proceedings of Australasian Conference on Robotics and Automation}{}{}{7,}
\newblock

\newblock

\PrintBackRefs{\CurrentBib}

\bibitem [\protect \citeauthoryear {%
Y.~Li%
\ \protect \BOthers {.}}{%
Y.~Li%
\ \protect \BOthers {.}}{%
{\protect \APACyear {2021}}%
}]{%
liComputerVisionBased2021}
\APACinsertmetastar {%
liComputerVisionBased2021}%
\begin{APACrefauthors}%
Li, Y.%
, Niu, Y.%
, Liu, Y.%
, Zheng, L.%
, Wang, Z.%
\BCBL {} Zhe, W.%
\end{APACrefauthors}%
\unskip\
\newblock
\APACrefYearMonthDay{2021}{{\APACmonth{09}}}{}.
\newblock
{\BBOQ}\APACrefatitle {Computer {{Vision Based Conveyor Belt Congestion Recognition}} in {{Logistics Industrial Parks}}} {Computer {{Vision Based Conveyor Belt Congestion Recognition}} in {{Logistics Industrial Parks}}}.{\BBCQ}
\newblock
 \APACrefbtitle {2021 26th {{IEEE International Conference}} on {{Emerging Technologies}} and {{Factory Automation}} ({{ETFA}} )} {2021 26th {{IEEE International Conference}} on {{Emerging Technologies}} and {{Factory Automation}} ({{ETFA}} )}\ (\BPGS\ 1--8).
\PrintBackRefs{\CurrentBib}

\bibitem [\protect \citeauthoryear {%
Lin%
\ \protect \BOthers {.}}{%
Lin%
\ \protect \BOthers {.}}{%
{\protect \APACyear {2017}}%
}]{%
linFeaturePyramidNetworks2017}
\APACinsertmetastar {%
linFeaturePyramidNetworks2017}%
\begin{APACrefauthors}%
Lin, T\BHBI Y.%
, Doll{\'a}r, P.%
, Girshick, R.%
, He, K.%
, Hariharan, B.%
\BCBL {} Belongie, S.%
\end{APACrefauthors}%
\unskip\
\newblock
\APACrefYearMonthDay{2017}{{\APACmonth{07}}}{}.
\newblock
{\BBOQ}\APACrefatitle {Feature {{Pyramid Networks}} for {{Object Detection}}} {Feature {{Pyramid Networks}} for {{Object Detection}}}.{\BBCQ}
\newblock
 \APACrefbtitle {{{IEEE Conference}} on {{Computer Vision}} and {{Pattern Recognition}} ({{CVPR}})} {{{IEEE Conference}} on {{Computer Vision}} and {{Pattern Recognition}} ({{CVPR}})}\ (\BPGS\ 936--944).
\PrintBackRefs{\CurrentBib}

\bibitem [\protect \citeauthoryear {%
Liu%
, Kim%
, Gu%
, Furukawa%
\BCBL {}\ \BBA {} Kautz%
}{%
Liu%
\ \protect \BOthers {.}}{%
{\protect \APACyear {2019}}%
}]{%
liuPlaneRCNN3DPlane2019}
\APACinsertmetastar {%
liuPlaneRCNN3DPlane2019}%
\begin{APACrefauthors}%
Liu, C.%
, Kim, K.%
, Gu, J.%
, Furukawa, Y.%
\BCBL {} Kautz, J.%
\end{APACrefauthors}%
\unskip\
\newblock
\APACrefYearMonthDay{2019}{{\APACmonth{06}}}{}.
\newblock
{\BBOQ}\APACrefatitle {{{PlaneRCNN}}: {{3D Plane Detection}} and {{Reconstruction From}} a {{Single Image}}} {{{PlaneRCNN}}: {{3D Plane Detection}} and {{Reconstruction From}} a {{Single Image}}}.{\BBCQ}
\newblock
 \APACrefbtitle {2019 {{IEEE}}/{{CVF Conference}} on {{Computer Vision}} and {{Pattern Recognition}} ({{CVPR}})} {2019 {{IEEE}}/{{CVF Conference}} on {{Computer Vision}} and {{Pattern Recognition}} ({{CVPR}})}\ (\BPGS\ 4445--4454).
\newblock
\APACaddressPublisher{{Long Beach, CA, USA}}{{IEEE}}.
\PrintBackRefs{\CurrentBib}

\bibitem [\protect \citeauthoryear {%
Lowe%
}{%
Lowe%
}{%
{\protect \APACyear {1999}}%
}]{%
loweObjectRecognitionLocal1999}
\APACinsertmetastar {%
loweObjectRecognitionLocal1999}%
\begin{APACrefauthors}%
Lowe, D.%
\end{APACrefauthors}%
\unskip\
\newblock
\APACrefYearMonthDay{1999}{{\APACmonth{09}}}{}.
\newblock
{\BBOQ}\APACrefatitle {Object Recognition from Local Scale-Invariant Features} {Object recognition from local scale-invariant features}.{\BBCQ}
\newblock
 \APACrefbtitle {Proceedings of the {{Seventh IEEE International Conference}} on {{Computer Vision}}} {Proceedings of the {{Seventh IEEE International Conference}} on {{Computer Vision}}}\ (\BVOL~2, \BPG~1150-1157 vol.2).
\PrintBackRefs{\CurrentBib}

\bibitem [\protect \citeauthoryear {%
Ma%
, Shu%
, Bai%
, Wang%
\BCBL {}\ \BBA {} Samaras%
}{%
Ma%
\ \protect \BOthers {.}}{%
{\protect \APACyear {2018}}%
}]{%
maDocUNetDocumentImage2018}
\APACinsertmetastar {%
maDocUNetDocumentImage2018}%
\begin{APACrefauthors}%
Ma, K.%
, Shu, Z.%
, Bai, X.%
, Wang, J.%
\BCBL {} Samaras, D.%
\end{APACrefauthors}%
\unskip\
\newblock
\APACrefYearMonthDay{2018}{}{}.
\newblock
{\BBOQ}\APACrefatitle {{{DocUNet}}: {{Document Image Unwarping}} via a {{Stacked U-Net}}} {{{DocUNet}}: {{Document Image Unwarping}} via a {{Stacked U-Net}}}.{\BBCQ}
\newblock
 \APACrefbtitle {Proceedings {{IEEE Conference}} on {{Computer Vision}} and {{Pattern Recognition}} ({{CVPR}})} {Proceedings {{IEEE Conference}} on {{Computer Vision}} and {{Pattern Recognition}} ({{CVPR}})}\ (\BPGS\ 4700--4709).
\PrintBackRefs{\CurrentBib}

\bibitem [\protect \citeauthoryear {%
Malyshev%
, Braginsky%
, Faddeeva%
\BCBL {}\ \BBA {} Gogolin%
}{%
Malyshev%
\ \protect \BOthers {.}}{%
{\protect \APACyear {2021}}%
}]{%
malyshevArtificialNeuralNetwork2021}
\APACinsertmetastar {%
malyshevArtificialNeuralNetwork2021}%
\begin{APACrefauthors}%
Malyshev, M.I.%
, Braginsky, S.A.%
, Faddeeva, E.{\relax Yu}.%
\BCBL {} Gogolin, S.S.%
\end{APACrefauthors}%
\unskip\
\newblock
\APACrefYearMonthDay{2021}{{\APACmonth{11}}}{}.
\newblock
{\BBOQ}\APACrefatitle {Artificial {{Neural Network Detection}} of {{Damaged Goods}} by {{Packaging State}}} {Artificial {{Neural Network Detection}} of {{Damaged Goods}} by {{Packaging State}}}.{\BBCQ}
\newblock
 \APACrefbtitle {2021 {{Intelligent Technologies}} and {{Electronic Devices}} in {{Vehicle}} and {{Road Transport Complex}} ({{TIRVED}})} {2021 {{Intelligent Technologies}} and {{Electronic Devices}} in {{Vehicle}} and {{Road Transport Complex}} ({{TIRVED}})}\ (\BPGS\ 1--7).
\PrintBackRefs{\CurrentBib}

\bibitem [\protect \citeauthoryear {%
Markovitz%
, Lavi%
, Perel%
, Mazor%
\BCBL {}\ \BBA {} Litman%
}{%
Markovitz%
\ \protect \BOthers {.}}{%
{\protect \APACyear {2020}}%
}]{%
markovitzCanYouRead2020}
\APACinsertmetastar {%
markovitzCanYouRead2020}%
\begin{APACrefauthors}%
Markovitz, A.%
, Lavi, I.%
, Perel, O.%
, Mazor, S.%
\BCBL {} Litman, R.%
\end{APACrefauthors}%
\unskip\
\newblock
\APACrefYearMonthDay{2020}{}{}.
\newblock
{\BBOQ}\APACrefatitle {Can {{You Read Me Now}}? {{Content Aware Rectification Using Angle Supervision}}} {Can {{You Read Me Now}}? {{Content Aware Rectification Using Angle Supervision}}}.{\BBCQ}
\newblock
 A.~Vedaldi, H.~Bischof, T.~Brox\BCBL {}\ \BBA {} J\BHBI M.~Frahm\ (\BEDS), \APACrefbtitle {Computer {{Vision}} \textendash{} {{ECCV}} 2020} {Computer {{Vision}} \textendash{} {{ECCV}} 2020}\ (\BPGS\ 208--223).
\newblock
\APACaddressPublisher{{Cham}}{{Springer International Publishing}}.
\PrintBackRefs{\CurrentBib}

\bibitem [\protect \citeauthoryear {%
M{\"a}ttig%
, Lorimer%
, Jost%
\BCBL {}\ \BBA {} Kirks%
}{%
M{\"a}ttig%
\ \protect \BOthers {.}}{%
{\protect \APACyear {2016}}%
}]{%
maettigUntersuchungEinsatzesAugmented2016}
\APACinsertmetastar {%
maettigUntersuchungEinsatzesAugmented2016}%
\begin{APACrefauthors}%
M{\"a}ttig, B.%
, Lorimer, I.%
, Jost, J.%
\BCBL {} Kirks, T.%
\end{APACrefauthors}%
\unskip\
\newblock
\APACrefYearMonthDay{2016}{}{}.
\newblock
{\BBOQ}\APACrefatitle {{Untersuchung des Einsatzes von Augmented Reality im Verpackungsprozess unter Ber\"ucksichtigung spezifischer Anforderungen an die Informationsdarstellung sowie die ergonomische Einbindung des Menschen in den Prozess}} {{Untersuchung des Einsatzes von Augmented Reality im Verpackungsprozess unter Ber\"ucksichtigung spezifischer Anforderungen an die Informationsdarstellung sowie die ergonomische Einbindung des Menschen in den Prozess}}.{\BBCQ}
\newblock
\APACjournalVolNumPages{Logistics Journal : Proceedings}{2016}{10}{,}
\newblock
\begin{APACrefDOI} \doi{10.2195/lj_proc_maettig_de_201610_01} \end{APACrefDOI}
\newblock

\newblock

\PrintBackRefs{\CurrentBib}

\bibitem [\protect \citeauthoryear {%
Mayershofer%
, Ge%
\BCBL {}\ \BBA {} Fottner%
}{%
Mayershofer%
, Ge%
\BCBL {}\ \BBA {} Fottner%
}{%
{\protect \APACyear {2020}}%
}]{%
mayershoferFullySyntheticTrainingIndustrial2020}
\APACinsertmetastar {%
mayershoferFullySyntheticTrainingIndustrial2020}%
\begin{APACrefauthors}%
Mayershofer, C.%
, Ge, T.%
\BCBL {} Fottner, J.%
\end{APACrefauthors}%
\unskip\
\newblock
\APACrefYearMonthDay{2020}{}{}.
\newblock
{\BBOQ}\APACrefatitle {Towards {{Fully-Synthetic Training}} for {{Industrial Applications}}} {Towards {{Fully-Synthetic Training}} for {{Industrial Applications}}}.{\BBCQ}
\newblock
\APACjournalVolNumPages{10th International Conference on Logistics, Informatics and Service Sciences (LISS)}{}{}{8,}
\newblock

\newblock

\PrintBackRefs{\CurrentBib}

\bibitem [\protect \citeauthoryear {%
Mayershofer%
, Holm%
, Molter%
\BCBL {}\ \BBA {} Fottner%
}{%
Mayershofer%
, Holm%
\BCBL {}\ \protect \BOthers {.}}{%
{\protect \APACyear {2020}}%
}]{%
mayershoferLOCOLogisticsObjects2020}
\APACinsertmetastar {%
mayershoferLOCOLogisticsObjects2020}%
\begin{APACrefauthors}%
Mayershofer, C.%
, Holm, D\BHBI M.%
, Molter, B.%
\BCBL {} Fottner, J.%
\end{APACrefauthors}%
\unskip\
\newblock
\APACrefYearMonthDay{2020}{}{}.
\newblock
{\BBOQ}\APACrefatitle {{{LOCO}}: {{Logistics}} Objects in Context} {{{LOCO}}: {{Logistics}} objects in context}.{\BBCQ}
\newblock
 \APACrefbtitle {2020 19th {{IEEE}} International Conference on Machine Learning and Applications ({{ICMLA}})} {2020 19th {{IEEE}} international conference on machine learning and applications ({{ICMLA}})}\ (\BPGS\ 612--617).
\PrintBackRefs{\CurrentBib}

\bibitem [\protect \citeauthoryear {%
Memon%
, Sami%
, Khan%
\BCBL {}\ \BBA {} Uddin%
}{%
Memon%
\ \protect \BOthers {.}}{%
{\protect \APACyear {2020}}%
}]{%
memonHandwrittenOpticalCharacter2020}
\APACinsertmetastar {%
memonHandwrittenOpticalCharacter2020}%
\begin{APACrefauthors}%
Memon, J.%
, Sami, M.%
, Khan, R.A.%
\BCBL {} Uddin, M.%
\end{APACrefauthors}%
\unskip\
\newblock
\APACrefYearMonthDay{2020}{}{}.
\newblock
{\BBOQ}\APACrefatitle {Handwritten {{Optical Character Recognition}} ({{OCR}}): {{A Comprehensive Systematic Literature Review}} ({{SLR}})} {Handwritten {{Optical Character Recognition}} ({{OCR}}): {{A Comprehensive Systematic Literature Review}} ({{SLR}})}.{\BBCQ}
\newblock
\APACjournalVolNumPages{IEEE Access}{8}{}{142642--142668,}
\newblock
\begin{APACrefDOI} \doi{10.1109/ACCESS.2020.3012542} \end{APACrefDOI}
\newblock

\newblock

\PrintBackRefs{\CurrentBib}

\bibitem [\protect \citeauthoryear {%
Meyer%
}{%
Meyer%
}{%
{\protect \APACyear {1992}}%
}]{%
meyerColorImageSegmentation1992}
\APACinsertmetastar {%
meyerColorImageSegmentation1992}%
\begin{APACrefauthors}%
Meyer, F.%
\end{APACrefauthors}%
\unskip\
\newblock
\APACrefYearMonthDay{1992}{{\APACmonth{04}}}{}.
\newblock
{\BBOQ}\APACrefatitle {Color Image Segmentation} {Color image segmentation}.{\BBCQ}
\newblock
 \APACrefbtitle {1992 {{International Conference}} on {{Image Processing}} and Its {{Applications}}} {1992 {{International Conference}} on {{Image Processing}} and its {{Applications}}}\ (\BPGS\ 303--306).
\PrintBackRefs{\CurrentBib}

\bibitem [\protect \citeauthoryear {%
Mieth%
, Humbeck%
\BCBL {}\ \BBA {} Herzwurm%
}{%
Mieth%
\ \protect \BOthers {.}}{%
{\protect \APACyear {2019}}%
}]{%
miethSurveyPotentialsIndoor2019}
\APACinsertmetastar {%
miethSurveyPotentialsIndoor2019}%
\begin{APACrefauthors}%
Mieth, C.%
, Humbeck, P.%
\BCBL {} Herzwurm, G.%
\end{APACrefauthors}%
\unskip\
\newblock
\APACrefYearMonthDay{2019}{}{}.
\newblock
{\BBOQ}\APACrefatitle {A {{Survey}} on the {{Potentials}} of {{Indoor Localization Systems}} in {{Production}}} {A {{Survey}} on the {{Potentials}} of {{Indoor Localization Systems}} in {{Production}}}.{\BBCQ}
\newblock
 U.~Clausen, S.~Langkau\BCBL {}\ \BBA {} F.~Kreuz\ (\BEDS), \APACrefbtitle {Advances in {{Production}}, {{Logistics}} and {{Traffic}}} {Advances in {{Production}}, {{Logistics}} and {{Traffic}}}\ (\BPGS\ 142--154).
\newblock
\APACaddressPublisher{{Cham}}{{Springer International Publishing}}.
\PrintBackRefs{\CurrentBib}

\bibitem [\protect \citeauthoryear {%
Mihalyi%
, Pathak%
, Vaskevicius%
, Fromm%
\BCBL {}\ \BBA {} Birk%
}{%
Mihalyi%
\ \protect \BOthers {.}}{%
{\protect \APACyear {2015}}%
}]{%
mihalyiRobust3DObject2015}
\APACinsertmetastar {%
mihalyiRobust3DObject2015}%
\begin{APACrefauthors}%
Mihalyi, R\BHBI G.%
, Pathak, K.%
, Vaskevicius, N.%
, Fromm, T.%
\BCBL {} Birk, A.%
\end{APACrefauthors}%
\unskip\
\newblock
\APACrefYearMonthDay{2015}{{\APACmonth{04}}}{}.
\newblock
{\BBOQ}\APACrefatitle {Robust {{3D}} Object Modeling with a Low-Cost {{RGBD-sensor}} and {{AR-markers}} for Applications with Untrained End-Users} {Robust {{3D}} object modeling with a low-cost {{RGBD-sensor}} and {{AR-markers}} for applications with untrained end-users}.{\BBCQ}
\newblock
\APACjournalVolNumPages{Robotics and Autonomous Systems}{66}{}{1--17,}
\newblock
\begin{APACrefDOI} \doi{10.1016/j.robot.2015.01.005} \end{APACrefDOI}
\newblock

\newblock

\PrintBackRefs{\CurrentBib}

\bibitem [\protect \citeauthoryear {%
Minaee%
\ \protect \BOthers {.}}{%
Minaee%
\ \protect \BOthers {.}}{%
{\protect \APACyear {2022}}%
}]{%
minaeeImageSegmentationUsing2022}
\APACinsertmetastar {%
minaeeImageSegmentationUsing2022}%
\begin{APACrefauthors}%
Minaee, S.%
, Boykov, Y.%
, Porikli, F.%
, Plaza, A.%
, Kehtarnavaz, N.%
\BCBL {} Terzopoulos, D.%
\end{APACrefauthors}%
\unskip\
\newblock
\APACrefYearMonthDay{2022}{{\APACmonth{07}}}{}.
\newblock
{\BBOQ}\APACrefatitle {Image {{Segmentation Using Deep Learning}}: {{A Survey}}} {Image {{Segmentation Using Deep Learning}}: {{A Survey}}}.{\BBCQ}
\newblock
\APACjournalVolNumPages{IEEE Transactions on Pattern Analysis and Machine Intelligence}{44}{7}{3523--3542,}
\newblock
\begin{APACrefDOI} \doi{10.1109/TPAMI.2021.3059968} \end{APACrefDOI}
\newblock

\newblock

\PrintBackRefs{\CurrentBib}

\bibitem [\protect \citeauthoryear {%
Mishra%
, Kapadia%
, Zaveri%
\BCBL {}\ \BBA {} Pinnamaneni%
}{%
Mishra%
\ \protect \BOthers {.}}{%
{\protect \APACyear {2019}}%
}]{%
mishraDevelopmentLowcostEmbedded2019}
\APACinsertmetastar {%
mishraDevelopmentLowcostEmbedded2019}%
\begin{APACrefauthors}%
Mishra, V.%
, Kapadia, H.K.%
, Zaveri, T.H.%
\BCBL {} Pinnamaneni, B.P.%
\end{APACrefauthors}%
\unskip\
\newblock
\APACrefYearMonthDay{2019}{}{}.
\newblock
{\BBOQ}\APACrefatitle {Development of Low-Cost Embedded Vision System with a Case Study on {{1D}} Barcode Detection} {Development of low-cost embedded vision system with a case study on {{1D}} barcode detection}.{\BBCQ}
\newblock
 \APACrefbtitle {Information and Communication Technology for Intelligent Systems: {{Proceedings}} of {{ICTIS}} 2018} {Information and communication technology for intelligent systems: {{Proceedings}} of {{ICTIS}} 2018}\ (\BVOL~1, \BPGS\ 505--513).
\newblock
\APACaddressPublisher{}{{Springer}}.
\PrintBackRefs{\CurrentBib}

\bibitem [\protect \citeauthoryear {%
Mohamed%
, Capitanelli%
, Mastrogiovanni%
, Rovetta%
\BCBL {}\ \BBA {} Zaccaria%
}{%
Mohamed%
\ \protect \BOthers {.}}{%
{\protect \APACyear {2020}}%
}]{%
mohamedDetectionLocalisationTracking2020}
\APACinsertmetastar {%
mohamedDetectionLocalisationTracking2020}%
\begin{APACrefauthors}%
Mohamed, I.S.%
, Capitanelli, A.%
, Mastrogiovanni, F.%
, Rovetta, S.%
\BCBL {} Zaccaria, R.%
\end{APACrefauthors}%
\unskip\
\newblock
\APACrefYearMonthDay{2020}{{\APACmonth{07}}}{}.
\newblock
{\BBOQ}\APACrefatitle {Detection, Localisation and Tracking of Pallets Using Machine Learning Techniques and {{2D}} Range Data} {Detection, localisation and tracking of pallets using machine learning techniques and {{2D}} range data}.{\BBCQ}
\newblock
\APACjournalVolNumPages{Neural Computing and Applications}{32}{13}{8811--8828,}
\newblock
\begin{APACrefDOI} \doi{10.1007/s00521-019-04352-0} \end{APACrefDOI}
\newblock

\newblock

\PrintBackRefs{\CurrentBib}

\bibitem [\protect \citeauthoryear {%
Molter%
\ \BBA {} Fottner%
}{%
Molter%
\ \BBA {} Fottner%
}{%
{\protect \APACyear {2018}}%
}]{%
molterRealtimePalletLocalization2018}
\APACinsertmetastar {%
molterRealtimePalletLocalization2018}%
\begin{APACrefauthors}%
Molter, B.%
\BCBT {}\ \BBA {} Fottner, J.%
\end{APACrefauthors}%
\unskip\
\newblock
\APACrefYearMonthDay{2018}{{\APACmonth{07}}}{}.
\newblock
{\BBOQ}\APACrefatitle {Real-Time {{Pallet Localization}} with {{3D Camera Technology}} for {{Forklifts}} in {{Logistic Environments}}} {Real-time {{Pallet Localization}} with {{3D Camera Technology}} for {{Forklifts}} in {{Logistic Environments}}}.{\BBCQ}
\newblock
 \APACrefbtitle {2018 {{IEEE International Conference}} on {{Service Operations}} and {{Logistics}}, and {{Informatics}} ({{SOLI}})} {2018 {{IEEE International Conference}} on {{Service Operations}} and {{Logistics}}, and {{Informatics}} ({{SOLI}})}\ (\BPGS\ 297--302).
\PrintBackRefs{\CurrentBib}

\bibitem [\protect \citeauthoryear {%
Molter%
\ \BBA {} Fottner%
}{%
Molter%
\ \BBA {} Fottner%
}{%
{\protect \APACyear {2019}}%
}]{%
molterSemiAutomaticPalletPickup2019}
\APACinsertmetastar {%
molterSemiAutomaticPalletPickup2019}%
\begin{APACrefauthors}%
Molter, B.%
\BCBT {}\ \BBA {} Fottner, J.%
\end{APACrefauthors}%
\unskip\
\newblock
\APACrefYearMonthDay{2019}{{\APACmonth{10}}}{}.
\newblock
{\BBOQ}\APACrefatitle {Semi-{{Automatic Pallet Pick-up}} as an {{Advanced Driver Assistance System}} for {{Forklifts}}} {Semi-{{Automatic Pallet Pick-up}} as an {{Advanced Driver Assistance System}} for {{Forklifts}}}.{\BBCQ}
\newblock
 \APACrefbtitle {2019 {{IEEE Intelligent Transportation Systems Conference}} ({{ITSC}})} {2019 {{IEEE Intelligent Transportation Systems Conference}} ({{ITSC}})}\ (\BPGS\ 4464--4469).
\PrintBackRefs{\CurrentBib}

\bibitem [\protect \citeauthoryear {%
Mor{\'e}%
}{%
Mor{\'e}%
}{%
{\protect \APACyear {1978}}%
}]{%
moreLevenbergMarquardtAlgorithmImplementation1978}
\APACinsertmetastar {%
moreLevenbergMarquardtAlgorithmImplementation1978}%
\begin{APACrefauthors}%
Mor{\'e}, J.J.%
\end{APACrefauthors}%
\unskip\
\newblock
\APACrefYearMonthDay{1978}{}{}.
\newblock
{\BBOQ}\APACrefatitle {The {{Levenberg-Marquardt}} Algorithm: {{Implementation}} and Theory} {The {{Levenberg-Marquardt}} algorithm: {{Implementation}} and theory}.{\BBCQ}
\newblock
 G.A.~Watson\ (\BED), \APACrefbtitle {Numerical {{Analysis}}} {Numerical {{Analysis}}}\ (\BPGS\ 105--116).
\newblock
\APACaddressPublisher{{Berlin, Heidelberg}}{{Springer}}.
\PrintBackRefs{\CurrentBib}

\bibitem [\protect \citeauthoryear {%
M{\"u}ller%
\ \BBA {} Voigtl{\"a}nder%
}{%
M{\"u}ller%
\ \BBA {} Voigtl{\"a}nder%
}{%
{\protect \APACyear {2019}}%
}]{%
mullerAutomatedTrucksRoad2019}
\APACinsertmetastar {%
mullerAutomatedTrucksRoad2019}%
\begin{APACrefauthors}%
M{\"u}ller, S.%
\BCBT {}\ \BBA {} Voigtl{\"a}nder, F.%
\end{APACrefauthors}%
\unskip\
\newblock
\APACrefYearMonthDay{2019}{}{}.
\newblock
{\BBOQ}\APACrefatitle {Automated {{Trucks}} in {{Road Freight Logistics}}: {{The User Perspective}}} {Automated {{Trucks}} in {{Road Freight Logistics}}: {{The User Perspective}}}.{\BBCQ}
\newblock
 U.~Clausen, S.~Langkau\BCBL {}\ \BBA {} F.~Kreuz\ (\BEDS), \APACrefbtitle {Advances in {{Production}}, {{Logistics}} and {{Traffic}}} {Advances in {{Production}}, {{Logistics}} and {{Traffic}}}\ (\BPGS\ 102--115).
\newblock
\APACaddressPublisher{{Cham}}{{Springer International Publishing}}.
\PrintBackRefs{\CurrentBib}

\bibitem [\protect \citeauthoryear {%
Naumann%
, D{\"o}rr%
, Salscheider%
\BCBL {}\ \BBA {} Furmans%
}{%
Naumann%
\ \protect \BOthers {.}}{%
{\protect \APACyear {2020}}%
}]{%
naumannRefinedPlaneSegmentation2020}
\APACinsertmetastar {%
naumannRefinedPlaneSegmentation2020}%
\begin{APACrefauthors}%
Naumann, A.%
, D{\"o}rr, L.%
, Salscheider, N.O.%
\BCBL {} Furmans, K.%
\end{APACrefauthors}%
\unskip\
\newblock
\APACrefYearMonthDay{2020}{{\APACmonth{03}}}{}.
\newblock
{\BBOQ}\APACrefatitle {Refined {{Plane Segmentation}} for {{Cuboid-Shaped Objects}} by {{Leveraging Edge Detection}}} {Refined {{Plane Segmentation}} for {{Cuboid-Shaped Objects}} by {{Leveraging Edge Detection}}}.{\BBCQ}
\newblock
 \APACrefbtitle {International {{Conference On Machine Learning And Applications}}.} {International {{Conference On Machine Learning And Applications}}.}
\PrintBackRefs{\CurrentBib}

\bibitem [\protect \citeauthoryear {%
Naumann%
, Hertlein%
, D{\"o}rr%
\BCBL {}\ \BBA {} Furmans%
}{%
Naumann%
\ \protect \BOthers {.}}{%
{\protect \APACyear {2023}}%
}]{%
naumannParcel3DShapeReconstruction2023}
\APACinsertmetastar {%
naumannParcel3DShapeReconstruction2023}%
\begin{APACrefauthors}%
Naumann, A.%
, Hertlein, F.%
, D{\"o}rr, L.%
\BCBL {} Furmans, K.%
\end{APACrefauthors}%
\unskip\
\newblock
\APACrefYearMonthDay{2023}{{\APACmonth{04}}}{}.
\newblock
\APACrefbtitle {{{Parcel3D}}: {{Shape Reconstruction}} from {{Single RGB Images}} for {{Applications}} in {{Transportation Logistics}}} {{{Parcel3D}}: {{Shape Reconstruction}} from {{Single RGB Images}} for {{Applications}} in {{Transportation Logistics}}}\ (\BNUM\ arXiv:2304.08994).
\newblock
\APACaddressPublisher{}{{arXiv}}.
\PrintBackRefs{\CurrentBib}

\bibitem [\protect \citeauthoryear {%
Naumann%
, Hertlein%
, Zhou%
, Dorr%
\BCBL {}\ \BBA {} Furmans%
}{%
Naumann%
\ \protect \BOthers {.}}{%
{\protect \APACyear {2022}}%
}]{%
naumannScrapeCutPasteLearn2022}
\APACinsertmetastar {%
naumannScrapeCutPasteLearn2022}%
\begin{APACrefauthors}%
Naumann, A.%
, Hertlein, F.%
, Zhou, B.%
, Dorr, L.%
\BCBL {} Furmans, K.%
\end{APACrefauthors}%
\unskip\
\newblock
\APACrefYearMonthDay{2022}{{\APACmonth{12}}}{}.
\newblock
{\BBOQ}\APACrefatitle {Scrape, {{Cut}}, {{Paste}} and {{Learn}}: {{Automated Dataset Generation Applied}} to {{Parcel Logistics}}} {Scrape, {{Cut}}, {{Paste}} and {{Learn}}: {{Automated Dataset Generation Applied}} to {{Parcel Logistics}}}.{\BBCQ}
\newblock
 \APACrefbtitle {2022 21st {{IEEE International Conference}} on {{Machine Learning}} and {{Applications}} ({{ICMLA}})} {2022 21st {{IEEE International Conference}} on {{Machine Learning}} and {{Applications}} ({{ICMLA}})}\ (\BPGS\ 1026--1031).
\PrintBackRefs{\CurrentBib}

\bibitem [\protect \citeauthoryear {%
Noceti%
, Zini%
\BCBL {}\ \BBA {} Odone%
}{%
Noceti%
\ \protect \BOthers {.}}{%
{\protect \APACyear {2018}}%
}]{%
nocetiMulticameraSystemDamage2018}
\APACinsertmetastar {%
nocetiMulticameraSystemDamage2018}%
\begin{APACrefauthors}%
Noceti, N.%
, Zini, L.%
\BCBL {} Odone, F.%
\end{APACrefauthors}%
\unskip\
\newblock
\APACrefYearMonthDay{2018}{{\APACmonth{02}}}{}.
\newblock
{\BBOQ}\APACrefatitle {A Multi-Camera System for Damage and Tampering Detection in a Postal Security Framework} {A multi-camera system for damage and tampering detection in a postal security framework}.{\BBCQ}
\newblock
\APACjournalVolNumPages{EURASIP Journal on Image and Video Processing}{2018}{1}{11,}
\newblock
\begin{APACrefDOI} \doi{10.1186/s13640-017-0242-x} \end{APACrefDOI}
\newblock

\newblock

\PrintBackRefs{\CurrentBib}

\bibitem [\protect \citeauthoryear {%
{\"O}zg{\"u}r%
, Alias%
\BCBL {}\ \BBA {} Noche%
}{%
{\"O}zg{\"u}r%
\ \protect \BOthers {.}}{%
{\protect \APACyear {2016}}%
}]{%
ozgurComparingSensorbasedCamerabased2016}
\APACinsertmetastar {%
ozgurComparingSensorbasedCamerabased2016}%
\begin{APACrefauthors}%
{\"O}zg{\"u}r, {\c C}.%
, Alias, C.%
\BCBL {} Noche, B.%
\end{APACrefauthors}%
\unskip\
\newblock
\APACrefYearMonthDay{2016}{}{}.
\newblock
{\BBOQ}\APACrefatitle {Comparing Sensor-Based and Camera-Based Approaches to Recognizing the Occupancy Status of the Load Handling Device of Forklift Trucks} {Comparing sensor-based and camera-based approaches to recognizing the occupancy status of the load handling device of forklift trucks}.{\BBCQ}
\newblock
\APACjournalVolNumPages{Logistics Journal : Proceedings}{2016}{05}{,}
\newblock
\begin{APACrefDOI} \doi{10.2195/lj_proc_oezguer_en_201605_01} \end{APACrefDOI}
\newblock

\newblock

\PrintBackRefs{\CurrentBib}

\bibitem [\protect \citeauthoryear {%
Pavlichenko%
, Garc{\'i}a%
, Koo%
\BCBL {}\ \BBA {} Behnke%
}{%
Pavlichenko%
\ \protect \BOthers {.}}{%
{\protect \APACyear {2019}}%
}]{%
pavlichenkoKittingBotMobileManipulation2019}
\APACinsertmetastar {%
pavlichenkoKittingBotMobileManipulation2019}%
\begin{APACrefauthors}%
Pavlichenko, D.%
, Garc{\'i}a, G.M.%
, Koo, S.%
\BCBL {} Behnke, S.%
\end{APACrefauthors}%
\unskip\
\newblock
\APACrefYearMonthDay{2019}{}{}.
\newblock
{\BBOQ}\APACrefatitle {{{KittingBot}}: {{A Mobile Manipulation Robot}} for {{Collaborative Kitting}} in {{Automotive Logistics}}} {{{KittingBot}}: {{A Mobile Manipulation Robot}} for {{Collaborative Kitting}} in {{Automotive Logistics}}}.{\BBCQ}
\newblock
 M.~Strand, R.~Dillmann, E.~Menegatti\BCBL {}\ \BBA {} S.~Ghidoni\ (\BEDS), \APACrefbtitle {Intelligent {{Autonomous Systems}} 15} {Intelligent {{Autonomous Systems}} 15}\ (\BVOL~867, \BPGS\ 849--864).
\newblock
\APACaddressPublisher{{Cham}}{{Springer International Publishing}}.
\PrintBackRefs{\CurrentBib}

\bibitem [\protect \citeauthoryear {%
Pinto%
, McCallum%
, Wei%
\BCBL {}\ \BBA {} Croft%
}{%
Pinto%
\ \protect \BOthers {.}}{%
{\protect \APACyear {2003}}%
}]{%
pintoTableExtractionUsing2003}
\APACinsertmetastar {%
pintoTableExtractionUsing2003}%
\begin{APACrefauthors}%
Pinto, D.%
, McCallum, A.%
, Wei, X.%
\BCBL {} Croft, W.B.%
\end{APACrefauthors}%
\unskip\
\newblock
\APACrefYearMonthDay{2003}{{\APACmonth{07}}}{}.
\newblock
{\BBOQ}\APACrefatitle {Table Extraction Using Conditional Random Fields} {Table extraction using conditional random fields}.{\BBCQ}
\newblock
 \APACrefbtitle {Proceedings of the 26th Annual International {{ACM SIGIR}} Conference on {{Research}} and Development in Informaion Retrieval} {Proceedings of the 26th annual international {{ACM SIGIR}} conference on {{Research}} and development in informaion retrieval}\ (\BPGS\ 235--242).
\newblock
\APACaddressPublisher{{New York, NY, USA}}{{Association for Computing Machinery}}.
\PrintBackRefs{\CurrentBib}

\bibitem [\protect \citeauthoryear {%
Prasse%
\ \protect \BOthers {.}}{%
Prasse%
\ \protect \BOthers {.}}{%
{\protect \APACyear {2015}}%
}]{%
prasseNewApproachesSingularization2015}
\APACinsertmetastar {%
prasseNewApproachesSingularization2015}%
\begin{APACrefauthors}%
Prasse, C.%
, Stenzel, J.%
, B{\"o}ckenkamp, A.%
, Rudak, B.%
, Lorenz, K.%
, Weichert, F.%
\BDBL {}{ten Hompel}, M.%
\end{APACrefauthors}%
\unskip\
\newblock
\APACrefYearMonthDay{2015}{}{}.
\newblock
{\BBOQ}\APACrefatitle {New {{Approaches}} for {{Singularization}} in {{Logistic Applications Using Low Cost 3D Sensors}}} {New {{Approaches}} for {{Singularization}} in {{Logistic Applications Using Low Cost 3D Sensors}}}.{\BBCQ}
\newblock
 A.~Mason, S.C.~Mukhopadhyay\BCBL {}\ \BBA {} K.P.~Jayasundera\ (\BEDS), \APACrefbtitle {Sensing {{Technology}}: {{Current Status}} and {{Future Trends IV}}} {Sensing {{Technology}}: {{Current Status}} and {{Future Trends IV}}}\ (\BPGS\ 191--215).
\newblock
\APACaddressPublisher{{Cham}}{{Springer International Publishing}}.
\PrintBackRefs{\CurrentBib}

\bibitem [\protect \citeauthoryear {%
Redmon%
\ \BBA {} Farhadi%
}{%
Redmon%
\ \BBA {} Farhadi%
}{%
{\protect \APACyear {2018}}%
}]{%
redmonYOLOv3IncrementalImprovement2018}
\APACinsertmetastar {%
redmonYOLOv3IncrementalImprovement2018}%
\begin{APACrefauthors}%
Redmon, J.%
\BCBT {}\ \BBA {} Farhadi, A.%
\end{APACrefauthors}%
\unskip\
\newblock
\APACrefYearMonthDay{2018}{{\APACmonth{04}}}{}.
\newblock
\APACrefbtitle {{{YOLOv3}}: {{An Incremental Improvement}}} {{{YOLOv3}}: {{An Incremental Improvement}}}\ (\BNUM\ arXiv:1804.02767).
\newblock
\APACaddressPublisher{}{{arXiv}}.
\PrintBackRefs{\CurrentBib}

\bibitem [\protect \citeauthoryear {%
Reif%
}{%
Reif%
}{%
{\protect \APACyear {2009}}%
}]{%
reifEntwicklungUndEvaluierung2009}
\APACinsertmetastar {%
reifEntwicklungUndEvaluierung2009}%
\begin{APACrefauthors}%
Reif, R.%
\end{APACrefauthors}%
\unskip\
\newblock
\APACrefYear{2009}.
\unskip\
\newblock
\APACrefbtitle {{Entwicklung und Evaluierung eines Augmented Reality unterst\"utzten Kommissioniersystems}} {{Entwicklung und Evaluierung eines Augmented Reality unterst\"utzten Kommissioniersystems}}\ \APACtypeAddressSchool {\BUPhD}{}{}.
\unskip\
\newblock
\APACaddressSchool {{Garching b. M\"unchen}}{Lehrstuhl f\"ur F\"ordertechnik Materialflu\ss{} Logistik (fml), Techn. Univ. M\"unchen}.
\PrintBackRefs{\CurrentBib}

\bibitem [\protect \citeauthoryear {%
Ren%
, He%
, Girshick%
\BCBL {}\ \BBA {} Sun%
}{%
Ren%
\ \protect \BOthers {.}}{%
{\protect \APACyear {2017}}%
}]{%
renFasterRCNNRealTime2017}
\APACinsertmetastar {%
renFasterRCNNRealTime2017}%
\begin{APACrefauthors}%
Ren, S.%
, He, K.%
, Girshick, R.%
\BCBL {} Sun, J.%
\end{APACrefauthors}%
\unskip\
\newblock
\APACrefYearMonthDay{2017}{{\APACmonth{06}}}{}.
\newblock
{\BBOQ}\APACrefatitle {Faster {{R-CNN}}: {{Towards Real-Time Object Detection}} with {{Region Proposal Networks}}} {Faster {{R-CNN}}: {{Towards Real-Time Object Detection}} with {{Region Proposal Networks}}}.{\BBCQ}
\newblock
\APACjournalVolNumPages{IEEE Transactions on Pattern Analysis and Machine Intelligence}{39}{6}{1137--1149,}
\newblock
\begin{APACrefDOI} \doi{10.1109/TPAMI.2016.2577031} \end{APACrefDOI}
\newblock

\newblock

\PrintBackRefs{\CurrentBib}

\bibitem [\protect \citeauthoryear {%
Rennie%
, Shome%
, Bekris%
\BCBL {}\ \BBA {} De~Souza%
}{%
Rennie%
\ \protect \BOthers {.}}{%
{\protect \APACyear {2016}}%
}]{%
rennieDatasetImprovedRGBDBased2016}
\APACinsertmetastar {%
rennieDatasetImprovedRGBDBased2016}%
\begin{APACrefauthors}%
Rennie, C.%
, Shome, R.%
, Bekris, K.E.%
\BCBL {} De~Souza, A.F.%
\end{APACrefauthors}%
\unskip\
\newblock
\APACrefYearMonthDay{2016}{{\APACmonth{07}}}{}.
\newblock
{\BBOQ}\APACrefatitle {A {{Dataset}} for {{Improved RGBD-Based Object Detection}} and {{Pose Estimation}} for {{Warehouse Pick-and-Place}}} {A {{Dataset}} for {{Improved RGBD-Based Object Detection}} and {{Pose Estimation}} for {{Warehouse Pick-and-Place}}}.{\BBCQ}
\newblock
\APACjournalVolNumPages{IEEE Robotics and Automation Letters}{1}{2}{1179--1185,}
\newblock
\begin{APACrefDOI} \doi{10.1109/LRA.2016.2532924} \end{APACrefDOI}
\newblock

\newblock

\PrintBackRefs{\CurrentBib}

\bibitem [\protect \citeauthoryear {%
Riad%
, Sporer%
, Bukhari%
\BCBL {}\ \BBA {} Dengel%
}{%
Riad%
\ \protect \BOthers {.}}{%
{\protect \APACyear {2017}}%
}]{%
riadClassificationInformationExtraction2017}
\APACinsertmetastar {%
riadClassificationInformationExtraction2017}%
\begin{APACrefauthors}%
Riad, A.%
, Sporer, C.%
, Bukhari, S.S.%
\BCBL {} Dengel, A.%
\end{APACrefauthors}%
\unskip\
\newblock
\APACrefYearMonthDay{2017}{{\APACmonth{11}}}{}.
\newblock
{\BBOQ}\APACrefatitle {Classification and {{Information Extraction}} for {{Complex}} and {{Nested Tabular Structures}} in {{Images}}} {Classification and {{Information Extraction}} for {{Complex}} and {{Nested Tabular Structures}} in {{Images}}}.{\BBCQ}
\newblock
 \APACrefbtitle {2017 14th {{IAPR International Conference}} on {{Document Analysis}} and {{Recognition}} ({{ICDAR}})} {2017 14th {{IAPR International Conference}} on {{Document Analysis}} and {{Recognition}} ({{ICDAR}})}\ (\BVOL~01, \BPGS\ 1156--1161).
\PrintBackRefs{\CurrentBib}

\bibitem [\protect \citeauthoryear {%
Riestock%
, Fessel%
, Depner%
\BCBL {}\ \BBA {} Borstell%
}{%
Riestock%
\ \protect \BOthers {.}}{%
{\protect \APACyear {2019}}%
}]{%
riestockSurveyDepthCameras2019}
\APACinsertmetastar {%
riestockSurveyDepthCameras2019}%
\begin{APACrefauthors}%
Riestock, M.%
, Fessel, K.%
, Depner, T.%
\BCBL {} Borstell, H.%
\end{APACrefauthors}%
\unskip\
\newblock
\APACrefYearMonthDay{2019}{{\APACmonth{06}}}{}.
\newblock
{\BBOQ}\APACrefatitle {Survey of {{Depth Cameras}} for {{Process-integrated State Detection}} in {{Logistics}}} {Survey of {{Depth Cameras}} for {{Process-integrated State Detection}} in {{Logistics}}}.{\BBCQ}
\newblock
 \APACrefbtitle {Smart {{SysTech}} 2019; {{European Conference}} on {{Smart Objects}}, {{Systems}} and {{Technologies}}} {Smart {{SysTech}} 2019; {{European Conference}} on {{Smart Objects}}, {{Systems}} and {{Technologies}}}\ (\BPGS\ 1--6).
\PrintBackRefs{\CurrentBib}

\bibitem [\protect \citeauthoryear {%
Rombach%
, Blattmann%
, Lorenz%
, Esser%
\BCBL {}\ \BBA {} Ommer%
}{%
Rombach%
\ \protect \BOthers {.}}{%
{\protect \APACyear {2022}}%
}]{%
rombachHighResolutionImageSynthesis2022}
\APACinsertmetastar {%
rombachHighResolutionImageSynthesis2022}%
\begin{APACrefauthors}%
Rombach, R.%
, Blattmann, A.%
, Lorenz, D.%
, Esser, P.%
\BCBL {} Ommer, B.%
\end{APACrefauthors}%
\unskip\
\newblock
\APACrefYearMonthDay{2022}{}{}.
\newblock
{\BBOQ}\APACrefatitle {High-{{Resolution Image Synthesis With Latent Diffusion Models}}} {High-{{Resolution Image Synthesis With Latent Diffusion Models}}}.{\BBCQ}
\newblock
 \APACrefbtitle {Proceedings of the {{IEEE}}/{{CVF Conference}} on {{Computer Vision}} and {{Pattern Recognition}}} {Proceedings of the {{IEEE}}/{{CVF Conference}} on {{Computer Vision}} and {{Pattern Recognition}}}\ (\BPGS\ 10684--10695).
\PrintBackRefs{\CurrentBib}

\bibitem [\protect \citeauthoryear {%
Rui%
, Zongyuan%
, Simon%
, Sridha%
\BCBL {}\ \BBA {} Clinton%
}{%
Rui%
\ \protect \BOthers {.}}{%
{\protect \APACyear {2020}}%
}]{%
ruiGeometryConstrainedCarRecognition2020}
\APACinsertmetastar {%
ruiGeometryConstrainedCarRecognition2020}%
\begin{APACrefauthors}%
Rui, Z.%
, Zongyuan, G.%
, Simon, D.%
, Sridha, S.%
\BCBL {} Clinton, F.%
\end{APACrefauthors}%
\unskip\
\newblock
\APACrefYearMonthDay{2020}{{\APACmonth{04}}}{}.
\newblock
{\BBOQ}\APACrefatitle {Geometry-{{Constrained Car Recognition Using}} a {{3D Perspective Network}}} {Geometry-{{Constrained Car Recognition Using}} a {{3D Perspective Network}}}.{\BBCQ}
\newblock
\APACjournalVolNumPages{Proceedings of the AAAI Conference on Artificial Intelligence}{34}{01}{1161--1168,}
\newblock
\begin{APACrefDOI} \doi{10.1609/aaai.v34i01.5468} \end{APACrefDOI}
\newblock

\newblock

\PrintBackRefs{\CurrentBib}

\bibitem [\protect \citeauthoryear {%
Rusu%
, Bradski%
, Thibaux%
\BCBL {}\ \BBA {} Hsu%
}{%
Rusu%
\ \protect \BOthers {.}}{%
{\protect \APACyear {2010}}%
}]{%
rusuFast3DRecognition2010}
\APACinsertmetastar {%
rusuFast3DRecognition2010}%
\begin{APACrefauthors}%
Rusu, R.B.%
, Bradski, G.%
, Thibaux, R.%
\BCBL {} Hsu, J.%
\end{APACrefauthors}%
\unskip\
\newblock
\APACrefYearMonthDay{2010}{{\APACmonth{10}}}{}.
\newblock
{\BBOQ}\APACrefatitle {Fast {{3D}} Recognition and Pose Using the {{Viewpoint Feature Histogram}}} {Fast {{3D}} recognition and pose using the {{Viewpoint Feature Histogram}}}.{\BBCQ}
\newblock
 \APACrefbtitle {2010 {{IEEE}}/{{RSJ International Conference}} on {{Intelligent Robots}} and {{Systems}}} {2010 {{IEEE}}/{{RSJ International Conference}} on {{Intelligent Robots}} and {{Systems}}}\ (\BPGS\ 2155--2162).
\PrintBackRefs{\CurrentBib}

\bibitem [\protect \citeauthoryear {%
Rutinowski%
, Pionzewski%
, Chilla%
, Reining%
\BCBL {}\ \BBA {} Hompel%
}{%
Rutinowski%
, Pionzewski%
\BCBL {}\ \protect \BOthers {.}}{%
{\protect \APACyear {2022}}%
}]{%
rutinowskiDeepLearningBased2022}
\APACinsertmetastar {%
rutinowskiDeepLearningBased2022}%
\begin{APACrefauthors}%
Rutinowski, J.%
, Pionzewski, C.%
, Chilla, T.%
, Reining, C.%
\BCBL {} Hompel, M.T.%
\end{APACrefauthors}%
\unskip\
\newblock
\APACrefYearMonthDay{2022}{{\APACmonth{12}}}{}.
\newblock
{\BBOQ}\APACrefatitle {Deep {{Learning Based Re-Identification}} of {{Wooden Euro-pallets}}} {Deep {{Learning Based Re-Identification}} of {{Wooden Euro-pallets}}}.{\BBCQ}
\newblock
 \APACrefbtitle {2022 21st {{IEEE International Conference}} on {{Machine Learning}} and {{Applications}} ({{ICMLA}})} {2022 21st {{IEEE International Conference}} on {{Machine Learning}} and {{Applications}} ({{ICMLA}})}\ (\BPGS\ 113--117).
\PrintBackRefs{\CurrentBib}

\bibitem [\protect \citeauthoryear {%
Rutinowski%
, Pionzewski%
, Chilla%
, Reining%
\BCBL {}\ \BBA {} ten Hompel%
}{%
Rutinowski%
\ \protect \BOthers {.}}{%
{\protect \APACyear {2021}}%
}]{%
rutinowskiReIdentificationWarehousingEntities2021}
\APACinsertmetastar {%
rutinowskiReIdentificationWarehousingEntities2021}%
\begin{APACrefauthors}%
Rutinowski, J.%
, Pionzewski, C.%
, Chilla, T.%
, Reining, C.%
\BCBL {} ten Hompel, M.%
\end{APACrefauthors}%
\unskip\
\newblock
\APACrefYearMonthDay{2021}{{\APACmonth{09}}}{}.
\newblock
{\BBOQ}\APACrefatitle {Towards {{Re-Identification}} for {{Warehousing Entities}} - {{A Work-in-Progress Study}}} {Towards {{Re-Identification}} for {{Warehousing Entities}} - {{A Work-in-Progress Study}}}.{\BBCQ}
\newblock
 \APACrefbtitle {2021 26th {{IEEE International Conference}} on {{Emerging Technologies}} and {{Factory Automation}} ({{ETFA}} )} {2021 26th {{IEEE International Conference}} on {{Emerging Technologies}} and {{Factory Automation}} ({{ETFA}} )}\ (\BPGS\ 1--4).
\PrintBackRefs{\CurrentBib}

\bibitem [\protect \citeauthoryear {%
Rutinowski%
, Youssef%
, Gouda%
, Reining%
\BCBL {}\ \BBA {} Roidl%
}{%
Rutinowski%
, Youssef%
\BCBL {}\ \protect \BOthers {.}}{%
{\protect \APACyear {2022}}%
}]{%
rutinowskijeromePotentialDeepLearning2022}
\APACinsertmetastar {%
rutinowskijeromePotentialDeepLearning2022}%
\begin{APACrefauthors}%
Rutinowski, J.%
, Youssef, H.%
, Gouda, A.%
, Reining, C.%
\BCBL {} Roidl, M.%
\end{APACrefauthors}%
\unskip\
\newblock
\APACrefYearMonthDay{2022}{}{}.
\newblock
{\BBOQ}\APACrefatitle {The {{Potential}} of {{Deep Learning}} Based {{Computer Vision}} in {{Warehousing Logistics}}} {The {{Potential}} of {{Deep Learning}} based {{Computer Vision}} in {{Warehousing Logistics}}}.{\BBCQ}
\newblock
\APACjournalVolNumPages{Volume 2022}{}{}{Issue 18,}
\newblock
\begin{APACrefDOI} \doi{10.2195/LJ_PROC_RUTINOWSKI_EN_202211_01} \end{APACrefDOI}
\newblock

\newblock

\PrintBackRefs{\CurrentBib}

\bibitem [\protect \citeauthoryear {%
Sabattini%
\ \protect \BOthers {.}}{%
Sabattini%
\ \protect \BOthers {.}}{%
{\protect \APACyear {2018}}%
}]{%
sabattiniPANRobotsProjectAdvanced2018}
\APACinsertmetastar {%
sabattiniPANRobotsProjectAdvanced2018}%
\begin{APACrefauthors}%
Sabattini, L.%
, Aikio, M.%
, Beinschob, P.%
, Boehning, M.%
, Cardarelli, E.%
, Digani, V.%
\BDBL {}Fuerstenberg, K.%
\end{APACrefauthors}%
\unskip\
\newblock
\APACrefYearMonthDay{2018}{{\APACmonth{03}}}{}.
\newblock
{\BBOQ}\APACrefatitle {The {{PAN-Robots Project}}: {{Advanced Automated Guided Vehicle Systems}} for {{Industrial Logistics}}} {The {{PAN-Robots Project}}: {{Advanced Automated Guided Vehicle Systems}} for {{Industrial Logistics}}}.{\BBCQ}
\newblock
\APACjournalVolNumPages{IEEE Robotics Automation Magazine}{25}{1}{55--64,}
\newblock
\begin{APACrefDOI} \doi{10.1109/MRA.2017.2700325} \end{APACrefDOI}
\newblock

\newblock

\PrintBackRefs{\CurrentBib}

\bibitem [\protect \citeauthoryear {%
Sarlin%
, DeTone%
, Malisiewicz%
\BCBL {}\ \BBA {} Rabinovich%
}{%
Sarlin%
\ \protect \BOthers {.}}{%
{\protect \APACyear {2020}}%
}]{%
sarlinSuperGlueLearningFeature2020}
\APACinsertmetastar {%
sarlinSuperGlueLearningFeature2020}%
\begin{APACrefauthors}%
Sarlin, P\BHBI E.%
, DeTone, D.%
, Malisiewicz, T.%
\BCBL {} Rabinovich, A.%
\end{APACrefauthors}%
\unskip\
\newblock
\APACrefYearMonthDay{2020}{{\APACmonth{06}}}{}.
\newblock
{\BBOQ}\APACrefatitle {{{SuperGlue}}: {{Learning Feature Matching With Graph Neural Networks}}} {{{SuperGlue}}: {{Learning Feature Matching With Graph Neural Networks}}}.{\BBCQ}
\newblock
 \APACrefbtitle {2020 {{IEEE}}/{{CVF Conference}} on {{Computer Vision}} and {{Pattern Recognition}} ({{CVPR}})} {2020 {{IEEE}}/{{CVF Conference}} on {{Computer Vision}} and {{Pattern Recognition}} ({{CVPR}})}\ (\BPGS\ 4937--4946).
\newblock
\APACaddressPublisher{{Seattle, WA, USA}}{{IEEE}}.
\PrintBackRefs{\CurrentBib}

\bibitem [\protect \citeauthoryear {%
Schnabel%
, Wahl%
, Wessel%
\BCBL {}\ \BBA {} Klein%
}{%
Schnabel%
\ \protect \BOthers {.}}{%
{\protect \APACyear {2007}}%
}]{%
schnabelShapeRecognition3D2007}
\APACinsertmetastar {%
schnabelShapeRecognition3D2007}%
\begin{APACrefauthors}%
Schnabel, R.%
, Wahl, R.%
, Wessel, R.%
\BCBL {} Klein, R.%
\end{APACrefauthors}%
\unskip\
\newblock
\APACrefYearMonthDay{2007}{}{}.
\newblock
\APACrefbtitle {Shape {{Recognition}} in {{3D Point Clouds}}} {Shape {{Recognition}} in {{3D Point Clouds}}}\ \APACbVolEdTR {}{Technical {{Report}}\ \BNUM\ CG-2007/1}.
\newblock
\APACaddressInstitution{{Bonn, Germany}}{}.
\PrintBackRefs{\CurrentBib}

\bibitem [\protect \citeauthoryear {%
{Scholz-Reiter}%
, Thamer%
\BCBL {}\ \BBA {} Uriarte%
}{%
{Scholz-Reiter}%
\ \protect \BOthers {.}}{%
{\protect \APACyear {2011}}%
}]{%
scholz-reiterApproach3DObject2011}
\APACinsertmetastar {%
scholz-reiterApproach3DObject2011}%
\begin{APACrefauthors}%
{Scholz-Reiter}, B.%
, Thamer, H.%
\BCBL {} Uriarte, C.%
\end{APACrefauthors}%
\unskip\
\newblock
\APACrefYearMonthDay{2011}{}{}.
\newblock
{\BBOQ}\APACrefatitle {An {{Approach}} for {{3D Object Recognition}} of {{Universal Goods}}} {An {{Approach}} for {{3D Object Recognition}} of {{Universal Goods}}}.{\BBCQ}
\newblock
\APACjournalVolNumPages{International Journal of Computers}{5}{2}{8,}
\newblock

\newblock

\PrintBackRefs{\CurrentBib}

\bibitem [\protect \citeauthoryear {%
Schwarz%
\ \protect \BOthers {.}}{%
Schwarz%
\ \protect \BOthers {.}}{%
{\protect \APACyear {2017}}%
}]{%
schwarzNimbRoPickingVersatile2017}
\APACinsertmetastar {%
schwarzNimbRoPickingVersatile2017}%
\begin{APACrefauthors}%
Schwarz, M.%
, Milan, A.%
, Lenz, C.%
, Munoz, A.%
, Periyasamy, A.S.%
, Schreiber, M.%
\BDBL {}Behnke, S.%
\end{APACrefauthors}%
\unskip\
\newblock
\APACrefYearMonthDay{2017}{{\APACmonth{05}}}{}.
\newblock
{\BBOQ}\APACrefatitle {{{NimbRo}} Picking: {{Versatile}} Part Handling for Warehouse Automation} {{{NimbRo}} picking: {{Versatile}} part handling for warehouse automation}.{\BBCQ}
\newblock
 \APACrefbtitle {2017 {{IEEE International Conference}} on {{Robotics}} and {{Automation}} ({{ICRA}})} {2017 {{IEEE International Conference}} on {{Robotics}} and {{Automation}} ({{ICRA}})}\ (\BPGS\ 3032--3039).
\newblock
\APACaddressPublisher{{Singapore, Singapore}}{{IEEE}}.
\PrintBackRefs{\CurrentBib}

\bibitem [\protect \citeauthoryear {%
Shetty%
, C{\'a}ceres%
, Pastrana%
\BCBL {}\ \BBA {} Rabelo%
}{%
Shetty%
\ \protect \BOthers {.}}{%
{\protect \APACyear {2012}}%
}]{%
shettyOpticalContainerCode2012}
\APACinsertmetastar {%
shettyOpticalContainerCode2012}%
\begin{APACrefauthors}%
Shetty, R.%
, C{\'a}ceres, R.%
, Pastrana, J.%
\BCBL {} Rabelo, L.%
\end{APACrefauthors}%
\unskip\
\newblock
\APACrefYearMonthDay{2012}{}{}.
\newblock
{\BBOQ}\APACrefatitle {Optical {{Container Code Recognition}} and Its {{Impact}} on the {{Maritime Supply Chain}}} {Optical {{Container Code Recognition}} and its {{Impact}} on the {{Maritime Supply Chain}}}.{\BBCQ}
\newblock
 \APACrefbtitle {2012 {{Industrial}} and {{Systems Engineering Research Conference}}} {2012 {{Industrial}} and {{Systems Engineering Research Conference}}}\ (\BPG~10).
\PrintBackRefs{\CurrentBib}

\bibitem [\protect \citeauthoryear {%
Shvarts%
\ \BBA {} Tamre%
}{%
Shvarts%
\ \BBA {} Tamre%
}{%
{\protect \APACyear {2014}}%
}]{%
shvartsBulkMaterialVolume2014}
\APACinsertmetastar {%
shvartsBulkMaterialVolume2014}%
\begin{APACrefauthors}%
Shvarts, D.%
\BCBT {}\ \BBA {} Tamre, M.%
\end{APACrefauthors}%
\unskip\
\newblock
\APACrefYearMonthDay{2014}{{\APACmonth{04}}}{}.
\newblock
{\BBOQ}\APACrefatitle {Bulk {{Material Volume Estimation Method}} and {{System}} for {{Logistic Applications}}} {Bulk {{Material Volume Estimation Method}} and {{System}} for {{Logistic Applications}}}.{\BBCQ}
\newblock
\APACjournalVolNumPages{9th International DAAAM Baltic Conference ``INDUSTRIAL ENGINEERING}{}{}{7,}
\newblock

\newblock

\PrintBackRefs{\CurrentBib}

\bibitem [\protect \citeauthoryear {%
Son%
, Anh%
, Ban%
\BCBL {}\ \BBA {} Duong%
}{%
Son%
\ \protect \BOthers {.}}{%
{\protect \APACyear {2017}}%
}]{%
sonMethodConstructAutomatic2017}
\APACinsertmetastar {%
sonMethodConstructAutomatic2017}%
\begin{APACrefauthors}%
Son, N.T.%
, Anh, B.N.%
, Ban, T.Q.%
\BCBL {} Duong, T.B.%
\end{APACrefauthors}%
\unskip\
\newblock
\APACrefYearMonthDay{2017}{{\APACmonth{07}}}{}.
\newblock
{\BBOQ}\APACrefatitle {A {{Method}} to {{Construct Automatic Object Bounding-Box Estimation System}} Using {{3D Cameras}}} {A {{Method}} to {{Construct Automatic Object Bounding-Box Estimation System}} using {{3D Cameras}}}.{\BBCQ}
\newblock
\APACjournalVolNumPages{International Journal of Science and Research (IJSR)}{6}{7}{961--965,}
\newblock
\begin{APACrefDOI} \doi{10.21275/ART20175316} \end{APACrefDOI}
\newblock

\newblock

\PrintBackRefs{\CurrentBib}

\bibitem [\protect \citeauthoryear {%
Soria%
, Riba%
\BCBL {}\ \BBA {} Sappa%
}{%
Soria%
\ \protect \BOthers {.}}{%
{\protect \APACyear {2020}}%
}]{%
soriaDenseExtremeInception2020}
\APACinsertmetastar {%
soriaDenseExtremeInception2020}%
\begin{APACrefauthors}%
Soria, X.%
, Riba, E.%
\BCBL {} Sappa, A.%
\end{APACrefauthors}%
\unskip\
\newblock
\APACrefYearMonthDay{2020}{{\APACmonth{03}}}{}.
\newblock
{\BBOQ}\APACrefatitle {Dense {{Extreme Inception Network}}: {{Towards}} a {{Robust CNN Model}} for {{Edge Detection}}} {Dense {{Extreme Inception Network}}: {{Towards}} a {{Robust CNN Model}} for {{Edge Detection}}}.{\BBCQ}
\newblock
 \APACrefbtitle {2020 {{IEEE Winter Conference}} on {{Applications}} of {{Computer Vision}} ({{WACV}})} {2020 {{IEEE Winter Conference}} on {{Applications}} of {{Computer Vision}} ({{WACV}})}\ (\BPGS\ 1912--1921).
\newblock
\APACaddressPublisher{{Snowmass Village, CO, USA}}{{IEEE}}.
\PrintBackRefs{\CurrentBib}

\bibitem [\protect \citeauthoryear {%
Staglian{\`o}%
, Noceti%
, Verri%
\BCBL {}\ \BBA {} Odone%
}{%
Staglian{\`o}%
\ \protect \BOthers {.}}{%
{\protect \APACyear {2015}}%
}]{%
staglianoOnlineSpaceVariantBackground2015}
\APACinsertmetastar {%
staglianoOnlineSpaceVariantBackground2015}%
\begin{APACrefauthors}%
Staglian{\`o}, A.%
, Noceti, N.%
, Verri, A.%
\BCBL {} Odone, F.%
\end{APACrefauthors}%
\unskip\
\newblock
\APACrefYearMonthDay{2015}{{\APACmonth{08}}}{}.
\newblock
{\BBOQ}\APACrefatitle {Online {{Space-Variant Background Modeling With Sparse Coding}}} {Online {{Space-Variant Background Modeling With Sparse Coding}}}.{\BBCQ}
\newblock
\APACjournalVolNumPages{IEEE Transactions on Image Processing}{24}{8}{2415--2428,}
\newblock
\begin{APACrefDOI} \doi{10/ggdmng} \end{APACrefDOI}
\newblock

\newblock

\PrintBackRefs{\CurrentBib}

\bibitem [\protect \citeauthoryear {%
Stoltz%
\ \protect \BOthers {.}}{%
Stoltz%
\ \protect \BOthers {.}}{%
{\protect \APACyear {2017}}%
}]{%
stoltzAugmentedRealityWarehouse2017}
\APACinsertmetastar {%
stoltzAugmentedRealityWarehouse2017}%
\begin{APACrefauthors}%
Stoltz, M\BHBI H.%
, Giannikas, V.%
, McFarlane, D.%
, Strachan, J.%
, Um, J.%
\BCBL {} Srinivasan, R.%
\end{APACrefauthors}%
\unskip\
\newblock
\APACrefYearMonthDay{2017}{{\APACmonth{07}}}{}.
\newblock
{\BBOQ}\APACrefatitle {Augmented {{Reality}} in {{Warehouse Operations}}: {{Opportunities}} and {{Barriers}}} {Augmented {{Reality}} in {{Warehouse Operations}}: {{Opportunities}} and {{Barriers}}}.{\BBCQ}
\newblock
\APACjournalVolNumPages{IFAC-PapersOnLine}{50}{1}{12979--12984,}
\newblock
\begin{APACrefDOI} \doi{10.1016/j.ifacol.2017.08.1807} \end{APACrefDOI}
\newblock

\newblock

\PrintBackRefs{\CurrentBib}

\bibitem [\protect \citeauthoryear {%
Subramani%
, Matton%
, Greaves%
\BCBL {}\ \BBA {} Lam%
}{%
Subramani%
\ \protect \BOthers {.}}{%
{\protect \APACyear {2021}}%
}]{%
subramaniSurveyDeepLearning2021}
\APACinsertmetastar {%
subramaniSurveyDeepLearning2021}%
\begin{APACrefauthors}%
Subramani, N.%
, Matton, A.%
, Greaves, M.%
\BCBL {} Lam, A.%
\end{APACrefauthors}%
\unskip\
\newblock
\APACrefYearMonthDay{2021}{{\APACmonth{02}}}{}.
\newblock
\APACrefbtitle {A {{Survey}} of {{Deep Learning Approaches}} for {{OCR}} and {{Document Understanding}}} {A {{Survey}} of {{Deep Learning Approaches}} for {{OCR}} and {{Document Understanding}}}\ (\BNUM\ arXiv:2011.13534).
\newblock
\APACaddressPublisher{}{{arXiv}}.
\PrintBackRefs{\CurrentBib}

\bibitem [\protect \citeauthoryear {%
Suh%
, Lee%
, Lee%
, Lukowicz%
\BCBL {}\ \BBA {} Hwang%
}{%
Suh%
\ \protect \BOthers {.}}{%
{\protect \APACyear {2019}}%
}]{%
suhRobustShippingLabel2019}
\APACinsertmetastar {%
suhRobustShippingLabel2019}%
\begin{APACrefauthors}%
Suh, S.%
, Lee, H.%
, Lee, Y.O.%
, Lukowicz, P.%
\BCBL {} Hwang, J.%
\end{APACrefauthors}%
\unskip\
\newblock
\APACrefYearMonthDay{2019}{{\APACmonth{09}}}{}.
\newblock
{\BBOQ}\APACrefatitle {Robust {{Shipping Label Recognition}} and {{Validation}} for {{Logistics}} by {{Using Deep Neural Networks}}} {Robust {{Shipping Label Recognition}} and {{Validation}} for {{Logistics}} by {{Using Deep Neural Networks}}}.{\BBCQ}
\newblock
 \APACrefbtitle {2019 {{IEEE International Conference}} on {{Image Processing}} ({{ICIP}})} {2019 {{IEEE International Conference}} on {{Image Processing}} ({{ICIP}})}\ (\BPGS\ 4509--4513).
\PrintBackRefs{\CurrentBib}

\bibitem [\protect \citeauthoryear {%
Sun%
\ \protect \BOthers {.}}{%
Sun%
\ \protect \BOthers {.}}{%
{\protect \APACyear {2020}}%
}]{%
sunObjectRecognitionVolume2020}
\APACinsertmetastar {%
sunObjectRecognitionVolume2020}%
\begin{APACrefauthors}%
Sun, Y.%
, Liu, Z.X.%
, Li, M.%
, Zeng, Z.T.%
, Zong, Z.X.%
\BCBL {} Ji, C.L.%
\end{APACrefauthors}%
\unskip\
\newblock
\APACrefYearMonthDay{2020}{{\APACmonth{11}}}{}.
\newblock
{\BBOQ}\APACrefatitle {An {{Object Recognition}} and {{Volume Calculation Method Based}} on {{Yolov3}} and {{Depth Vision}}} {An {{Object Recognition}} and {{Volume Calculation Method Based}} on {{Yolov3}} and {{Depth Vision}}}.{\BBCQ}
\newblock
\APACjournalVolNumPages{Journal of Physics: Conference Series}{1684}{1}{012009,}
\newblock
\begin{APACrefDOI} \doi{10/gpc9mq} \end{APACrefDOI}
\newblock

\newblock

\PrintBackRefs{\CurrentBib}

\bibitem [\protect \citeauthoryear {%
Sun%
, Zheng%
, Yang%
, Tian%
\BCBL {}\ \BBA {} Wang%
}{%
Sun%
\ \protect \BOthers {.}}{%
{\protect \APACyear {2018}}%
}]{%
sunPartModelsPerson2018}
\APACinsertmetastar {%
sunPartModelsPerson2018}%
\begin{APACrefauthors}%
Sun, Y.%
, Zheng, L.%
, Yang, Y.%
, Tian, Q.%
\BCBL {} Wang, S.%
\end{APACrefauthors}%
\unskip\
\newblock
\APACrefYearMonthDay{2018}{}{}.
\newblock
{\BBOQ}\APACrefatitle {Beyond {{Part Models}}: {{Person Retrieval}} with {{Refined Part Pooling}} (and {{A Strong Convolutional Baseline}})} {Beyond {{Part Models}}: {{Person Retrieval}} with {{Refined Part Pooling}} (and {{A Strong Convolutional Baseline}})}.{\BBCQ}
\newblock
 V.~Ferrari, M.~Hebert, C.~Sminchisescu\BCBL {}\ \BBA {} Y.~Weiss\ (\BEDS), \APACrefbtitle {Computer {{Vision}} \textendash{} {{ECCV}} 2018} {Computer {{Vision}} \textendash{} {{ECCV}} 2018}\ (\BVOL\ 11208, \BPGS\ 501--518).
\newblock
\APACaddressPublisher{{Cham}}{{Springer International Publishing}}.
\PrintBackRefs{\CurrentBib}

\bibitem [\protect \citeauthoryear {%
Tang%
\ \BBA {} Veelenturf%
}{%
Tang%
\ \BBA {} Veelenturf%
}{%
{\protect \APACyear {2019}}%
}]{%
tangStrategicRoleLogistics2019}
\APACinsertmetastar {%
tangStrategicRoleLogistics2019}%
\begin{APACrefauthors}%
Tang, C.S.%
\BCBT {}\ \BBA {} Veelenturf, L.P.%
\end{APACrefauthors}%
\unskip\
\newblock
\APACrefYearMonthDay{2019}{{\APACmonth{09}}}{}.
\newblock
{\BBOQ}\APACrefatitle {The Strategic Role of Logistics in the Industry 4.0 Era} {The strategic role of logistics in the industry 4.0 era}.{\BBCQ}
\newblock
\APACjournalVolNumPages{Transportation Research Part E: Logistics and Transportation Review}{129}{}{1--11,}
\newblock
\begin{APACrefDOI} \doi{10.1016/j.tre.2019.06.004} \end{APACrefDOI}
\newblock

\newblock

\PrintBackRefs{\CurrentBib}

\bibitem [\protect \citeauthoryear {%
Thamer%
, Kost%
, Weimer%
\BCBL {}\ \BBA {} {Scholz-Reiter}%
}{%
Thamer%
\ \protect \BOthers {.}}{%
{\protect \APACyear {2013}}%
}]{%
thamer3DrobotVisionSystem2013}
\APACinsertmetastar {%
thamer3DrobotVisionSystem2013}%
\begin{APACrefauthors}%
Thamer, H.%
, Kost, H.%
, Weimer, D.%
\BCBL {} {Scholz-Reiter}, B.%
\end{APACrefauthors}%
\unskip\
\newblock
\APACrefYearMonthDay{2013}{{\APACmonth{09}}}{}.
\newblock
{\BBOQ}\APACrefatitle {A {{3D-robot}} Vision System for Automatic Unloading of Containers} {A {{3D-robot}} vision system for automatic unloading of containers}.{\BBCQ}
\newblock
 \APACrefbtitle {2013 {{IEEE}} 18th {{Conference}} on {{Emerging Technologies Factory Automation}} ({{ETFA}})} {2013 {{IEEE}} 18th {{Conference}} on {{Emerging Technologies Factory Automation}} ({{ETFA}})}\ (\BPGS\ 1--7).
\PrintBackRefs{\CurrentBib}

\bibitem [\protect \citeauthoryear {%
Thamer%
, Weimer%
, Kost%
\BCBL {}\ \BBA {} {Scholz-Reiter}%
}{%
Thamer%
\ \protect \BOthers {.}}{%
{\protect \APACyear {2014}}%
}]{%
thamer3DComputerVisionAutomation2014}
\APACinsertmetastar {%
thamer3DComputerVisionAutomation2014}%
\begin{APACrefauthors}%
Thamer, H.%
, Weimer, D.%
, Kost, H.%
\BCBL {} {Scholz-Reiter}, B.%
\end{APACrefauthors}%
\unskip\
\newblock
\APACrefYearMonthDay{2014}{}{}.
\newblock
{\BBOQ}\APACrefatitle {{{3D-Computer Vision}} for {{Automation}} of {{Logistic Processes}}} {{{3D-Computer Vision}} for {{Automation}} of {{Logistic Processes}}}.{\BBCQ}
\newblock
 U.~Clausen, M.~{ten Hompel}\BCBL {}\ \BBA {} J.F.~Meier\ (\BEDS), \APACrefbtitle {Efficiency and {{Innovation}} in {{Logistics}}} {Efficiency and {{Innovation}} in {{Logistics}}}\ (\BPGS\ 67--75).
\newblock
\APACaddressPublisher{{Cham}}{{Springer International Publishing}}.
\PrintBackRefs{\CurrentBib}

\bibitem [\protect \citeauthoryear {%
Thomas%
, MacDonald%
\BCBL {}\ \BBA {} Stol%
}{%
Thomas%
\ \protect \BOthers {.}}{%
{\protect \APACyear {2011}}%
}]{%
thomasRealtimeRobustImage2011}
\APACinsertmetastar {%
thomasRealtimeRobustImage2011}%
\begin{APACrefauthors}%
Thomas, S.%
, MacDonald, B.%
\BCBL {} Stol, K.%
\end{APACrefauthors}%
\unskip\
\newblock
\APACrefYearMonthDay{2011}{}{}.
\newblock
{\BBOQ}\APACrefatitle {Real-Time Robust Image Feature Description and Matching} {Real-time robust image feature description and matching}.{\BBCQ}
\newblock
\APACaddressPublisher{}{{Springer-Verlag Berlin Heidelberg}}.
\PrintBackRefs{\CurrentBib}

\bibitem [\protect \citeauthoryear {%
Tsolaki%
, Vafeiadis%
, Nizamis%
, Ioannidis%
\BCBL {}\ \BBA {} Tzovaras%
}{%
Tsolaki%
\ \protect \BOthers {.}}{%
{\protect \APACyear {2022}}%
}]{%
tsolakiUtilizingMachineLearning2022a}
\APACinsertmetastar {%
tsolakiUtilizingMachineLearning2022a}%
\begin{APACrefauthors}%
Tsolaki, K.%
, Vafeiadis, T.%
, Nizamis, A.%
, Ioannidis, D.%
\BCBL {} Tzovaras, D.%
\end{APACrefauthors}%
\unskip\
\newblock
\APACrefYearMonthDay{2022}{{\APACmonth{02}}}{}.
\newblock
{\BBOQ}\APACrefatitle {Utilizing Machine Learning on Freight Transportation and Logistics Applications: {{A}} Review} {Utilizing machine learning on freight transportation and logistics applications: {{A}} review}.{\BBCQ}
\newblock
\APACjournalVolNumPages{ICT Express}{}{}{S2405959522000200,}
\newblock
\begin{APACrefDOI} \doi{10.1016/j.icte.2022.02.001} \end{APACrefDOI}
\newblock

\newblock

\PrintBackRefs{\CurrentBib}

\bibitem [\protect \citeauthoryear {%
United~Nations%
}{%
United~Nations%
}{%
{\protect \APACyear {2019}}%
}]{%
unitednationsRecommendationsTransportDangerous2019}
\APACinsertmetastar {%
unitednationsRecommendationsTransportDangerous2019}%
\begin{APACrefauthors}%
United~Nations, W.G.%
\end{APACrefauthors}%
\unskip\
\newblock
\APACrefYear{2019}.
\newblock
\APACrefbtitle {Recommendations on the Transport of Dangerous Goods: Model Regulations} {Recommendations on the transport of dangerous goods: Model regulations}\ (\BNUM\ SPECA/PWG-TBC(19)/8).
\newblock
\APACaddressPublisher{{New York and Geneva}}{{United Nations}}.
\PrintBackRefs{\CurrentBib}

\bibitem [\protect \citeauthoryear {%
Varga%
, Costea%
\BCBL {}\ \BBA {} Nedevschi%
}{%
Varga%
\ \protect \BOthers {.}}{%
{\protect \APACyear {2015}}%
}]{%
vargaImprovedAutonomousLoad2015}
\APACinsertmetastar {%
vargaImprovedAutonomousLoad2015}%
\begin{APACrefauthors}%
Varga, R.%
, Costea, A.%
\BCBL {} Nedevschi, S.%
\end{APACrefauthors}%
\unskip\
\newblock
\APACrefYearMonthDay{2015}{{\APACmonth{09}}}{}.
\newblock
{\BBOQ}\APACrefatitle {Improved Autonomous Load Handling with Stereo Cameras} {Improved autonomous load handling with stereo cameras}.{\BBCQ}
\newblock
 \APACrefbtitle {2015 {{IEEE International Conference}} on {{Intelligent Computer Communication}} and {{Processing}} ({{ICCP}})} {2015 {{IEEE International Conference}} on {{Intelligent Computer Communication}} and {{Processing}} ({{ICCP}})}\ (\BPGS\ 251--256).
\newblock
\APACaddressPublisher{{Cluj-Napoca, Romania}}{{IEEE}}.
\PrintBackRefs{\CurrentBib}

\bibitem [\protect \citeauthoryear {%
Varga%
\ \BBA {} Nedevschi%
}{%
Varga%
\ \BBA {} Nedevschi%
}{%
{\protect \APACyear {2014}}%
}]{%
vargaVisionbasedAutonomousLoad2014}
\APACinsertmetastar {%
vargaVisionbasedAutonomousLoad2014}%
\begin{APACrefauthors}%
Varga, R.%
\BCBT {}\ \BBA {} Nedevschi, S.%
\end{APACrefauthors}%
\unskip\
\newblock
\APACrefYearMonthDay{2014}{{\APACmonth{09}}}{}.
\newblock
{\BBOQ}\APACrefatitle {Vision-Based Autonomous Load Handling for Automated Guided Vehicles} {Vision-based autonomous load handling for automated guided vehicles}.{\BBCQ}
\newblock
 \APACrefbtitle {2014 {{IEEE}} 10th {{International Conference}} on {{Intelligent Computer Communication}} and {{Processing}} ({{ICCP}})} {2014 {{IEEE}} 10th {{International Conference}} on {{Intelligent Computer Communication}} and {{Processing}} ({{ICCP}})}\ (\BPGS\ 239--244).
\PrintBackRefs{\CurrentBib}

\bibitem [\protect \citeauthoryear {%
Varga%
\ \BBA {} Nedevschi%
}{%
Varga%
\ \BBA {} Nedevschi%
}{%
{\protect \APACyear {2016}}%
}]{%
vargaRobustPalletDetection2016}
\APACinsertmetastar {%
vargaRobustPalletDetection2016}%
\begin{APACrefauthors}%
Varga, R.%
\BCBT {}\ \BBA {} Nedevschi, S.%
\end{APACrefauthors}%
\unskip\
\newblock
\APACrefYearMonthDay{2016}{}{}.
\newblock
{\BBOQ}\APACrefatitle {Robust {{Pallet Detection}} for {{Automated Logistics Operations}}:} {Robust {{Pallet Detection}} for {{Automated Logistics Operations}}:}.{\BBCQ}
\newblock
 \APACrefbtitle {Proceedings of the 11th {{Joint Conference}} on {{Computer Vision}}, {{Imaging}} and {{Computer Graphics Theory}} and {{Applications}}} {Proceedings of the 11th {{Joint Conference}} on {{Computer Vision}}, {{Imaging}} and {{Computer Graphics Theory}} and {{Applications}}}\ (\BPGS\ 470--477).
\newblock
\APACaddressPublisher{{Rome, Italy}}{{SCITEPRESS - Science and and Technology Publications}}.
\PrintBackRefs{\CurrentBib}

\bibitem [\protect \citeauthoryear {%
{Vargas-Osorio}%
\ \BBA {} Zuniga%
}{%
{Vargas-Osorio}%
\ \BBA {} Zuniga%
}{%
{\protect \APACyear {2016}}%
}]{%
vargas-osorioLiteratureReviewPallet2016}
\APACinsertmetastar {%
vargas-osorioLiteratureReviewPallet2016}%
\begin{APACrefauthors}%
{Vargas-Osorio}, S.%
\BCBT {}\ \BBA {} Zuniga, C.%
\end{APACrefauthors}%
\unskip\
\newblock
\APACrefYearMonthDay{2016}{}{}.
\newblock
{\BBOQ}\APACrefatitle {A Literature Review on the Pallet Loading Problem} {A literature review on the pallet loading problem}.{\BBCQ}
\newblock
\APACjournalVolNumPages{L\'ampsakos}{15}{}{,}
\newblock
\begin{APACrefDOI} \doi{http://dx.doi.org/10.21501/issn.2145-4086} \end{APACrefDOI}
\newblock

\newblock

\PrintBackRefs{\CurrentBib}

\bibitem [\protect \citeauthoryear {%
Wahrmann%
, Hildebrandt%
, Schuetz%
, Wittmann%
\BCBL {}\ \BBA {} Rixen%
}{%
Wahrmann%
\ \protect \BOthers {.}}{%
{\protect \APACyear {2019}}%
}]{%
wahrmannAutonomousFlexibleRobotic2019}
\APACinsertmetastar {%
wahrmannAutonomousFlexibleRobotic2019}%
\begin{APACrefauthors}%
Wahrmann, D.%
, Hildebrandt, A\BHBI C.%
, Schuetz, C.%
, Wittmann, R.%
\BCBL {} Rixen, D.%
\end{APACrefauthors}%
\unskip\
\newblock
\APACrefYearMonthDay{2019}{{\APACmonth{03}}}{}.
\newblock
{\BBOQ}\APACrefatitle {An {{Autonomous}} and {{Flexible Robotic Framework}} for {{Logistics Applications}}} {An {{Autonomous}} and {{Flexible Robotic Framework}} for {{Logistics Applications}}}.{\BBCQ}
\newblock
\APACjournalVolNumPages{Journal of Intelligent \& Robotic Systems}{93}{3}{419--431,}
\newblock
\begin{APACrefDOI} \doi{10/gpc9mh} \end{APACrefDOI}
\newblock

\newblock

\PrintBackRefs{\CurrentBib}

\bibitem [\protect \citeauthoryear {%
N.~Wang%
\ \protect \BOthers {.}}{%
N.~Wang%
\ \protect \BOthers {.}}{%
{\protect \APACyear {2018}}%
}]{%
wangPixel2MeshGenerating3D2018}
\APACinsertmetastar {%
wangPixel2MeshGenerating3D2018}%
\begin{APACrefauthors}%
Wang, N.%
, Zhang, Y.%
, Li, Z.%
, Fu, Y.%
, Liu, W.%
\BCBL {} Jiang, Y\BHBI G.%
\end{APACrefauthors}%
\unskip\
\newblock
\APACrefYearMonthDay{2018}{}{}.
\newblock
{\BBOQ}\APACrefatitle {{{Pixel2Mesh}}: {{Generating 3D Mesh Models}} from {{Single RGB Images}}} {{{Pixel2Mesh}}: {{Generating 3D Mesh Models}} from {{Single RGB Images}}}.{\BBCQ}
\newblock
 V.~Ferrari, M.~Hebert, C.~Sminchisescu\BCBL {}\ \BBA {} Y.~Weiss\ (\BEDS), \APACrefbtitle {European {{Conference}} on {{Computer Vision}} ({{ECCV}})} {European {{Conference}} on {{Computer Vision}} ({{ECCV}})}\ (\BVOL\ 11215, \BPGS\ 55--71).
\newblock
\APACaddressPublisher{{Cham}}{{Springer International Publishing}}.
\PrintBackRefs{\CurrentBib}

\bibitem [\protect \citeauthoryear {%
Y.~Wang%
, Zhou%
, Lu%
\BCBL {}\ \BBA {} Li%
}{%
Y.~Wang%
\ \protect \BOthers {.}}{%
{\protect \APACyear {2022}}%
}]{%
wangUDocGANUnpairedDocument2022}
\APACinsertmetastar {%
wangUDocGANUnpairedDocument2022}%
\begin{APACrefauthors}%
Wang, Y.%
, Zhou, W.%
, Lu, Z.%
\BCBL {} Li, H.%
\end{APACrefauthors}%
\unskip\
\newblock
\APACrefYearMonthDay{2022}{{\APACmonth{10}}}{}.
\newblock
{\BBOQ}\APACrefatitle {{{UDoc-GAN}}: {{Unpaired Document Illumination Correction}} with {{Background Light Prior}}} {{{UDoc-GAN}}: {{Unpaired Document Illumination Correction}} with {{Background Light Prior}}}.{\BBCQ}
\newblock
 \APACrefbtitle {Proceedings of the 30th {{ACM International Conference}} on {{Multimedia}}} {Proceedings of the 30th {{ACM International Conference}} on {{Multimedia}}}\ (\BPGS\ 5074--5082).
\newblock
\APACaddressPublisher{{New York, NY, USA}}{{Association for Computing Machinery}}.
\PrintBackRefs{\CurrentBib}

\bibitem [\protect \citeauthoryear {%
Weichert%
\ \protect \BOthers {.}}{%
Weichert%
\ \protect \BOthers {.}}{%
{\protect \APACyear {2010}}%
}]{%
weichertMarkerbasedTrackingSupport2010}
\APACinsertmetastar {%
weichertMarkerbasedTrackingSupport2010}%
\begin{APACrefauthors}%
Weichert, F.%
, Fiedler, D.%
, Hegenberg, J.%
, M{\"u}ller, H.%
, Prasse, C.%
, Roidl, M.%
\BCBL {} {ten Hompel}, M.%
\end{APACrefauthors}%
\unskip\
\newblock
\APACrefYearMonthDay{2010}{{\APACmonth{06}}}{}.
\newblock
{\BBOQ}\APACrefatitle {Marker-Based Tracking in Support of {{RFID}} Controlled Material Flow Systems} {Marker-based tracking in support of {{RFID}} controlled material flow systems}.{\BBCQ}
\newblock
\APACjournalVolNumPages{Logistics Research}{2}{1}{13--21,}
\newblock
\begin{APACrefDOI} \doi{10.1007/s12159-010-0025-6} \end{APACrefDOI}
\newblock

\newblock

\PrintBackRefs{\CurrentBib}

\bibitem [\protect \citeauthoryear {%
Woschank%
, Rauch%
\BCBL {}\ \BBA {} Zsifkovits%
}{%
Woschank%
\ \protect \BOthers {.}}{%
{\protect \APACyear {2020}}%
}]{%
woschankReviewFurtherDirections2020}
\APACinsertmetastar {%
woschankReviewFurtherDirections2020}%
\begin{APACrefauthors}%
Woschank, M.%
, Rauch, E.%
\BCBL {} Zsifkovits, H.%
\end{APACrefauthors}%
\unskip\
\newblock
\APACrefYearMonthDay{2020}{{\APACmonth{01}}}{}.
\newblock
{\BBOQ}\APACrefatitle {A {{Review}} of {{Further Directions}} for {{Artificial Intelligence}}, {{Machine Learning}}, and {{Deep Learning}} in {{Smart Logistics}}} {A {{Review}} of {{Further Directions}} for {{Artificial Intelligence}}, {{Machine Learning}}, and {{Deep Learning}} in {{Smart Logistics}}}.{\BBCQ}
\newblock
\APACjournalVolNumPages{Sustainability}{12}{9}{3760,}
\newblock
\begin{APACrefDOI} \doi{10/ghf4b3} \end{APACrefDOI}
\newblock

\newblock

\PrintBackRefs{\CurrentBib}

\bibitem [\protect \citeauthoryear {%
Wudhikarn%
, Charoenkwan%
\BCBL {}\ \BBA {} Malang%
}{%
Wudhikarn%
\ \protect \BOthers {.}}{%
{\protect \APACyear {2022}}%
}]{%
wudhikarnDeepLearningBarcode2022}
\APACinsertmetastar {%
wudhikarnDeepLearningBarcode2022}%
\begin{APACrefauthors}%
Wudhikarn, R.%
, Charoenkwan, P.%
\BCBL {} Malang, K.%
\end{APACrefauthors}%
\unskip\
\newblock
\APACrefYearMonthDay{2022}{}{}.
\newblock
{\BBOQ}\APACrefatitle {Deep {{Learning}} in {{Barcode Recognition}}: {{A Systematic Literature Review}}} {Deep {{Learning}} in {{Barcode Recognition}}: {{A Systematic Literature Review}}}.{\BBCQ}
\newblock
\APACjournalVolNumPages{IEEE Access}{10}{}{8049--8072,}
\newblock
\begin{APACrefDOI} \doi{10.1109/ACCESS.2022.3143033} \end{APACrefDOI}
\newblock

\newblock

\PrintBackRefs{\CurrentBib}

\bibitem [\protect \citeauthoryear {%
Xiao%
, Lu%
, Zhang%
\BCBL {}\ \BBA {} Zhang%
}{%
Xiao%
\ \protect \BOthers {.}}{%
{\protect \APACyear {2017}}%
}]{%
xiaoPalletRecognitionLocalization2017}
\APACinsertmetastar {%
xiaoPalletRecognitionLocalization2017}%
\begin{APACrefauthors}%
Xiao, J.%
, Lu, H.%
, Zhang, L.%
\BCBL {} Zhang, J.%
\end{APACrefauthors}%
\unskip\
\newblock
\APACrefYearMonthDay{2017}{{\APACmonth{11}}}{}.
\newblock
{\BBOQ}\APACrefatitle {Pallet Recognition and Localization Using an {{RGB-D}} Camera} {Pallet recognition and localization using an {{RGB-D}} camera}.{\BBCQ}
\newblock
\APACjournalVolNumPages{International Journal of Advanced Robotic Systems}{14}{6}{172988141773779,}
\newblock
\begin{APACrefDOI} \doi{10.1177/1729881417737799} \end{APACrefDOI}
\newblock

\newblock

\PrintBackRefs{\CurrentBib}

\bibitem [\protect \citeauthoryear {%
Xiao%
, Zhang%
, Adler%
, Zhang%
\BCBL {}\ \BBA {} Zhang%
}{%
Xiao%
\ \protect \BOthers {.}}{%
{\protect \APACyear {2013}}%
}]{%
xiaoThreedimensionalPointCloud2013}
\APACinsertmetastar {%
xiaoThreedimensionalPointCloud2013}%
\begin{APACrefauthors}%
Xiao, J.%
, Zhang, J.%
, Adler, B.%
, Zhang, H.%
\BCBL {} Zhang, J.%
\end{APACrefauthors}%
\unskip\
\newblock
\APACrefYearMonthDay{2013}{{\APACmonth{12}}}{}.
\newblock
{\BBOQ}\APACrefatitle {Three-Dimensional Point Cloud Plane Segmentation in Both Structured and Unstructured Environments} {Three-dimensional point cloud plane segmentation in both structured and unstructured environments}.{\BBCQ}
\newblock
\APACjournalVolNumPages{Robotics and Autonomous Systems}{61}{12}{1641--1652,}
\newblock
\begin{APACrefDOI} \doi{10.1016/j.robot.2013.07.001} \end{APACrefDOI}
\newblock

\newblock

\PrintBackRefs{\CurrentBib}

\bibitem [\protect \citeauthoryear {%
E.~Xie%
\ \protect \BOthers {.}}{%
E.~Xie%
\ \protect \BOthers {.}}{%
{\protect \APACyear {2021}}%
}]{%
xieSegFormerSimpleEfficient2021}
\APACinsertmetastar {%
xieSegFormerSimpleEfficient2021}%
\begin{APACrefauthors}%
Xie, E.%
, Wang, W.%
, Yu, Z.%
, Anandkumar, A.%
, Alvarez, J.M.%
\BCBL {} Luo, P.%
\end{APACrefauthors}%
\unskip\
\newblock
\APACrefYearMonthDay{2021}{}{}.
\newblock
{\BBOQ}\APACrefatitle {{{SegFormer}}: {{Simple}} and {{Efficient Design}} for {{Semantic Segmentation}} with {{Transformers}}} {{{SegFormer}}: {{Simple}} and {{Efficient Design}} for {{Semantic Segmentation}} with {{Transformers}}}.{\BBCQ}
\newblock
 \APACrefbtitle {Proceedings of the 35th {{Conference}} on {{Neural Information Processing Systems}}.} {Proceedings of the 35th {{Conference}} on {{Neural Information Processing Systems}}.}
\PrintBackRefs{\CurrentBib}

\bibitem [\protect \citeauthoryear {%
G\BHBI W.~Xie%
, Yin%
, Zhang%
\BCBL {}\ \BBA {} Liu%
}{%
G\BHBI W.~Xie%
\ \protect \BOthers {.}}{%
{\protect \APACyear {2020}}%
}]{%
xieDewarpingDocumentImage2020}
\APACinsertmetastar {%
xieDewarpingDocumentImage2020}%
\begin{APACrefauthors}%
Xie, G\BHBI W.%
, Yin, F.%
, Zhang, X\BHBI Y.%
\BCBL {} Liu, C\BHBI L.%
\end{APACrefauthors}%
\unskip\
\newblock
\APACrefYearMonthDay{2020}{}{}.
\newblock
{\BBOQ}\APACrefatitle {Dewarping {{Document Image}} by {{Displacement Flow Estimation}} with {{Fully Convolutional Network}}} {Dewarping {{Document Image}} by {{Displacement Flow Estimation}} with {{Fully Convolutional Network}}}.{\BBCQ}
\newblock
 X.~Bai, D.~Karatzas\BCBL {}\ \BBA {} D.~Lopresti\ (\BEDS), \APACrefbtitle {Document {{Analysis Systems}}} {Document {{Analysis Systems}}}\ (\BPGS\ 131--144).
\newblock
\APACaddressPublisher{{Cham}}{{Springer International Publishing}}.
\PrintBackRefs{\CurrentBib}

\bibitem [\protect \citeauthoryear {%
Xue%
, Tian%
, Zhan%
, Lu%
\BCBL {}\ \BBA {} Bai%
}{%
Xue%
\ \protect \BOthers {.}}{%
{\protect \APACyear {2022}}%
}]{%
xueFourierDocumentRestoration2022}
\APACinsertmetastar {%
xueFourierDocumentRestoration2022}%
\begin{APACrefauthors}%
Xue, C.%
, Tian, Z.%
, Zhan, F.%
, Lu, S.%
\BCBL {} Bai, S.%
\end{APACrefauthors}%
\unskip\
\newblock
\APACrefYearMonthDay{2022}{{\APACmonth{06}}}{}.
\newblock
{\BBOQ}\APACrefatitle {Fourier {{Document Restoration}} for {{Robust Document Dewarping}} and {{Recognition}}} {Fourier {{Document Restoration}} for {{Robust Document Dewarping}} and {{Recognition}}}.{\BBCQ}
\newblock
 \APACrefbtitle {2022 {{IEEE}}/{{CVF Conference}} on {{Computer Vision}} and {{Pattern Recognition}} ({{CVPR}})} {2022 {{IEEE}}/{{CVF Conference}} on {{Computer Vision}} and {{Pattern Recognition}} ({{CVPR}})}\ (\BPGS\ 4563--4572).
\newblock
\APACaddressPublisher{{New Orleans, LA, USA}}{{IEEE}}.
\PrintBackRefs{\CurrentBib}

\bibitem [\protect \citeauthoryear {%
B.~Yang%
\ \protect \BOthers {.}}{%
B.~Yang%
\ \protect \BOthers {.}}{%
{\protect \APACyear {2019}}%
}]{%
yangLearningObjectBounding2019}
\APACinsertmetastar {%
yangLearningObjectBounding2019}%
\begin{APACrefauthors}%
Yang, B.%
, Wang, J.%
, Clark, R.%
, Hu, Q.%
, Wang, S.%
, Markham, A.%
\BCBL {} Trigoni, N.%
\end{APACrefauthors}%
\unskip\
\newblock
\APACrefYearMonthDay{2019}{}{}.
\newblock
{\BBOQ}\APACrefatitle {Learning {{Object Bounding Boxes}} for {{3D Instance Segmentation}} on {{Point Clouds}}} {Learning {{Object Bounding Boxes}} for {{3D Instance Segmentation}} on {{Point Clouds}}}.{\BBCQ}
\newblock
 \APACrefbtitle {Proceedings of {{Advances}} in Neural Information Processing Systems} {Proceedings of {{Advances}} in neural information processing systems}\ (\BVOL~32).
\PrintBackRefs{\CurrentBib}

\bibitem [\protect \citeauthoryear {%
S.~Yang%
\ \BBA {} Scherer%
}{%
S.~Yang%
\ \BBA {} Scherer%
}{%
{\protect \APACyear {2019}}%
}]{%
yangCubeSLAMMonocular3D2019}
\APACinsertmetastar {%
yangCubeSLAMMonocular3D2019}%
\begin{APACrefauthors}%
Yang, S.%
\BCBT {}\ \BBA {} Scherer, S.%
\end{APACrefauthors}%
\unskip\
\newblock
\APACrefYearMonthDay{2019}{{\APACmonth{08}}}{}.
\newblock
{\BBOQ}\APACrefatitle {{{CubeSLAM}}: {{Monocular}} 3-{{D Object SLAM}}} {{{CubeSLAM}}: {{Monocular}} 3-{{D Object SLAM}}}.{\BBCQ}
\newblock
\APACjournalVolNumPages{IEEE Transactions on Robotics}{35}{4}{925--938,}
\newblock
\begin{APACrefDOI} \doi{10/gg652k} \end{APACrefDOI}
\newblock

\newblock

\PrintBackRefs{\CurrentBib}

\bibitem [\protect \citeauthoryear {%
Ye%
\ \protect \BOthers {.}}{%
Ye%
\ \protect \BOthers {.}}{%
{\protect \APACyear {2022}}%
}]{%
yeDeepLearningPerson2022}
\APACinsertmetastar {%
yeDeepLearningPerson2022}%
\begin{APACrefauthors}%
Ye, M.%
, Shen, J.%
, Lin, G.%
, Xiang, T.%
, Shao, L.%
\BCBL {} Hoi, S.C.H.%
\end{APACrefauthors}%
\unskip\
\newblock
\APACrefYearMonthDay{2022}{{\APACmonth{06}}}{}.
\newblock
{\BBOQ}\APACrefatitle {Deep {{Learning}} for {{Person Re-Identification}}: {{A Survey}} and {{Outlook}}} {Deep {{Learning}} for {{Person Re-Identification}}: {{A Survey}} and {{Outlook}}}.{\BBCQ}
\newblock
\APACjournalVolNumPages{IEEE Transactions on Pattern Analysis and Machine Intelligence}{44}{6}{2872--2893,}
\newblock
\begin{APACrefDOI} \doi{10.1109/TPAMI.2021.3054775} \end{APACrefDOI}
\newblock

\newblock

\PrintBackRefs{\CurrentBib}

\bibitem [\protect \citeauthoryear {%
Zeng%
\ \protect \BOthers {.}}{%
Zeng%
\ \protect \BOthers {.}}{%
{\protect \APACyear {2017}}%
}]{%
zeng2016multi}
\APACinsertmetastar {%
zeng2016multi}%
\begin{APACrefauthors}%
Zeng, A.%
, Yu, K\BHBI T.%
, Song, S.%
, Suo, D.%
, Walker~Jr, E.%
, Rodriguez, A.%
\BCBL {} Xiao, J.%
\end{APACrefauthors}%
\unskip\
\newblock
\APACrefYearMonthDay{2017}{}{}.
\newblock
{\BBOQ}\APACrefatitle {Multi-View Self-Supervised Deep Learning for {{6D}} Pose Estimation in the Amazon Picking Challenge} {Multi-view self-supervised deep learning for {{6D}} pose estimation in the amazon picking challenge}.{\BBCQ}
\newblock
 \APACrefbtitle {Proceedings of the {{IEEE}} International Conference on Robotics and Automation.} {Proceedings of the {{IEEE}} international conference on robotics and automation.}
\PrintBackRefs{\CurrentBib}

\bibitem [\protect \citeauthoryear {%
Zhao%
, Zheng%
, Xu%
\BCBL {}\ \BBA {} Wu%
}{%
Zhao%
\ \protect \BOthers {.}}{%
{\protect \APACyear {2019}}%
}]{%
zhaoObjectDetectionDeep2019}
\APACinsertmetastar {%
zhaoObjectDetectionDeep2019}%
\begin{APACrefauthors}%
Zhao, Z\BHBI Q.%
, Zheng, P.%
, Xu, S\BHBI T.%
\BCBL {} Wu, X.%
\end{APACrefauthors}%
\unskip\
\newblock
\APACrefYearMonthDay{2019}{{\APACmonth{11}}}{}.
\newblock
{\BBOQ}\APACrefatitle {Object {{Detection With Deep Learning}}: {{A Review}}} {Object {{Detection With Deep Learning}}: {{A Review}}}.{\BBCQ}
\newblock
\APACjournalVolNumPages{IEEE Transactions on Neural Networks and Learning Systems}{30}{11}{3212--3232,}
\newblock
\begin{APACrefDOI} \doi{10.1109/TNNLS.2018.2876865} \end{APACrefDOI}
\newblock

\newblock

\PrintBackRefs{\CurrentBib}

\bibitem [\protect \citeauthoryear {%
Zhou%
, Wang%
, Klein%
\BCBL {}\ \BBA {} Kai%
}{%
Zhou%
\ \protect \BOthers {.}}{%
{\protect \APACyear {2021}}%
}]{%
zhouObjectDetectionMapping2021}
\APACinsertmetastar {%
zhouObjectDetectionMapping2021}%
\begin{APACrefauthors}%
Zhou, B.%
, Wang, A.%
, Klein, J\BHBI F.%
\BCBL {} Kai, F.%
\end{APACrefauthors}%
\unskip\
\newblock
\APACrefYearMonthDay{2021}{{\APACmonth{09}}}{}.
\newblock
{\BBOQ}\APACrefatitle {Object {{Detection}} and {{Mapping}} with {{Bounding Box Constraints}}} {Object {{Detection}} and {{Mapping}} with {{Bounding Box Constraints}}}.{\BBCQ}
\newblock
 \APACrefbtitle {2021 {{IEEE International Conference}} on {{Multisensor Fusion}} and {{Integration}} for {{Intelligent Systems}} ({{MFI}})} {2021 {{IEEE International Conference}} on {{Multisensor Fusion}} and {{Integration}} for {{Intelligent Systems}} ({{MFI}})}\ (\BPGS\ 1--6).
\PrintBackRefs{\CurrentBib}

\bibitem [\protect \citeauthoryear {%
Zou%
, Chen%
, Shi%
, Guo%
\BCBL {}\ \BBA {} Ye%
}{%
Zou%
\ \protect \BOthers {.}}{%
{\protect \APACyear {2023}}%
}]{%
zouObjectDetection202023}
\APACinsertmetastar {%
zouObjectDetection202023}%
\begin{APACrefauthors}%
Zou, Z.%
, Chen, K.%
, Shi, Z.%
, Guo, Y.%
\BCBL {} Ye, J.%
\end{APACrefauthors}%
\unskip\
\newblock
\APACrefYearMonthDay{2023}{}{}.
\newblock
{\BBOQ}\APACrefatitle {Object {{Detection}} in 20 {{Years}}: {{A Survey}}} {Object {{Detection}} in 20 {{Years}}: {{A Survey}}}.{\BBCQ}
\newblock
\APACjournalVolNumPages{Proceedings of the IEEE}{}{}{1--20,}
\newblock
\begin{APACrefDOI} \doi{10.1109/JPROC.2023.3238524} \end{APACrefDOI}
\newblock

\newblock

\PrintBackRefs{\CurrentBib}

\end{thebibliography}

\end{document}